\documentclass{article} 
\usepackage[T1]{fontenc}
\usepackage{iclr2027_conference,times}
\usepackage[toc,page,header]{appendix}
\usepackage{minitoc}
\usepackage{etoc}

\usepackage{amsmath,amsfonts,bm}

\def\eqref#1{equation~\ref{#1}}

\def\1{\bm{1}}

\DeclareMathAlphabet{\mathsfit}{\encodingdefault}{\sfdefault}{m}{sl}
\SetMathAlphabet{\mathsfit}{bold}{\encodingdefault}{\sfdefault}{bx}{n}

\usepackage{hyperref}
\usepackage{url}
\usepackage{booktabs}
\usepackage{multirow}
\usepackage{graphicx}
\usepackage{wrapfig}
\usepackage{marvosym}

\title{When Time Series Foundation Models Fail
\\Under Cross-Channel Dependence}

\author{
  Jinmyeong Choi\qquad Jinkwan Jang\qquad Seul Lee\qquad Taesup Kim\textsuperscript{\Letter} \\ 
  Graduate School of Data Science \\ 
  Seoul National University \\ 
  \texttt{\{jinmyeongchoi, jkjang22, ds\_seul, taesup.kim\}@snu.ac.kr}
}

\iclrfinalcopy
\begin{document}

\maketitle
\begingroup
\makeatletter
\NoHyper
\renewcommand\thefootnote{\Letter }
\footnotetext{
    Corresponding author.
}
\endNoHyper
\makeatother
\endgroup

\begin{abstract}
Multivariate time series foundation models (TSFMs) can access related series and covariates, but whether their forecasts reliably reflect the predictive value of cross-channel information remains unclear. We examine how {TSFMs} use cross-channel information through controlled synthetic experiments and real-world forecasting tasks. Our analysis covers cross-channel temporal relationships, dependence structures among observed channels, and the effects of known-future conditioning and covariate reliability. Useful cross-channel relationships are not consistently reflected in forecasts, while redundant or conditionally irrelevant inputs can still influence predictions. Statistical baselines fitted to the same observed contexts recover the relationships that the models underuse, indicating that these shortfalls reflect limited use of available information rather than its absence. On real-world tasks, conditioning on each model’s own median covariate forecasts increases quantile loss and reduces interval coverage. Propagating covariate uncertainty recovers much of this loss, but not all. Together, these findings reveal model-dependent limitations in exploiting temporal dependencies, combining information across channels, and producing reliable forecasts under covariate uncertainty.
\end{abstract}

\section{Introduction}
\label{sec:introduction}

Related time series can provide predictive information beyond what is available in a target's own history, motivating the use of multivariate forecasting. Yet it remains unclear when and how this information should be exploited. Some approaches explicitly model cross-channel dependencies~\citep{zhang2023crossformer, liu2024itransformerinvertedtransformerseffective,wang2024timemixer++}, while channel-independent approaches that process each series separately with shared parameters have also achieved strong forecasting performance \citep{Yuqietal-2023-PatchTST}. 
Prior analyses suggest a trade-off between modeling capacity and robustness to distribution shifts \citep{han2023capacityrobustnesstradeoffrevisiting}.
These lines of work reflect a common challenge in multivariate forecasting: how to translate the information available across related series into improved predictions.

Early time series foundation models (TSFMs) were largely developed around univariate forecasting~\citep{garza2024timegpt1, goswami2024moment, liu2024timergenerativepretrainedtransformers, liu2025sundial, ansari2024chronoslearninglanguagetime,pmlr-v235-das24c}.
Other approaches subsequently incorporated covariates into pretrained forecasters through modular adaptation, in-context learning or cross-channel attention~\citep{arango2025chronosxadaptingpretrainedtime, auer2025zeroshottimeseriesforecasting, potosnak2026micamultivariateinfinicompressive}.
More recent TSFMs have expanded native support for multivariate forecasting and historical or known-future covariates \citep{ansari2025chronos2univariateuniversalforecasting, jain2026timesfm3, podest2026tirex2generalizingtirexmultivariate, meyer2026t0timeseriesfoundationmodel, yamaguchi2026citrasfmtinytimeseries}.
This expansion raises the prospect of using pretrained models not only to forecast individual series, but also to exploit relationships among them in new tasks.
Yet the ability to accept cross-channel inputs does not mean that a model uses them in accordance with their predictive value.

Assessing this alignment is challenging because cross-channel relationships vary in both their temporal structure and their predictive value.
A signal in one series may anticipate a later change in the target~\citep{LIFT}, while several series may provide complementary evidence about its future~\citep{taga2025timepfn, stock2002forecasting}.
Other inputs may instead repeat information already available or contribute no additional predictive information~\citep{944919.944968, lim2020temporalfusiontransformersinterpretable}.
Moreover, future covariate values may themselves be estimates, so their availability does not guarantee their reliability~\citep{hyndman2021fpp3}.
These cases make it necessary to distinguish the predictive information
provided by an input from the influence that input has on a model's forecast:
useful information may be underused, while redundant, irrelevant, or unreliable
inputs may still affect predictions. We therefore ask: \textit{does a model's use of cross-channel inputs align with the predictive value of those inputs?}

Forecasting benchmarks provide broad assessments of model performance across diverse tasks \citep{aksu2024giftevalbenchmarkgeneraltime,cohen2025timedifferentobservabilityperspective,shchur2026fevbenchrealisticbenchmarktime}, but aggregate accuracy alone offers limited insight into how specific cross-channel relationships contribute to a forecast.
The empirical benefits of cross-channel modeling also depend on the evaluation setting: TSMixer reports limited gains from feature mixing on common long-term forecasting benchmarks, but larger gains from cross-variate information on the M5 demand forecasting task \citep{Chen2023}.
More fundamentally, an improvement after adding related series does not establish which predictive relationships the model exploited.
Conversely, an absence of improvement does not distinguish a lack of useful information from a failure to use the information available.
These ambiguities motivate evaluations that examine how model forecasts respond to cross-channel inputs under explicitly specified predictive relationships, complementing comparisons of overall forecasting performance.

Prior diagnostics examine simple target--covariate
mappings~\citep{berthelier2026investigatingsimpletargetcovariaterelationships},
as well as leading indicators and robustness to additional
independent channels~\citep{ghoroghchian2026foundations}.
Our analysis instead centers on an alignment-based diagnostic:
whether the influence of a cross-channel input on the forecast
matches the predictive information it contributes.
This diagnostic distinguishes underuse of informative inputs from
sensitivity to inputs that add no predictive information.

We examine four complementary aspects of cross-channel information use.
First, responding to a temporally aligned covariate does not establish
that a model can connect an earlier signal to the appropriate later
target response \citep{LIFT}.
This motivates \textbf{(RQ1)}:
\emph{How do temporal relationships affect TSFMs' use of cross-channel information?} Second, multiple observations may provide complementary evidence
about a shared predictive state \citep{stock2002forecasting},
whereas duplicates or irrelevant channels add no predictive information
beyond the informative observations.
This distinction leads to \textbf{(RQ2)}:
\emph{How do TSFMs use cross-channel information under different
dependence structures?} A third consideration is that models may use historical information
differently across input and conditioning configurations, even when
the available predictive information is unchanged.
Accordingly, \textbf{(RQ3)} asks:
\emph{How does the conditioning interface affect TSFMs' use of
cross-channel information?} Finally, future covariates may themselves vary in reliability, particularly
when their future values are estimated rather than observed
\citep{hyndman2021fpp3}.
This motivates \textbf{(RQ4)}:
\emph{How does covariate reliability affect TSFMs' use of
cross-channel information?}

We evaluate frozen Chronos-2, TimesFM-3, TiRex-2{, $t_0$, and
CITRAS-FM} on controlled synthetic processes with explicitly specified cross-channel
relationships.
The generating processes establish the predictive value of the
supplied inputs, while matched input comparisons measure their influence on forecasts.
Oracle references use knowledge of the data-generating process, whereas statistical baselines fitted separately to the same
observed contexts assess achievable performance under explicit model-class assumptions
(Appendix~\ref{app:context_fitted_protocol}).
On three real-world tasks, we compare direct multivariate forecasting with conditioning on each model's own covariate forecasts.

\vspace{-0.26cm}

\begin{figure}[t]
    \centering
    \includegraphics[width=\linewidth]{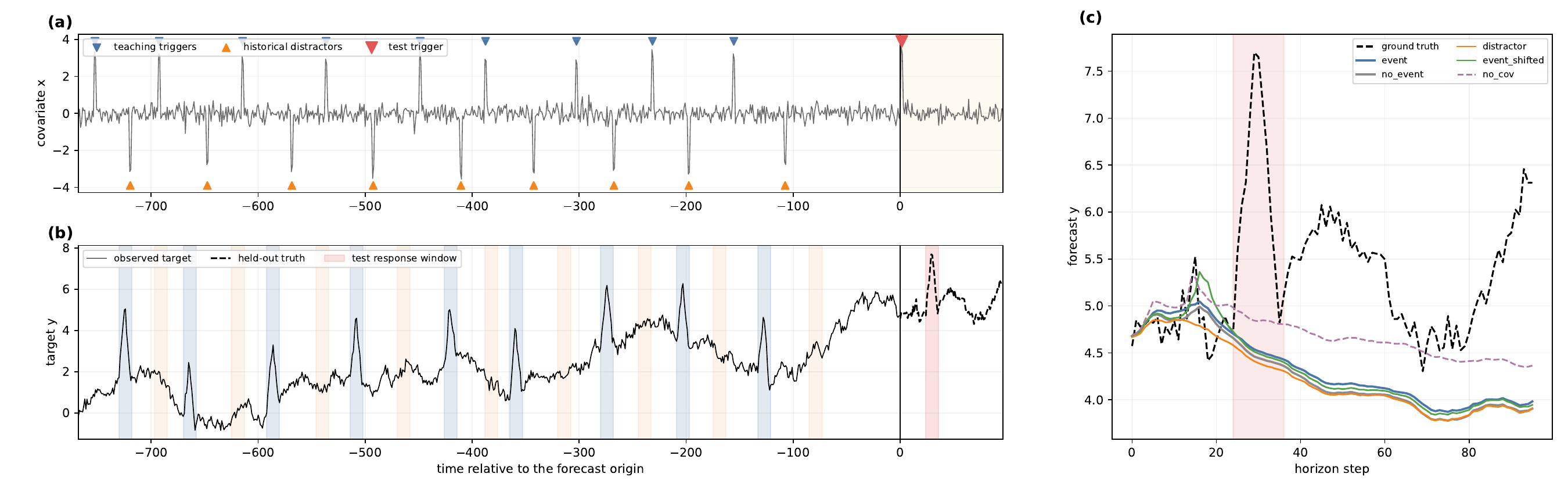}
\caption{
A spike trigger with a positive target response at $\tau=24$.
(a) Covariate triggers and matched distractors.
(b) Delayed target responses, with the test response fixed in the forecast horizon and the test trigger anchored at the forecast origin.
(c) Counterfactual TimesFM-3 forecasts for a single window.
Aggregate response gains are reported in Figure~\ref{fig:temporal_response_gain}.
}
    \label{fig:temporal_event_response}
\end{figure}

\section{Temporal covariate event--target response underutilization}
\label{sec:temporal_underutilization}

A pattern in a covariate series can signal a corresponding change in a target series.
For example, a brief burst of rainfall can signal a streamflow
response that rises, peaks, and recedes \citep{weaver2003methods}.
To address \textbf{RQ1}, we examine whether TSFMs use such cues to produce the
corresponding target response across different event--response patterns,
and how this use changes as the temporal separation between the trigger
and response increases.
We first evaluate responses across different event--response patterns,
and then vary the trigger--response lag while holding the target-response
position fixed within each sweep.

\subsection{Event--Response Pattern Combinations}

\paragraph{Trigger Patterns, Target Responses, and Historical Examples.}
We cross three auxiliary-trigger patterns---a positive spike,
a negative dip, and a biphasic (positive--negative) pulse---with
positive, negative, and biphasic target responses, yielding nine
event--response combinations.
Each condition contains nine historical teaching pairs that
repeat its single trigger--response mapping, together with
sign-reversed distractor patterns that produce no target response.
Figure~\ref{fig:temporal_event_response} illustrates a spike
trigger with a positive target response.
The kernels and generation settings are specified in
Appendix~\ref{app:temporal_setup}.

\paragraph{Varying the Delay between Events and Responses.}
We denote the target-response anchor by $t_{\mathrm{resp}}$,
measured relative to the forecast origin $t=0$, and place the
trigger at $t_{\mathrm{trig}}=t_{\mathrm{resp}}-\tau$.
We hold $t_{\mathrm{resp}}$ fixed within each lag sweep, so that
changing the lag does not move the target response within the
forecast horizon.
At $\tau=0$, the trigger and response anchors coincide.
The covariate history and future continuation are supplied
through the same conditioning interface at every lag.
The trigger lies in the known-future segment for
$\tau<t_{\mathrm{resp}}$, at the forecast origin for
$\tau=t_{\mathrm{resp}}$, and in history for $\tau>t_{\mathrm{resp}}$.
We compare lags within the future and historical segments
separately to keep the trigger's input segment fixed.
The main experiments use $t_{\mathrm{resp}}=24$, with additional
response-position controls reported in
Appendix~\ref{app:temporal_position_controls}.

\paragraph{Measuring Event-Induced Target Responses.}
We subtract the no-event forecast from the event forecast
for paired inputs that differ only in the presence of the test trigger.
For each combination $c$ and seed $s$, averaging these differences
over windows gives the response profile $\overline D_{c,s}$.
With $Q_c$ denoting the true injected response at the fixed anchor,
we measure its signed strength using projection gain:
\begin{equation}
    G_{c,s}
    =\frac{\langle\overline D_{c,s},Q_c\rangle}
           {\langle Q_c,Q_c\rangle}.
\end{equation}
The ideal response has unit gain; negative gain indicates a
component opposite to the expected response pattern.
We compute gain within each combination and seed, then average
equally across the nine combinations before reporting the
mean and SEM across five seeds (Figure~\ref{fig:temporal_response_gain}).
Timing and profile-shape measurements are provided as supplementary
diagnostics in Appendix~\ref{app:temporal_results}.

\subsection{Target Responses across Patterns and Delays}

\begin{figure}[t]
    \centering
    \includegraphics[width=0.8\linewidth]{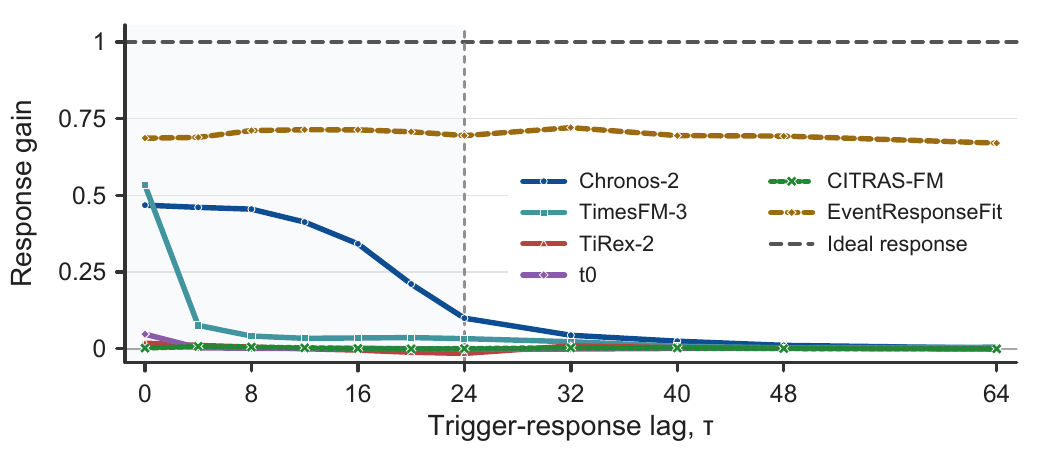}
    \caption{Response gain, averaged equally over nine
    event--response combinations (mean $\pm$ SEM across five seeds).
    Black dashed lines mark unit gain and the trigger at the forecast
    origin ($\tau=24$).}
    \label{fig:temporal_response_gain}
\end{figure}

\paragraph{Baseline Responses across Event--Response Patterns.}
At $\tau=0$, Chronos-2 and TimesFM-3 produce positive projection
gains across all nine combinations, indicating a response
component aligned with the expected target pattern.
Their responses are attenuated, and their strength depends
on the event--response mapping.
Both models respond more weakly to biphasic triggers across
the target-response families, while producing pronounced
biphasic target responses when the trigger is a spike or dip.
Chronos-2 and TimesFM-3 respond preferentially
to event triggers, with consistently weaker absolute
projection gains for sign-reversed distractors
(Appendix~\ref{app:temporal_results}).
{TiRex-2, $t_0$, and CITRAS-FM exhibit small mean projection gains
across the tested lags, including at zero lag
(Figure~\ref{fig:temporal_response_gain}).}

\paragraph{Increasing Delay within the Known-Future Segment.}
TimesFM-3's gain drops sharply at short nonzero lags in all
nine combinations.
This attenuation also recurs across the tested response anchors
in the spike-to-positive control
(Appendix~\ref{app:temporal_position_controls}).
Chronos-2 largely maintains its average gain over the shortest
lags, but its gain decreases as the lag grows further.

\paragraph{Increasing Delay within Observed History.}
At larger lags within the historical segment, the responses of
Chronos-2 and TimesFM-3 become weak despite the continued
availability of predictive information.
A simple event-response regression fitted to each window's
observed target and covariate histories
(EventResponseFit; Appendix~\ref{app:temporal_fitted_baseline})
retains substantial gains across lags.
Responsiveness at short lags therefore does not ensure that a
model preserves its response to the same predictive relationship
at larger lags.
Combination-specific results and controls are reported in
Appendix~\ref{app:temporal_results}. 
{Related fine-tuning experiments are reported in
Appendix~\ref{app:temporal_finetuning}.
Context-length sensitivity results are reported in
Appendix~\ref{app:bc_temporal_context}.}

\section{Structural Dependency Failures}
\label{sec:distributed_underutilization}
\label{sec:diag_distributed}

To investigate \textbf{RQ2}, we examine how TSFMs use predictive information
under different cross-channel dependence structures. We study a controlled single-factor setting inspired by dynamic
factor models, which describe multiple series through common
dynamic factors and series-specific components
\citep{stock2016dynamic}.
Factor-based forecasting uses these observations to estimate
predictive factors \citep{stock2002forecasting}.
We first provide the underlying state directly and then replace
it with noisy observations.

\subsection{Direct and Noisy Observations of a Shared Latent State}

We generate a shared latent process and a target series as
\begin{align}
    z_t &= \rho z_{t-1}+\sqrt{1-\rho^2}\,u_t, \\
    Y_t &= \kappa z_t+\sigma_y\epsilon_t,
\end{align}
where $u_t$ and $\epsilon_t$ are mutually independent standard
Gaussian white-noise processes.
We vary latent persistence $\rho$ while holding the target
loading and noise level fixed and preserving unit stationary
variance for $z_t$.
All comparisons use the same multivariate history interface
and matched target windows.
Full settings and numerical results are provided in
Appendices~\ref{app:evidence_scaling}
and~\ref{app:structural_numerical_results}.

\paragraph{{Context-Fitted References.}}
{As references for what is achievable from the same finite contexts,
we fit a Gaussian state-space model (SSM) and SparseDirect to each
window's observed histories at $\rho=0.95$.
The SSM's model class contains the generating process, but its
parameters, including the persistence, are estimated from each
context alone (Appendix~\ref{app:structural_fitted_baselines}).
SparseDirect is an exploratory sparse linear reference selected
by rolling validation within the observed context
(Appendix~\ref{app:structural_sparse_direct}). The SSM removes exact duplicates before fitting;
SparseDirect retains them.}

\paragraph{Providing the Latent State Directly.}
We compare target-only forecasting with forecasts that
additionally receive the true latent history $z_{1:T}$,
without supplying future latent values.
We measure the benefit as the {percentage} MAE reduction relative
to {each model's own} target-only baseline. This condition provides a reference for interpreting performance
when the state must instead be inferred from noisy observations.
\begin{figure}[t]
    \centering
    \includegraphics[width=\linewidth]{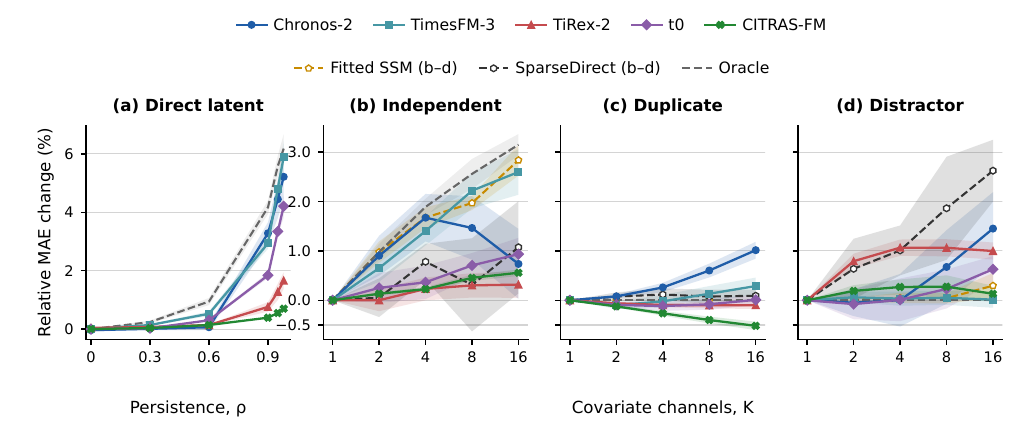}
    \caption{{Relative MAE reductions (\%) with (a) true latent history,
    relative to target-only input, and (b) independent observations,
    relative to noisy $K=1$; (c--d) relative MAE increases (\%)
    with duplicates and distractors, relative to noisy $K=1$.
    Each method uses its own reference MAE.
    Ratios are computed within each seed before aggregation.
    Panel (a) varies $\rho$; (b--d) use $\rho=0.95$ and
    $\lambda=0.75$.
    Curves show means, with shading for $\pm$ SEM over five seeds;
    dashed gray lines denote the oracle.
    Fitted SSM and SparseDirect are shown in (b--d).
    The fitted SSM removes exact duplicates before fitting;
    SparseDirect retains them.}}
    \label{fig:evidence_scaling}
\end{figure}

\paragraph{Adding Noisy Observation Channels of the Same Latent State.}
We replace direct latent access with auxiliary observations
\begin{equation}
    X_{j,t}=\lambda z_t+\sqrt{1-\lambda^2}\,\eta_{j,t},
\end{equation}
where the $\eta_j$ are mutually independent AR(1) processes,
independent of $z$ and the target noise, with the same persistence
and unit stationary variance as $z$.
The fixed loading $\lambda$ controls the contribution of the
shared state, while the normalization preserves unit variance
for each observation.
The \emph{independent} condition varies the number $K$ of
observations, each with its own noise realization.
This supplies complementary evidence about the same latent state.

\paragraph{Adding Duplicate or Irrelevant Channels.}
The \emph{duplicate} condition repeats one noisy observation,
while the \emph{distractor} condition retains one informative
channel and adds target-irrelevant series with matched
temporal statistics.
At each $K$, the three channel compositions have matched
channel counts and individual-channel marginal statistics.
Neither duplicates nor distractors add predictive information
beyond the single informative observation.
For all channel-composition comparisons, we {express} paired MAE
changes {as percentages of} each model's {own} noisy single-channel
baseline {MAE} ($K=1$). The oracle conditions on exactly the histories available in
each condition, providing a reference for the predictive value
of additional observations.
\subsection{Forecast Benefits and Channel Interference}

\paragraph{Benefits of Direct Latent Access.}
The predictive value of the latent history grows with persistence, and so do
the models' gains from direct access (Figure~\ref{fig:evidence_scaling}a).
At $\rho \geq 0.9$, Chronos-2 and TimesFM-3 obtain the {largest
relative gains among the TSFMs}, {while $t_0$ also benefits
substantially.} TiRex-2 {and CITRAS-FM obtain smaller gains}.
At intermediate persistence ($\rho = 0.6$), TimesFM-3 already obtains
part of the available gain, while Chronos-2's gain remains near zero
({Figure}~{\ref{fig:evidence_scaling}}a).
Because the oracle gain itself varies with $\rho$, model gains across
persistence levels must be interpreted alongside the corresponding
oracle gains.

\paragraph{Gains from Additional Noisy Observation Channels.}
In the persistent setting shown in Figure~\ref{fig:evidence_scaling}b,
TimesFM-3 continues to benefit from additional independent observations,
whereas Chronos-2's mean gain declines beyond an intermediate channel
count despite increasing oracle utility
(Appendix~\ref{app:paired_channel_count}).
{The fitted SSM} continues to improve as independent observations are
added and {attains gains close to} the oracle.
TiRex-2 gains little from additional noisy observations, and its
modest benefit from direct latent access indicates that this
limitation is already visible before multiple observations must
be combined.
{$t_0$ and CITRAS-FM also obtain smaller mean gains than TimesFM-3
from additional noisy observations.
SparseDirect also shows modest mean gains with substantial
variation across seeds.}

\paragraph{Interference from Duplicate and Irrelevant Channels.}
Distractor channels increase the error of TiRex-2 as soon as one
is added and that of Chronos-2 {and $t_0$} at larger distractor
counts, although {Chronos-2} varies considerably across seeds{.
CITRAS-FM changes comparatively little}
(Figure~\ref{fig:evidence_scaling}d){.
TimesFM-3 remains nearly unchanged while benefiting from
informative channels, showing that this interference can
be avoided zero-shot}.
The fitted state-space model, which is not told which channels
are irrelevant, incurs almost no distractor penalty.
{SparseDirect incurs a larger mean penalty than every tested TSFM
at} the {largest channel count}.
Exact duplicates add no information beyond the single informative
observation, yet Chronos-2's {mean} error increases steadily with their
number, while the {mean errors of TimesFM-3, TiRex-2, $t_0$,
and SparseDirect} change {comparatively} little
(Figure~\ref{fig:evidence_scaling}c).
{CITRAS-FM's mean error instead decreases as copies are added.
Accidental duplicates can therefore improve or worsen forecast accuracy.

Context-length sensitivity results} are reported in
Appendix~{\ref{app:bc_structural_context}}.

\section{When Known-Future Conditioning Misleads}
\label{sec:known_future_conditioning}
\label{sec:diag_known_future}

This section addresses \textbf{RQ3} and \textbf{RQ4}.
To study \textbf{RQ3}, we first examine how TSFM forecasts change
when cross-channel information is presented through different
conditioning interfaces, using settings where the supplied future
covariate adds no predictive information beyond the observed histories.
We then turn to \textbf{RQ4} by examining how forecast behavior changes
as future-covariate reliability deteriorates, first in controlled settings
and then in real-world tasks where future covariates are themselves
predicted.
Detailed numerical comparisons for the controlled settings are
provided in Appendices~\ref{app:known_future_numerical_results}
and~\ref{app:reliability_experiments}.

\subsection{When Future Covariates Have No Predictive Information}
\label{sec:irrelevant_future_covariates}

\paragraph{Lagged Copies of the Covariate and Conditioning Interfaces.}
We generate the target as
\begin{equation}
    Y_t=\beta X_{t-L}+\sqrt{1-\beta^2}\,\epsilon_t,
\end{equation}
where $X_t$ and $\epsilon_t$ are independent standard Gaussian
white-noise processes.
We vary the relation strength over $\beta\in\{0.4,0.6,0.8,1\}$.
At $\beta=1$, the target is an exact lagged copy of the covariate;
smaller positive values introduce target noise while preserving
unit marginal target variance.
With horizon $H=8$ and lags $L\in\{8,16\}$, all source values
needed to predict the target horizon lie in the observed history;
the future continuation adds no predictive information.
We compare \textsc{MV}, which supplies both histories as multivariate
channels; \textsc{PastCov}, which supplies the covariate history
as a past covariate; and \textsc{KC}, which additionally provides
its future continuation.
A channel-independent (\textsc{CI}) baseline uses only the target
history.
The protocols are detailed in Appendix~\ref{app:known_future_conditioning}.

\paragraph{{Context-Fitted References.}}
{At $\beta\in\{0.8,1\}$, we use lag-selected ordinary least
squares (OLS), which fits a single-lag covariate regression
(Appendix~\ref{app:lagged_copy_fitted_baselines}),
and SparseDirect, which fits a sparse linear regression for each
forecast step over channel--lag predictors.
Both are fitted and selected within each observed context,
without receiving the true lag.}

\paragraph{Forecasting an Exact Lagged Copy of the Covariate.}
The noise-free case tests whether models exploit a fully predictive
historical relationship without irreducible target uncertainty.
Chronos-2 benefits strongly from the covariate history under
\textsc{MV}/\textsc{PastCov}, while TimesFM-3{, TiRex-2, $t_0$,}
and {CITRAS-FM} remain close to their target-only baselines
(Table~\ref{tab:irrelevant_future_main}).
{SparseDirect predicts the lagged copy almost exactly from
history alone at both lags.}
\textsc{KC} improves {TimesFM-3 and TiRex-2} despite adding no
predictive information, whereas its benefit for Chronos-2 is
not consistent across lags{.
For $t_0$ and CITRAS-FM, \textsc{KC} instead increases mean MAE
relative to \textsc{PastCov}}.
We measure whether changing a predictive historical value
produces the expected change in the forecast.
Let $J_{h,j}$ denote the average forecast change at step $h$
per unit perturbation of the historical source value for
target step $j$.
Ideally, only step $j$ responds, with sensitivity $\beta$,
giving the response matrix $J^\star=\beta I_H$.
We measure the mismatch as
\begin{equation}
    E_J
    = \frac{\|J-\beta I_H\|_F}{\|\beta I_H\|_F}.
\end{equation}
This error captures both incorrect response strength and
changes at forecast steps that should remain unaffected.
It is zero when the average response matches the ideal
pattern; lower values indicate closer agreement
(Appendix~\ref{app:known_future_conditioning}).

\begin{figure}[t]
    \centering
    \includegraphics[width=\linewidth]{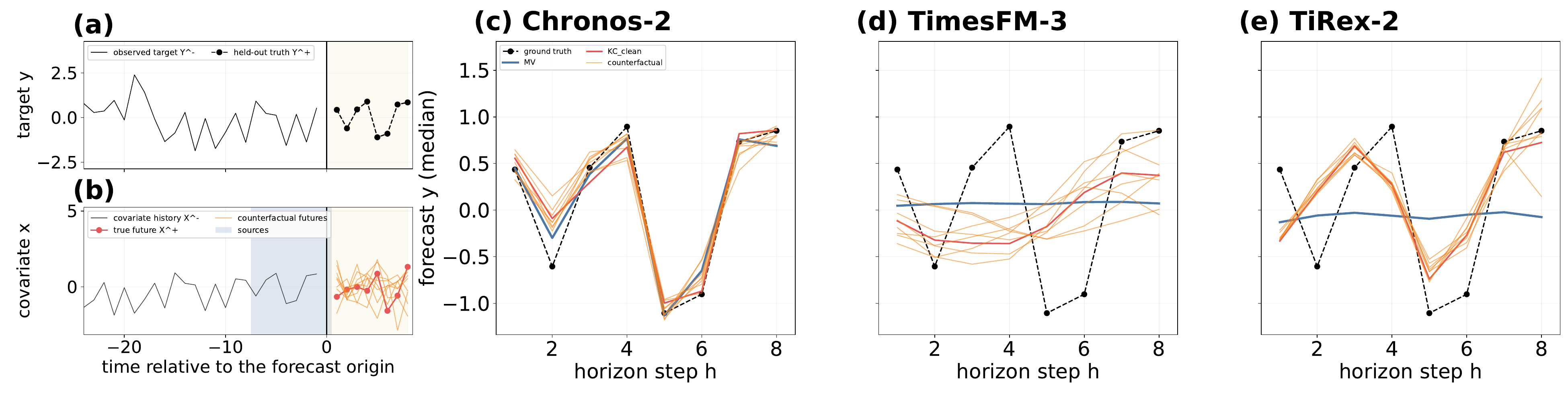}
    \caption{
    A single-window exact-copy example ($\beta=1$, $L=8$).
    (a--b) Target and covariate series; blue shading marks the historical source values for the forecast horizon.
    (c--e) Median forecasts under MV, KC with the original future, and KC with eight resampled futures, holding the histories fixed.
    }
    \label{fig:known_future_beta1_example}
\end{figure}

For TimesFM-3 and TiRex-2, \textsc{KC} lowers both MAE and
$E_J$ relative to \textsc{PastCov}, improving forecast accuracy
alongside the response to predictive historical values.
Thus, historical information can be used differently across
input and conditioning configurations even when the available
predictive information is unchanged.
Figure~\ref{fig:known_future_beta1_example} illustrates these
differences for a single window.
\paragraph{Interface Dependence across Relation Strengths.}
{The lag-selected OLS reference} identifies the true lag
in every tested window for $\beta\in\{0.8,1\}$ and $L\in\{8,16\}$.
It attains near-oracle accuracy and produces identical forecasts
with or without the future continuation
(Appendix~\ref{app:lagged_copy_fitted_baselines}).
{SparseDirect also attains near-oracle accuracy from history
alone at $\beta=0.8$. Its history-only and \textsc{KC} routes yield similar mean MAEs,
but not identical forecasts (Table~\ref{tab:irrelevant_future_main}).}
TSFM accuracy varies with the conditioning interface.
Across the noisy-copy settings, \textsc{KC} yields lower mean MAE
than \textsc{PastCov} for {Chronos-2, TimesFM-3, and TiRex-2} at both
lags, although some gains are small relative to seed-level variation{.
For $t_0$ and CITRAS-FM, mean MAE is higher under \textsc{KC}
than under \textsc{PastCov} at every tested positive relation strength}.
For TimesFM-3 and TiRex-2, mean gains increase as the relationship
strengthens, and \textsc{KC} also lowers $E_J$ relative to
\textsc{PastCov}
(Appendix~\ref{app:known_future_beta_sweep}).
These improvements occur without additional predictive
information from future covariates and are accompanied by
responses to predictive historical values that more closely
match the oracle.

\paragraph{Sensitivity to Irrelevant Future Covariate Values.}
We hold the histories and \textsc{KC} interface fixed and resample
the future continuation.
Forecast change under future resampling, $S_{\mathrm{future}}$,
is the mean pairwise absolute difference between the resulting
median forecasts.
The oracle is invariant to this resampling, as is the fitted
regression in the tested $\beta\in\{0.8,1\}$,
$L\in\{8,16\}$ settings. {SparseDirect is also invariant at $\beta=1$, but responds to future resampling at $\beta=0.8$.}
All {five} TSFMs remain sensitive at every tested positive
relation strength.
At $L=8$, {the mean sensitivities of Chronos-2, TimesFM-3, and
TiRex-2 exceed those of} their no-dependence {controls}
({Appendix}~{\ref{app:known_future_beta_sweep}}).
For TimesFM-3 and TiRex-2, mean sensitivity increases as the
historical relationship strengthens{.
At $\beta=0.8$, Chronos-2, TimesFM-3, TiRex-2, and $t_0$
have larger mean sensitivity than SparseDirect, while CITRAS-FM
has smaller mean sensitivity (Table~\ref{tab:irrelevant_future_main})}.
Known-future conditioning can therefore improve accuracy
while making forecasts sensitive to inputs that leave the
oracle predictive distribution unchanged.
{Context-length sensitivity results are reported in
Appendix~\ref{app:kc_context_sensitivity}.}

\subsection{Changes in Covariate Reliability at Forecast Time}
\label{sec:kc_reliability_shift}

\paragraph{Predictive Covariates and Reliability Degradation.}
Future covariates supplied through a known-future interface may
themselves be forecasts.
For example, electricity-price tasks in fev-bench include day-ahead
load and generation forecasts as known covariates
\citep{shchur2026fevbenchrealisticbenchmarktime,Lago_2021}.
Availability at forecast time therefore does not imply error-free
values.
We study a controlled setting in which the clean covariate and
target share a persistent latent state, making clean future
covariates predictive of the target.
Holding the target process fixed, we degrade the supplied
covariate by mixing it with an independent realization having
matched temporal statistics.
The mixing weight $\lambda$ controls reliability, from the clean
covariate at $\lambda=1$ to an independent series at $\lambda=0$.

\paragraph{Degrading Covariate History, Future Values, or Both.}
\textsc{Stationary} degrades both history and future,
\textsc{Future-Only} degrades only the future, and
\textsc{History-Only} degrades only the history.
\textsc{Stationary} and \textsc{Future-Only} receive exactly
the same degraded future values, differing only in whether
the historical covariate has also been degraded.
Their comparison measures how the effect of those future values
depends on the preceding history.
\textsc{History-Only} evaluates the reverse mismatch.
The generation process and degradation conditions are specified
in Appendix~\ref{app:reliability_shift}.

\paragraph{{Fitted Conditional References.}}
{We also evaluate a fitted joint Gaussian state-space model (SSM)
and a lag-selected vector autoregression (VAR).
Fitting and lag selection use only the observed histories;
forecasts condition on the supplied future covariate values.}

\paragraph{Forecast Accuracy and Coverage under Reliability Mismatch.}
Under \textsc{Stationary} degradation, forecast error approaches
the channel-independent baseline as covariate reliability decreases.
{For most models, the} same future values produce larger errors
when they follow a clean history, while the reverse mismatch is less harmful
(Table~\ref{tab:kc_reliability_shift}).
{The fitted SSM and VAR also exhibit this mismatch.
At $\lambda=0$, the mean penalties for Chronos-2, TimesFM-3,
and TiRex-2 are close to the SSM's and exceed the VAR's;
$t_0$ is comparable to the VAR, while CITRAS-FM is less affected.}
At matched degraded future values, \textsc{Future-Only} also
produces narrower prediction intervals and lower empirical
coverage than \textsc{Stationary} {for Chronos-2, TimesFM-3,
TiRex-2, and $t_0$} (Table~{\ref{tab:ade_reliability_intervals_tsfm}}).
A relationship supported by history can therefore become a
source of forecasting risk when future input reliability changes.
This comparison characterizes risk under an unannounced change
without assuming that the change is identifiable from the
available inputs.
{Context-length sensitivity results are reported in
Appendix~\ref{app:ade_reliability_context}.}

\begin{wrapfigure}{r}{0.40\textwidth}
    \centering
    \includegraphics[width=\linewidth]
        {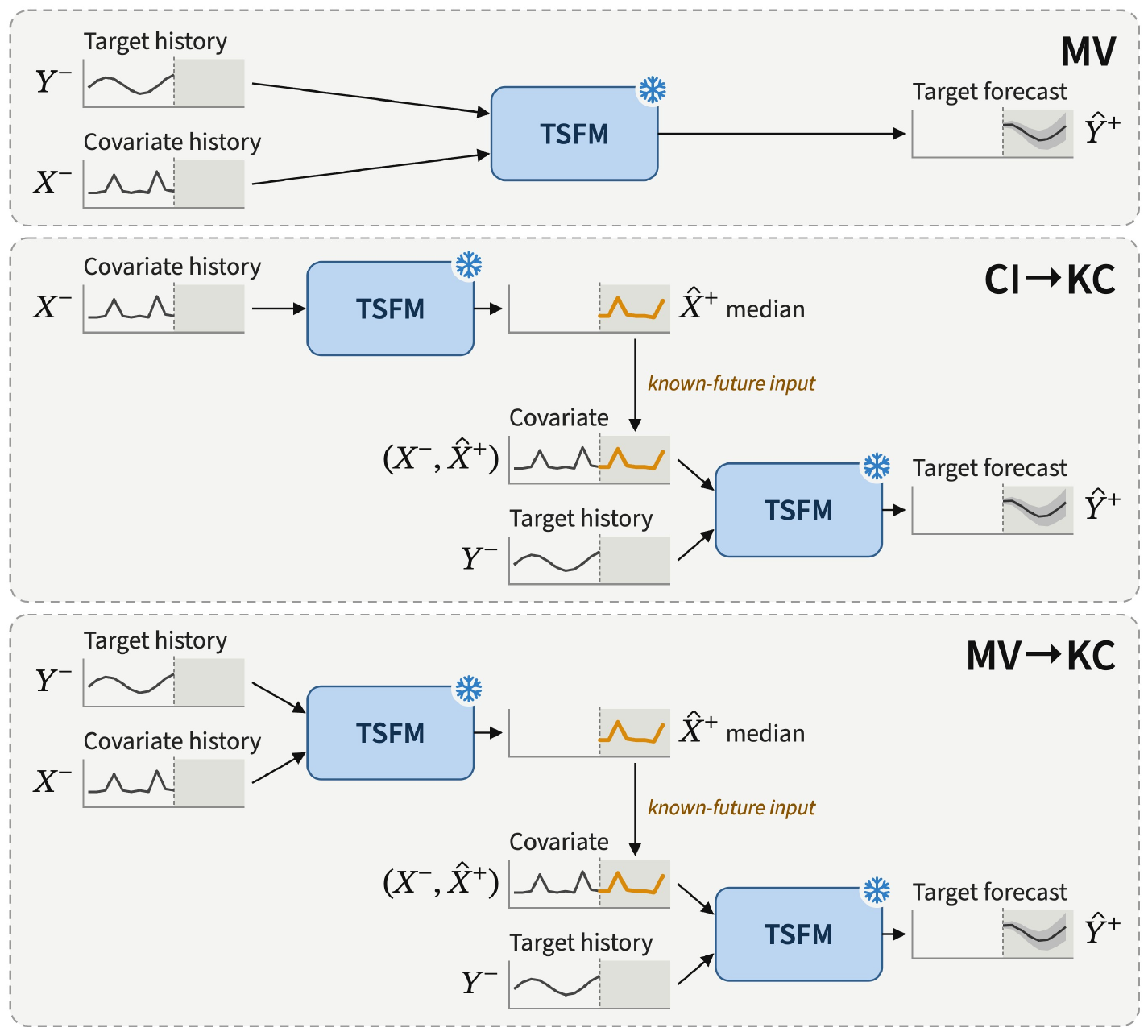}
    \caption{Direct forecasting and median-path reuse.
    Orange paths are predicted covariates.}
    \label{fig:forecast_reuse_protocol}
\end{wrapfigure}

\subsection{Using the Model's Own Covariate Forecasts as Known-Future Inputs}
\label{sec:forecast_reuse}

\paragraph{Real-world Tasks with Unobserved Future Covariates.}
Forecasting pipelines may estimate future covariate values
and supply these estimates to a target forecaster
\citep{hyndman2021fpp3}.
The reliability of these future inputs may differ from that
of the observed histories.
Related approaches in load forecasting pass temperature
scenarios through a demand model \citep{dimoulkas2019neural}.
We examine this practice on three paired-channel tasks from
fev-bench \citep{shchur2026fevbenchrealisticbenchmarktime}:
air temperature $\to$ electricity load (GFC17), wind speed
$\to$ active power (KDD), and customer count $\to$ sales (Rossmann).
We retain the benchmark horizons and rolling forecast origins,
restrict inputs to the selected source $X$ and target $Y$,
and cap the context at 512 observations.
All routes in the main comparison withhold realized future
source values {and are evaluated on the same cases with complete future covariates}.
Dataset selection and task adaptations are detailed in
Appendix~\ref{app:forecast_reuse_data}.
\begin{table}[t]
    \centering
    \caption{
    Known-future conditioning for exact ($\beta=1$) and noisy
    ($\beta=0.8$) lagged copies ($H=8$).
    CI uses only the target history; MV and PastCov are reported separately.
    $S_{\mathrm{future}}$ is mean $\pm$ SEM across five seeds.
    Dashes denote conditions or metrics not evaluated.
    }
    \label{tab:irrelevant_future_main}
    \resizebox{\linewidth}{!}{%
        \scriptsize
\setlength{\tabcolsep}{2pt}
\renewcommand{\arraystretch}{1.0}
\begin{tabular}{@{}lc*{6}{r}c@{\hspace{6pt}}*{6}{r}c@{}}
    \toprule
    \multirow{3}{*}{Model} & \multirow{3}{*}{$L$}
    & \multicolumn{7}{c}{Exact lagged copy ($\beta=1$)}
    & \multicolumn{7}{c}{Noisy lagged copy ($\beta=0.8$)} \\
    \cmidrule(lr){3-9}\cmidrule(lr){10-16}
    & & \multicolumn{4}{c}{MAE $\downarrow$}
    & \multicolumn{2}{c}{$E_J$ $\downarrow$}
    & \multirow{2}{*}{$S_{\mathrm{future}}$ $\downarrow$}
    & \multicolumn{4}{c}{MAE $\downarrow$}
    & \multicolumn{2}{c}{$E_J$ $\downarrow$}
    & \multirow{2}{*}{$S_{\mathrm{future}}$ $\downarrow$} \\
    \cmidrule(lr){3-6}\cmidrule(lr){7-8}
    \cmidrule(lr){10-13}\cmidrule(lr){14-15}
    & & CI & MV & PastCov & KC & PastCov & KC &
    & CI & MV & PastCov & KC & PastCov & KC & \\
    \midrule
    \multirow{2}{*}{Chronos-2}
    & 8  & 0.804 & 0.171 & 0.171 & 0.185 & 0.084 & 0.092
    & $0.148 \pm 0.001$
    & 0.801 & 0.540 & 0.540 & 0.533 & 0.144 & 0.062
    & $0.162 \pm 0.001$ \\
    & 16 & 0.801 & 0.323 & 0.323 & 0.306 & 0.256 & 0.222
    & $0.146 \pm 0.000$
    & 0.803 & 0.615 & 0.615 & 0.587 & 0.390 & 0.247
    & $0.154 \pm 0.000$ \\
    \addlinespace[1pt]
    \multirow{2}{*}{TimesFM-3}
    & 8  & 0.804 & 0.790 & 0.790 & 0.702 & 0.981 & 0.855
    & $0.164 \pm 0.001$
    & 0.801 & 0.796 & 0.796 & 0.755 & 0.991 & 0.890
    & $0.114 \pm 0.001$ \\
    & 16 & 0.801 & 0.790 & 0.790 & 0.738 & 0.985 & 0.894
    & $0.085 \pm 0.001$
    & 0.802 & 0.799 & 0.799 & 0.779 & 0.994 & 0.939
    & $0.053 \pm 0.001$ \\
    \addlinespace[1pt]
    \multirow{2}{*}{TiRex-2}
    & 8  & 0.804 & 0.801 & 0.801 & 0.551 & 0.997 & 0.666
    & $0.163 \pm 0.001$
    & 0.801 & 0.799 & 0.799 & 0.668 & 0.997 & 0.673
    & $0.128 \pm 0.001$ \\
    & 16 & 0.801 & 0.800 & 0.800 & 0.642 & 0.999 & 0.769
    & $0.075 \pm 0.001$
    & 0.803 & 0.802 & 0.802 & 0.730 & 0.999 & 0.801
    & $0.061 \pm 0.001$ \\
    \addlinespace[1pt]
    \multirow{2}{*}{$t_0$}
    & 8  & 0.805 & 0.804 & 0.805 & 0.808 & 0.999 & 0.999
    & $0.078 \pm 0.001$
    & 0.803 & 0.802 & 0.803 & 0.805 & 0.999 & 0.998
    & $0.074 \pm 0.001$ \\
    & 16 & 0.802 & 0.802 & 0.803 & 0.806 & 1.000 & 1.000
    & $0.081 \pm 0.001$
    & 0.804 & 0.804 & 0.804 & 0.807 & 1.000 & 1.000
    & $0.077 \pm 0.001$ \\
    \addlinespace[1pt]
    \multirow{2}{*}{CITRAS-FM}
    & 8  & 0.807 & 0.802 & 0.799 & 0.803 & 0.991 & 0.996
    & $0.029 \pm 0.001$
    & 0.804 & 0.801 & 0.798 & 0.802 & 0.992 & 0.997
    & $0.020 \pm 0.001$ \\
    & 16 & 0.803 & 0.799 & 0.799 & 0.802 & 0.994 & 1.000
    & $0.007 \pm 0.000$
    & 0.805 & 0.803 & 0.803 & 0.805 & 0.995 & 1.000
    & $0.007 \pm 0.000$ \\
    \addlinespace[1pt]
    \multirow{2}{*}{SparseDirect}
    & 8  & --- & 0.001 & --- & 0.001 & --- & ---
    & $0.000 \pm 0.000$
    & --- & 0.493 & --- & 0.493 & --- & ---
    & $0.030 \pm 0.001$ \\
    & 16 & --- & 0.001 & --- & 0.001 & --- & ---
    & $0.000 \pm 0.000$
    & --- & 0.491 & --- & 0.490 & --- & ---
    & $0.021 \pm 0.001$ \\
    \bottomrule
\end{tabular}%
    }
\end{table}

\begin{table}[t]
    \centering
    \caption{
    Forecast-time covariate reliability mismatch at $\lambda=0$ ($H=8$).
    Stationary and Future-only share identical degraded future inputs.
    CI uses only the target history; Clean denotes $\lambda=1$.
    $\Delta_{\mathrm{shift}}$ is Future-only minus Stationary MAE.
    {$\Delta_{\mathrm{VAR}} =
    \Delta_{\mathrm{shift}} - \Delta_{\mathrm{shift}}^{\mathrm{VAR}}$
    measures the difference from the VAR's shift penalty.}
    Uncertainties are SEMs of five paired seed-level differences.
    }
    \label{tab:kc_reliability_shift}
    \resizebox{\linewidth}{!}{%
            \small
    \setlength{\tabcolsep}{3pt}
    \renewcommand{\arraystretch}{1.10}
    \begin{tabular}{@{}lrrrrrcc@{}}
        \toprule
        & \multicolumn{5}{c}{MAE $\downarrow$} & & \\
        \cmidrule(lr){2-6}
        Model & CI & Clean & Stationary
        & \shortstack{Future} & \shortstack{History}
        & $\Delta_{\mathrm{shift}}$ & {$\Delta_{\mathrm{VAR}}$} \\
        \midrule
        Chronos-2 & 0.802 & 0.589 & 0.800 & 1.259 & 0.808 & $0.459 \pm 0.020$ & $+0.083 \pm 0.013$ \\
        TimesFM-3 & 0.797 & 0.580 & 0.796 & 1.271 & 0.792 & $0.475 \pm 0.018$ & $+0.099 \pm 0.009$ \\
        TiRex-2 & 0.796 & 0.585 & 0.797 & 1.232 & 0.798 & $0.435 \pm 0.016$ & $+0.058 \pm 0.009$ \\
        $t_0$ & 0.808 & 0.599 & 0.803 & 1.151 & 0.820 & $0.348 \pm 0.021$ & $-0.029 \pm 0.016$ \\
        CITRAS-FM & 0.802 & 0.694 & 0.801 & 0.865 & 0.801 & $0.065 \pm 0.009$ & $-0.312 \pm 0.019$ \\
        \midrule
        Fitted joint SSM & 0.791 & 0.582 & 0.794 & 1.238 & 0.797 & $0.444 \pm 0.015$ & $+0.067 \pm 0.008$ \\
        Lag-selected VAR & 0.824 & 0.609 & 0.829 & 1.206 & 0.830 & $0.377 \pm 0.020$ & --- \\
        \bottomrule
    \end{tabular}
    }
\end{table}

\paragraph{Direct Target Forecasting versus Conditioning on Predicted Covariates.}
{\leavevmode}
Direct \textsc{MV} forecasting predicts both channels from their
observed histories (Figure~\ref{fig:forecast_reuse_protocol}).
In median-path \textsc{MV}$\to$\textsc{KC}, we supply the
covariate median from this forecast as a known-future input and
predict the target again with the same frozen model and unchanged
histories, so the second stage receives no new observations.
We also use {trajectory} propagation to examine whether any loss
of probabilistic quality persists beyond single-path conditioning.
We {draw 100 covariate trajectories from the \textsc{MV} forecast
marginals using a Gaussian copula
\citep{baron2025efficientlygeneratingcorrelatedsample, choi2026interweavingmarginalsmultivariatesample}.
Its AR(1)-style temporal correlation is estimated from the
observed covariate history.
We} condition separately on each {trajectory} and obtain target
quantiles by inverting the equally weighted mixture of the
resulting target CDFs.
This {trajectory construction} and its sensitivity checks are
detailed in Appendix~\ref{app:quantile_path_propagation}.

\paragraph{Quantile Loss and Interval Coverage with Predicted Covariates.}
We evaluate scaled quantile loss (SQL) and empirical coverage
of the nominal $80\%$ prediction interval.
SQL normalizes quantile loss by the seasonal naive error of
the target context; scoring and aggregation details are given
in Appendix~\ref{app:forecast_reuse_data}.
Conditioning on predicted covariate medians increases SQL and
lowers coverage relative to direct \textsc{MV} across all evaluated
dataset--model
combinations (Table~\ref{tab:forecast_reuse_sql_coverage}).
These losses in probabilistic quality can occur even when
point forecasts improve: on KDD, median-path {\textsc{MV}$\to$\textsc{KC}
reduces} target MAE for every model while producing narrower
intervals with lower coverage.
Using the same model in both stages therefore does not by
itself preserve probabilistic forecast quality. {The fitted joint SSM and VAR also show this deterioration.
SQL increases and coverage drops are generally larger for the
TSFMs than for the VAR (Appendix~\ref{app:forecast_reuse_results}).}

Compared with median-path conditioning, {trajectory} propagation
reduces SQL and moves coverage closer to the nominal level
in every evaluated dataset--model combination.
Nevertheless, coverage {generally} remains below both the nominal
level and that of direct \textsc{MV} forecasting.
{For} the {SSM and VAR, propagating their fitted joint trajectories
also brings SQL close to direct \textsc{MV}}
(Appendix~{\ref{app:forecast_reuse_results}}).
As a diagnostic, this comparison shows that accounting for
covariate uncertainty recovers a substantial portion of the
probabilistic quality lost under single-path conditioning,
{with remaining differences from} direct multivariate
forecasting {varying across models.
Source-only recalibration results are reported in
Appendix~\ref{app:forecast_reuse_experiments}.
Context-length sensitivity results are reported in
Appendix~\ref{app:f_context_sensitivity}}.
\begin{table}[t]
\centering
\caption{Conditioning on predicted covariates with a median path (Median)
or propagation of 100 sampled covariate trajectories (Trajectory),
compared with direct \textsc{MV}.
Bold columns show the Median-to-Trajectory change: relative SQL change
(\%) and coverage change in percentage points (pp), computed
from unrounded scores.}
\label{tab:forecast_reuse_sql_coverage}
\resizebox{\linewidth}{!}{%
\begingroup
\small
\setlength{\tabcolsep}{3pt}
\renewcommand{\arraystretch}{1.12}
\begin{tabular}{@{}llrrrrrrrr@{}}
\toprule
& & \multicolumn{4}{c}{SQL $\downarrow$}
  & \multicolumn{4}{c}{Coverage (\%; nominal $80\%$)} \\
\cmidrule(lr){3-6}\cmidrule(lr){7-10}
Dataset & Model
& \textsc{MV} & Median & Trajectory & $\Delta$ (\%)
& \textsc{MV} & Median & Trajectory & $\Delta$ (pp) \\
\midrule
GFC17 & Chronos-2 & 0.968 & 1.062 & 1.032 & $\mathbf{-2.9}$ & 69.4 & 46.2 & 65.7 & $\mathbf{+19.5}$ \\
 & TimesFM-3 & 0.954 & 0.996 & 0.953 & $\mathbf{-4.3}$ & 71.2 & 53.2 & 66.5 & $\mathbf{+13.3}$ \\
 & TiRex-2 & 0.930 & 0.942 & 0.930 & $\mathbf{-1.3}$ & 71.4 & 60.9 & 66.0 & $\mathbf{+5.2}$ \\
 & $t_0$ & 0.957 & 0.983 & 0.974 & $\mathbf{-0.9}$ & 70.3 & 59.3 & 65.1 & $\mathbf{+5.8}$ \\
 & CITRAS-FM & 0.990 & 1.021 & 0.989 & $\mathbf{-3.1}$ & 46.2 & 39.0 & 43.0 & $\mathbf{+4.0}$ \\
\midrule
KDD & Chronos-2 & 0.828 & 0.926 & 0.888 & $\mathbf{-4.1}$ & 71.9 & 46.1 & 65.2 & $\mathbf{+19.1}$ \\
 & TimesFM-3 & 0.807 & 0.931 & 0.799 & $\mathbf{-14.2}$ & 71.8 & 25.5 & 72.5 & $\mathbf{+47.0}$ \\
 & TiRex-2 & 0.735 & 0.839 & 0.728 & $\mathbf{-13.2}$ & 84.1 & 30.5 & 79.8 & $\mathbf{+49.3}$ \\
 & $t_0$ & 0.732 & 0.808 & 0.720 & $\mathbf{-10.9}$ & 79.9 & 30.2 & 68.5 & $\mathbf{+38.4}$ \\
 & CITRAS-FM & 0.802 & 0.849 & 0.798 & $\mathbf{-6.1}$ & 53.0 & 34.4 & 43.5 & $\mathbf{+9.1}$ \\
\midrule
Rossmann & Chronos-2 & 0.459 & 0.507 & 0.469 & $\mathbf{-7.5}$ & 76.5 & 37.4 & 73.9 & $\mathbf{+36.5}$ \\
 & TimesFM-3 & 0.490 & 0.545 & 0.497 & $\mathbf{-8.8}$ & 83.0 & 36.9 & 78.2 & $\mathbf{+41.3}$ \\
 & TiRex-2 & 0.520 & 0.562 & 0.524 & $\mathbf{-6.8}$ & 81.8 & 47.9 & 77.8 & $\mathbf{+29.9}$ \\
 & $t_0$ & 0.475 & 0.539 & 0.485 & $\mathbf{-10.0}$ & 82.4 & 36.6 & 80.3 & $\mathbf{+43.7}$ \\
 & CITRAS-FM & 0.527 & 0.538 & 0.534 & $\mathbf{-0.7}$ & 66.5 & 56.5 & 57.1 & $\mathbf{+0.6}$ \\
\bottomrule
\end{tabular}
\endgroup
}
\end{table}

\section{Conclusion}

Across the four research questions, we find that cross-channel
information can be available without being effectively or reliably
used; statistical baselines fitted to the same observed contexts
recover several relationships that the models underuse.
Depending on the model, TSFMs underuse delayed signals or
complementary observations while remaining sensitive to redundant
or conditionally irrelevant inputs.
The use of historical information can change across conditioning
interfaces even without additional predictive information,
and forecast-time changes in covariate reliability can compromise
both accuracy and probabilistic forecast quality.
{On real-world tasks, conditioning on a single predicted covariate
path degrades probabilistic forecasts. Trajectory propagation
recovers a substantial portion of this loss, while coverage
generally remains below the nominal level.}

These findings motivate evaluating not only whether cross-channel
inputs improve forecasts, but also what predictive information they
contribute beyond the observations already available.
Promising directions include connecting these behavioral findings to architectural and pretraining choices, and developing training and inference procedures that selectively exploit cross-channel information while accounting for uncertainty in predicted covariates.

\newpage

\subsection*{AI use statement}
We used generative AI tools to assist with experimental design,
code development, experiment execution, interpretation of results,
literature search, and the preparation of manuscript text,
figures, and tables.
Generative AI was not used to generate the initial research idea.
The authors reviewed intermediate experimental results at multiple
stages of the study and take responsibility for the final content,
including all AI-assisted text, code, analyses, and other artifacts.

\subsection*{Ethics statement}
This study uses controlled synthetic data and existing forecasting
benchmarks to examine the reliability of cross-channel forecasting.
A potential downstream risk is overconfidence when predicted
covariates are supplied as known-future inputs without accounting
for their uncertainty.
Our analysis documents this risk and shows that propagating covariate uncertainty recovers a substantial portion of the lost probabilistic quality but does not fully close the gap, motivating careful evaluation of probabilistic forecasts in downstream applications.

\subsection*{Reproducibility statement}
We document the model versions, experimental settings, random seeds,
and evaluation protocols in
Appendix~\ref{app:general_protocol} and the experiment-specific
appendices.
The synthetic data-generating processes, oracle references,
controls, baseline-fitting procedures, and detailed numerical
results are provided in
Appendices~\ref{app:temporal_experiments}--\ref{app:reliability_experiments}.
Data preparation and evaluation protocols for the real-data
experiments are described in
Appendix~\ref{app:forecast_reuse_data}, and representation-probe
procedures are detailed in
Appendix~\ref{app:representation_experiments}.



\newpage
\bibliography{iclr2027_conference}
\bibliographystyle{iclr2027_conference}
\newpage

\appendix

\section{Experimental Protocol and Reporting}
\label{app:general_protocol}

The appendices follow the behavioral experiments in the main text:
temporal event--response diagnostics (Appendix~\ref{app:temporal_experiments}),
cross-channel dependency structure (Appendix~\ref{app:structural_experiments}),
conditionally irrelevant future covariates (Appendix~\ref{app:known_future_experiments}),
reliability shifts (Appendix~\ref{app:reliability_experiments}), and
conditioning on predicted covariates in real data (Appendix~\ref{app:forecast_reuse_experiments}).
The separate representation experiments are collected in
Appendix~\ref{app:representation_experiments}.

\subsection{Models and Input Conditions}

We evaluate frozen checkpoints of Chronos-2, TimesFM-3, TiRex-2,
CITRAS-FM, and $t_0$, identified as
\texttt{amazon/chronos-2},
\texttt{google/timesfm-3.0-pytorch},
\texttt{NX-AI/TiRex-2},
\texttt{hitachi-nlp/citras-fm}, and
\texttt{theforecastingcompany/t0-beta}, respectively.
Backbones remain frozen in the behavioral evaluations.
Task-specific fine-tuning is evaluated separately in
Appendix~\ref{app:temporal_finetuning}.
Representation probes are fitted separately, using the splits
described in Appendix~\ref{app:representation_experiments}.

\textsc{CI} uses only the scored target's history; \textsc{MV} supplies
the histories of the target and other channels as a multivariate input.
\textsc{PastCov} supplies the covariate history through the past-covariate
interface, and \textsc{KC} additionally supplies future covariate values.
All five models support \textsc{CI}, \textsc{MV}, and \textsc{KC}, and
use attention mechanisms to exchange information across channels.
Their temporal backbones, cross-channel attention mechanisms, and
treatment of known-future covariates differ, as summarized below.
These condition names describe inputs rather than assuming that every
backbone implements the interfaces with the same internal computation.
The information supplied by each condition is specified within its
experiment. Point-forecast measurements use the native median forecast,
or the available quantile closest to 0.5 as specified for the temporal task.

\paragraph{Chronos-2.}
Chronos-2~\citep{ansari2025chronos2univariateuniversalforecasting} is a patch-based, encoder-only
Transformer that alternates temporal attention within each series with
group attention across related series at corresponding patch positions.
Observed future covariate values are supplied alongside masked future
target values. Pretraining combines real-world and synthetic univariate
data; multivariate and covariate-informed training tasks are generated
synthetically by imposing contemporaneous and sequential dependencies
among series.

\paragraph{TimesFM-3.}
TimesFM-3~\citep{jain2026timesfm3} uses a patch-based, decoder-only
Transformer with alternating causal temporal attention and full attention
across variates. Its lookahead token construction incorporates upcoming
known-covariate patches, while masked future target tokens enable
forecasting the horizon in a single forward pass.
Pretraining uses a mixture of real-world and synthetic time series.

\paragraph{TiRex-2.}
TiRex-2~\citep{podest2026tirex2generalizingtirexmultivariate} combines recurrent xLSTM temporal
processing with asymmetric grouped attention across variates.
Known-future covariates additionally receive reverse temporal processing,
while the attention mask preserves causality for target variables.
Pretraining uses real-world and synthetic univariate corpora together
with an on-the-fly coupling pipeline that constructs multivariate
dependencies, including shared-factor and lagged relationships.

\paragraph{CITRAS-FM.}
CITRAS-FM~\citep{yamaguchi2026citrasfmtinytimeseries} is a compact, patch-based,
decoder-only Transformer with separate temporal and cross-variate
attention modules. Its Shifted Attention aligns target tokens with
known covariates in the next patch being predicted.
Pretraining combines augmented real-world data and synthetic multivariate
series, including CovSynth, which constructs covariates from decomposed
target-series components.

\paragraph{$t_0$.}
We use the $t_0$-beta checkpoint~\citep{meyer2026t0timeseriesfoundationmodel}, a decoder-style
patch Transformer that alternates attention along the temporal and
variate axes to incorporate target histories, past covariates, and
known-future covariates.
The model family is pretrained on curated public data and synthetic
generator families designed to contain covariate-to-target dependencies.
\subsection{Baselines Fitted to the Observed Context}
\label{app:context_fitted_protocol}

We distinguish three types of reference: an analytic oracle that knows
the data-generating process, a statistical model fitted to the available
observations under a specified model class, and a frozen pretrained TSFM.
The fitted baselines assess performance achievable from the finite
observed context with explicit modeling assumptions.
Each fit uses only one window's history; training examples are not pooled
across windows, conditions, or seeds.
Supplied future covariates may be used for prediction when allowed by
the corresponding input condition, but target values in the final
forecast horizon are never used for fitting or model selection.
Validation targets used for hyperparameter selection lie entirely
inside the observed context. Fitting objectives and selection criteria
are specified separately for each estimator below.

Generating parameters, true lags, event labels, kernels, and event
annotations are excluded from estimator inputs.
The known-lag OLS reference is an explicitly privileged exception:
it receives the true lag, while estimating its regression coefficients
from the observed history.
In the structural direct-access condition, the true latent history is
available because it is part of that condition's observed input.
All other structural conditions use only their specified target and covariate histories.
These distinctions separate a model-class assumption from knowledge
of the realized generating parameters.

The baselines use the same windows and five generation seeds as their
corresponding behavioral experiments. Lagged-copy and structural tasks
use a 512-step context, an eight-step horizon, and 256 windows per seed;
event-response and position tasks use a 768-step context, a 96-step
horizon, and 300 windows per seed.
TSFM forecasts are taken from the existing experiment records.
Comparisons are formed within matched windows and summarized within
each seed before reporting the mean and SEM across the five seeds.
For response profiles, the original equal-combination aggregation is
retained. Additional model fits, lags, anchors, and channel counts are
not treated as independent repetitions.

{The event-response, structural, and lagged-copy reference
evaluations report point forecasts, scored by MAE, event-response metrics,
and input sensitivity as appropriate. Probabilistic joint SSM and VAR
references for reliability mismatch and predicted-covariate conditioning
are reported separately in Appendices~\ref{app:reliability_experiments}
and~\ref{app:forecast_reuse_experiments}.}
Successful fitted baselines demonstrate achievable performance under
their stated assumptions, while unsuccessful fits do not establish
that predictive information is absent from the context.
Method details and results follow the corresponding behavioral
experiments in Appendices~\ref{app:temporal_fitted_baseline},
\ref{app:structural_fitted_baselines}, and~\ref{app:lagged_copy_fitted_baselines}.

\subsection{Evaluation Populations and Uncertainty}
Synthetic comparisons share realizations between their compared conditions.
{
For the temporal, structural, lagged-copy, and reliability-shift behavioral
experiments, uncertainty is computed from five generation-seed quantities.
Paired differences are formed within each seed before reporting their mean
and sample standard deviation divided by $\sqrt{5}$.
Excess penalties relative to a fitted reference are also paired within seed.
Windows and forecast steps are not treated as independent generation seeds.
}
Metric-specific definitions and aggregation order are given with each experiment.

All five-seed experiments, including temporal responses, use generation seeds $0$--$4$.
Probe uncertainties summarize held-out-seed folds whose training sets overlap;
the folds are not additional independently trained foundation models.
Real-data cohorts follow the dataset-specific rolling origins and
series-level aggregation in Appendix~\ref{app:forecast_reuse_data}, rather than
the synthetic generation-seed protocol.

\clearpage
\section{Temporal Event--Response Diagnostics}
\label{app:temporal_experiments}
\subsection{Temporal Cross-Channel Utilization Task}
\label{app:temporal_setup}

\paragraph{Data generation.}
For condition $c=(p,r)$, comprising trigger family $p$ and response family $r$, the covariate and target channels follow
\begin{align}
X^{(p)}_t
&= \sigma_x\epsilon^x_t
 + A\sum_k P_p(t-t_k)
 + A\sum_j P_{\mathrm{dis},p}(t-u_j), \\
Y^{(r)}_t
&= L_t+B\sum_k R_r(t-t_k-\tau)+\sigma_y\epsilon^y_t, \\
L_t&=L_{t-1}+\sigma_{\mathrm{level}}\epsilon^L_t,
\end{align}
where the noise sequences are independent standard Gaussian sequences, $t_k$ are trigger anchors, and $u_j$ are distractor anchors.
The target level is initialized at zero immediately before the generated context.
The covariate background is white noise, and the target background is an independent random walk.
The amplitudes are $A=B=3$, with noise scales $\sigma_x=0.3$, $\sigma_y=0.1$, and $\sigma_{\mathrm{level}}=0.15$.

\paragraph{Trigger and response families.}
The width-four trigger kernels are
\begin{align}
P_{\mathrm{spike}}&=[0,1,1,0], \\
P_{\mathrm{dip}}&=[0,-1,-1,0], \\
P_{\mathrm{biphasic}}&=[1,1,-1,-1].
\end{align}
Each family uses the sign-reversed distractor $P_{\mathrm{dis},p}=-P_p$, which produces no target response.
We cross the three triggers with three width-twelve response kernels:
\begin{align}
R_{\mathrm{positive}}&=[0,0.2,0.4,0.6,0.8,1,1,0.8,0.6,0.4,0.2,0], \\
R_{\mathrm{negative}}&=-R_{\mathrm{positive}}, \\
R_{\mathrm{biphasic}}&=[0,0.5,1,1,0.5,0,0,-0.5,-1,-1,-0.5,0].
\end{align}
Every kernel is zero outside its listed support.
The biphasic response concatenates a width-six positive triangle with its negative counterpart.
Each of the nine combinations is a separate condition: all teaching pairs and the test event within one window use the same trigger--response mapping.
The nine conditions are not mixed within a window.
The injected trigger energies $A^2\lVert P_p\rVert_2^2$ are 18, 18, and 36 for spike, dip, and biphasic triggers; the response energies $B^2\lVert R_r\rVert_2^2$ are 39.6, 39.6, and 45 for positive, negative, and biphasic responses.
Within each set, absolute peak amplitudes and nominal widths agree, but total energies and nonzero supports differ.

\paragraph{Windows and event placement.}
We generate 300 windows for each seed $s\in\{0,1,2,3,4\}$, trigger--response combination, and lag, using context length 768 and forecast horizon 96.
Time is indexed relative to the forecast origin: history occupies $t=-768,\ldots,-1$ and the forecast horizon occupies $t=0,\ldots,95$.
We denote the test-response anchor by $t_{\mathrm{resp}}$ and its
trigger anchor by $t_{\mathrm{trig}}=t_{\mathrm{resp}}-\tau$.
The main nine-combination sweep fixes $t_{\mathrm{resp}}=24$,
the 25th forecast sample.
The additional spike-to-positive position control varies
$t_{\mathrm{resp}}\in\{16,24,32,40\}$ at short lags
$\tau\in\{0,4,8,16\}$
(Appendix~\ref{app:temporal_position_controls}).
The main lag grid is
\begin{equation}
\tau\in\{0,4,8,12,16,20,24,32,40,48,64,96,128\}.
\end{equation}
The trigger is in the known-future segment for $\tau<t_{\mathrm{resp}}$,
at its first sample for $\tau=t_{\mathrm{resp}}$, and in history
for $\tau>t_{\mathrm{resp}}$.
Lag refers to the separation between injection anchors; the first nonzero samples and peaks need not coincide at $\tau=0$.

Each context contains nine teaching triggers on a grid with spacing 76, starting at $-762$, with independent integer jitter drawn uniformly from $\{-8,\ldots,8\}$.
Distractor anchors use the midpoint grid, offset by 38 steps, with independently drawn jitter.
Distractor grid points later than $-1-\tau-12$ are excluded, giving nine distractors through $\tau=96$ and eight at $\tau=128$.
For a fixed seed, lag, and window, background realizations and event placements are shared across all nine combinations.
Holding the response family fixed gives identical target arrays across triggers; holding the trigger fixed gives identical covariate arrays across responses.
Changing the response modifies both historical teaching responses and the test response.
Across lags, teaching responses move to $t_k+\tau$, so the target history also changes.

The original placement rules are retained for every combination.
For seeds 0 through 4, respectively, 44, 33, 41, 36, and 32 of the 2,700 scheduled teaching triggers extend beyond the left context boundary at each lag.
The retained teaching-trigger energy is approximately $99.61$--$99.72\%$ for spike and dip, and $99.40$--$99.56\%$ for biphasic triggers.
At zero lag, teaching responses at these clipped anchors can also be truncated; this placement rule is shared across response families.
The test trigger, shifted trigger, and test response are fully contained in the supplied window.

\paragraph{Conditioning and counterfactual inputs.}
Each frozen model receives the target history together with covariate history and known-future values through the same conditioning interface at every lag.
We use the native forecast quantile closest to 0.5 and each backbone's internal scaling, without additional experiment-level normalization.
The event and no-event inputs differ only by removal of the test trigger; the target history, teaching examples, distractors, and background values remain fixed within this pair.
For combination $c$, seed $s$, and window $w$, define
\begin{equation}
D_{csw,\tau}(h)
=\widehat Y_{\mathrm{event},csw,\tau}(h)
-\widehat Y_{\mathrm{no\text{-}event},csw,\tau}(h).
\end{equation}
The primary response profile is the within-seed window average,
\begin{equation}
\overline D_{cs,\tau}(h)
=\frac{1}{300}\sum_{w=1}^{300}D_{csw,\tau}(h).
\end{equation}

\paragraph{Gain, onset, and profile measurement.}
For a fixed response anchor, let
$Q_c(h)=BR_r(h-t_{\mathrm{resp}})$ be the true injected response
over the full forecast horizon.
We measure the signed projection gain
\begin{equation}
g_{cs,\tau}
=\frac{\langle\overline D_{cs,\tau},Q_c\rangle}
{\langle Q_c,Q_c\rangle}.
\end{equation}
This definition applies to positive, negative, and zero-sum biphasic responses.
The gain is 1 for the exact response, 0 for a zero response, and $-1$ for its exact sign reversal.
All nine combinations use this definition, including the reused positive-response forecasts.
For each valid response anchor $o\in\{0,\ldots,84\}$, the response-specific matched-filter score is
\begin{equation}
a_{cs,\tau}(o)
=\frac{\sum_{j=0}^{11}\overline D_{cs,\tau}(o+j)R_r(j)}
{\lVert R_r\rVert_2}.
\end{equation}
We select
$\widehat t_{\mathrm{resp},cs,\tau}=\arg\max_o|a_{cs,\tau}(o)|$,
resolving ties at the earliest anchor, and define signed onset error
$e_{cs,\tau}=\widehat t_{\mathrm{resp},cs,\tau}-t_{\mathrm{resp}}$.
Positive errors indicate a late response and negative errors an early response.
The absolute onset error is $|e_{cs,\tau}|$.
These metrics localize each combination's window-averaged profile; timing is not averaged over separately localized windows.
No detection threshold filters the reported model results.
An exactly zero response has no meaningful onset, even though an unmasked argmax returns a numerical index.
To capture deviations in response shape as well as amplitude, we also report
\begin{equation}
E_{cs,\tau}
=\frac{\lVert\overline D_{cs,\tau}-Q_c\rVert_2}{\lVert Q_c\rVert_2},
\end{equation}
with norms taken over the full horizon.
The exact response has $E=0$ and a zero response has $E=1$.

\paragraph{Aggregation and uncertainty.}
We use equal-combination averages within each seed,
\begin{equation}
G_{s,\tau}=\frac{1}{9}\sum_c g_{cs,\tau},\qquad
T_{s,\tau}=\frac{1}{9}\sum_c|e_{cs,\tau}|.
\end{equation}
Figure~\ref{fig:temporal_response_gain} reports the across-seed mean and SEM of $G_{s,\tau}$.
{Timing is a supplementary diagnostic, summarized by the across-seed mean of $T_{s,\tau}$.}
For any seed-level quantity $q_s$, SEM is
\begin{equation}
\operatorname{SEM}(q)
=\sqrt{\frac{\sum_s(q_s-\overline q)^2}{S(S-1)}},
\qquad S=5.
\end{equation}
Absolute timing errors are taken before averaging over combinations or seeds, so early and late errors do not cancel.
Profiles with different response signs are never averaged together before computing their metrics.
Figure~\ref{fig:temporal_gain_heatmap} shows combination-specific
gain means. {Appendix~\ref{app:temporal_fitted_baseline}
summarizes macro gain and profile error for the fitted baseline.}
This uncertainty describes five seeds of a fixed nine-combination evaluation.
Within-segment lag differences use paired seed-level quantities, with reference lag 0 for the known-future segment and 32 for history; lag 24 is reported separately.

\paragraph{Control arms and analytic references.}
The distractor arm replaces the test trigger with $-P_p$.
The shifted-trigger arm moves it eight steps later while retaining
the original historical demonstrations, implying an ideal response
anchor of $t_{\mathrm{resp}}+8$ (32 in the main setting).
Each control response is measured relative to the same no-event input.
We distinguish the shifted response's onset error relative to
$t_{\mathrm{resp}}+8$ from its onset shift relative to the
unshifted forecast.
The no-covariate arm removes both covariate history and covariate future, including all covariate teaching evidence; it is a global ablation rather than a test-trigger intervention.
The analytic event response is $Q_c(h)=BR_r(h-t_{\mathrm{resp}})$,
with gain 1 and onset error 0.
For the shifted analytic response
$Q_c^{\mathrm{shift}}(h)=BR_r(h-t_{\mathrm{resp}}-8)$,
gain relative to the original template is
$\langle Q_c^{\mathrm{shift}},Q_c\rangle/\lVert Q_c\rVert_2^2$.
This equals approximately $0.03636$ for the positive and negative responses and $-0.2$ for the biphasic response.
Scoring against its own shifted template instead gives gain 1 and
onset error 0 relative to anchor $t_{\mathrm{resp}}+8$.
The distractor has zero analytic response and no defined onset.

\paragraph{Scope.}
Lag variation changes the trigger's absolute position and the temporal arrangement of teaching responses, and traverses the future--history boundary.
Within-segment comparisons keep the trigger's input segment fixed but do not isolate temporal distance from absolute position.
The trigger and response families also differ in energy and support, so their comparison does not isolate shape at equal signal energy.
Detailed results are reported in Appendix~\ref{app:temporal_results}.

\subsection{Detailed Responses and Controls}
\label{app:temporal_results}

\begin{figure}[tbp]
    \centering
    \includegraphics[width=\linewidth]{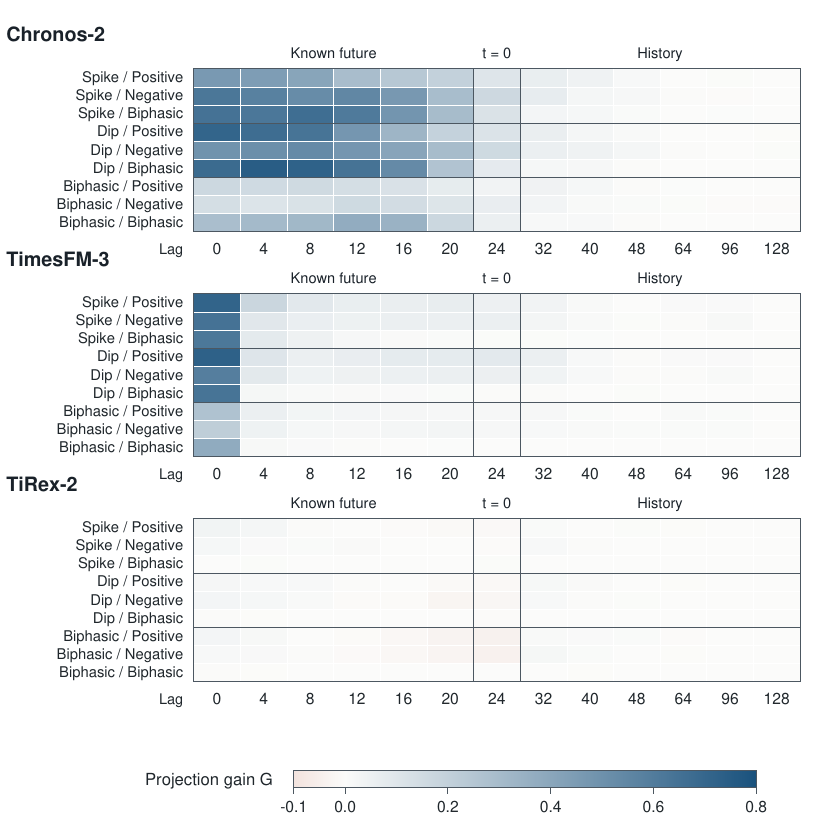}
    \caption{Signed projection gain across all nine trigger--response
    combinations and 13 lags. Rows identify trigger / target-response
    families; cells are means across five seeds on a shared color scale.
    Columns show discrete lag settings. With $t_{\mathrm{resp}}=24$,
    the trigger is in the known future for $\tau<24$, at the forecast
    origin for $\tau=24$, and in history for $\tau>24$.}
    \label{fig:temporal_gain_heatmap}
\end{figure}

Figure~\ref{fig:temporal_gain_heatmap} shows all nine
trigger--response combinations across the 13 tested lags.
Metrics are computed from each seed's window-averaged response
profile before aggregation across five seeds.
{Figure~\ref{fig:temporal_response_gain} reports
equal-combination gains through $\tau=64$.}
Definitions are given in Appendix~\ref{app:temporal_setup}.

\paragraph{Combination-specific patterns.}
At zero lag, Chronos-2 and TimesFM-3 respond less strongly to
biphasic triggers than to spikes or dips for each target-response
family, even though both can produce substantial biphasic target
responses. This distinguishes trigger and response roles without
isolating shape from energy or support.
TimesFM-3's reduction at $\tau=4$ occurs in every combination
and seed. The paired macro-gain changes from $\tau=0$ to $8$
are $-0.01299\pm0.00209$ for Chronos-2 and
$-0.49167\pm0.00507$ for TimesFM-3.
Chronos-2's attenuation varies across combinations and is not
uniformly monotonic; its onset error remains zero through
$\tau=20$ in every combination and seed.
When gain becomes small, a matched-filter onset estimate does
not independently establish a meaningful target response.

{
\paragraph{Responses of $t_0$ and CITRAS-FM.}
$t_0$ and CITRAS-FM have small mean projection gains across
the tested lags, including at $\tau=0$.
At zero lag, $t_0$'s mean gain is largest for biphasic target
responses within each trigger family.
CITRAS-FM's combination-specific mean gains remain small
throughout the lag grid.
At the terminal historical lags $\tau=96$ and $128$, the
five-seed mean gains are $0.00188$ and $0.00021$ for $t_0$,
and $-0.00013$ and $-0.00014$ for CITRAS-FM, respectively.
}

\paragraph{Distractor and shifted-trigger controls.}
At zero lag, absolute distractor gain is smaller than event gain
for Chronos-2 and TimesFM-3 in every combination and seed.
TiRex-2's distractor gain exceeds its event gain in the
biphasic-to-negative condition in every seed.
Shifting the trigger eight steps later shifts Chronos-2's
estimated response anchor by eight steps in every combination
and seed; TimesFM-3's shifted onset error is zero or $-1$.
These controls distinguish responsiveness from selective use
of the trigger.

{
The distractor and shifted-trigger controls also evaluate
$t_0$ and CITRAS-FM under the same protocol.
}

\subsection{Event-Response Regression Fitted to the Observed Context}
\label{app:temporal_fitted_baseline}

\paragraph{Method and prior.}
We fit a polarity-gated event-response regression to each window's
target and covariate histories.
The model class assumes sparse, large covariate events, transient
target responses, and a local-level background.
Detected events can induce separate response kernels for positive
and negative onset polarities.
The estimator is not given the generating lag, trigger or response
family, kernels, event annotations, teaching-event locations, or
noiseless target background.

The event detector uses a strict threshold of four on the standardized
covariate and a median-absolute-deviation scaling constant of $1.4826$.
Bursts separated by at most two non-flagged samples are merged, and
the onset's first polarity is retained.
The onset rule reads only the current and preceding three samples,
so extending the supplied path into the future does not revise
previously fixed onsets.
\paragraph{Fitting the response kernels.}
Let $N_q(t)$ be the detected onset sequence for polarity
$q\in\{+,-\}$, and let $k_{q,\ell}$ denote its response kernel
at lag $\ell\in\{0,\ldots,160\}$.
The event contribution is
\begin{equation}
    g_t=\sum_{q\in\{+,-\}}\sum_{\ell=0}^{160}
    k_{q,\ell}N_q(t-\ell).
\end{equation}
We fit the stacked kernel vector $k$ to the target's first
differences $\Delta Y_t=Y_t-Y_{t-1}$.
The design has entries
$D_{t,(q,\ell)}=N_q(t-\ell)-N_q(t-1-\ell)$, and the fitting
objective is
\begin{equation}
    \min_k\;\|\Delta Y-Dk\|_2^2
    +d\lambda_{\mathrm{sm}}\|D_2k\|_2^2
    +d\,10^{-4}\|k\|_2^2.
\end{equation}
Here $D_2$ applies separate second differences to the two
polarity kernels, with zero extension at their boundaries.
The scale $d$ is the mean of the positive diagonal entries
of $D^\top D$ on the retained design, or one if none are positive;
it does not use a generating amplitude or noise scale.
The first 161 difference rows are removed in each fit,
leaving 606 rows for the full 768-step context.

\paragraph{Selecting regularization and forecasting.}
The smoothness weight $\lambda_{\mathrm{sm}}$ is selected from
$\{0,0.1,1,10\}$ using history prefixes of length 480, 576,
and 672, each followed by a 96-step validation interval within
the observed context.
For each prefix, covariate centering and scaling are estimated from
that prefix only. The detector uses these statistics with the fixed
threshold, and response kernels are fitted to the same prefix.
The following covariate values are supplied as known-future inputs
to the validation forecast, without entering the fitted scaler or kernels.
Selection minimizes the unweighted mean of the three folds'
96-step \emph{level-forecast MSE in raw target units}, rather than
the penalized first-difference objective used for fitting.
Ties favor the larger smoothness weight, with comparison tolerance
$10^{-12}$ relative to score magnitude and a unit floor.
The final held-out target horizon is never used in this selection.

After refitting at the selected weight on the full observed history,
the level forecast is
\begin{equation}
    \widehat Y_h=Y_{-1}-\widehat g_{-1}+\widehat g_h.
\end{equation}
Future covariate values enter the detected-event sequence used
for prediction. Counterfactual changes confined to the future
share the historical fit; changes to the covariate history
recompute detection, regularization selection, and the kernels.
If no burst is detected, the procedure reduces to a local-level
forecast. No such fallback occurred in the reported temporal grid.
The no-covariate arm directly repeats the last observed target value.

\paragraph{Response gains and profile errors.}
We evaluate the same nine event--response combinations, 13 lags,
five seeds, and 300 windows per combination and seed as the
frozen-model experiment.
Gains and profile errors are computed from the window-averaged
event-minus-no-event response, using the original signed template.
Equal-combination mean gain remains between $0.670$ and $0.780$
over the full lag grid, while mean profile error ranges from
$0.256$ to $0.358$.
At each lag, its mean gain is higher and its mean profile error
is lower than those of each TSFM.
These are comparisons of the nine-combination averages, not
claims that every individual combination favors the baseline.
For example, Chronos-2 has lower profile error for the dip-to-biphasic
combination at $\tau=4$.

All 117 combination--lag mean gains are positive.
Matched-filter onset error is zero for every seed's window-averaged
profile, which does not imply exact timing in every individual window.
At zero lag, mean distractor gains range from $-0.0027$ to $0.0019$;
shifting the trigger by eight steps shifts the baseline's averaged
response by eight steps in every combination and seed.
The results demonstrate substantial recoverability from the supplied
contexts under an explicit event-response model class.

{
\subsection{Forecast-Position Control}
\label{app:temporal_position_controls}
For spike-to-positive responses, we evaluate anchors
$\{16,24,32,40\}$ and lags $\{0,4,8,16\}$ with Chronos-2,
TimesFM-3, TiRex-2, and EventResponseFit under the same sample counts.
TimesFM-3 attenuates at nonzero lags at every anchor.
Averaging paired gain changes from lag zero to 16 over the strictly
future anchors $\{24,32,40\}$ gives $-0.2501\pm0.0046$ for Chronos-2,
$-0.7721\pm0.0033$ for TimesFM-3, $-0.0318\pm0.0031$ for TiRex-2,
and $0.0050\pm0.0046$ for EventResponseFit (five-seed SEM).
Thus attenuation is not confined to anchor 24, although lag, trigger
position, and response position are not independently identified.
}

{
\subsection{Observed-Background Event Responses}
\label{app:bc_temporal_real_background}
We plant the same nine mappings in GFC17 electricity-load and KDD
power records. Source and target backgrounds come from different
items and nonoverlapping calendar intervals of the same measured
variable, rather than natural temperature--load or wind--power pairs.
Records are divided into 864-sample blocks. Historical eligibility
requires two finite samples, at most $5\%$ missingness, and filled
standard deviation above $10^{-8}$.
Each seed samples 300 eligible pairs with replacement, shared across
models and conditions. Causal forward filling initializes leading gaps
by the history median; each channel is normalized by its filled history.
The source-trigger and target-response amplitudes are fixed at 10 and
$1.7647113485840686$ normalized units, respectively; the target amplitude is
calibrated to the expected synthetic-background history variance, independently
of forecast outcomes. Missing source-future values are forward-filled, while missing
target-future values retain an explicit scoring mask.
We retain anchor 24 and lags $\{0,4,16,32,64\}$ and the same
window-profile, equal-mapping, five-seed aggregation.
Table~\ref{tab:slim_bc_real_temporal} shows attenuation on both backgrounds;
EventResponseFit retains larger positive gains at historical lags.
}
\begin{table}[tbp]
\centering
\caption{Projection gain on observed backgrounds. Entries are equal-mapping mean $\pm$ five-seed SEM.}
\label{tab:slim_bc_real_temporal}
\footnotesize\setlength{\tabcolsep}{3pt}
\resizebox{\linewidth}{!}{%
\begin{tabular}{llrrrrr}
\toprule
Dataset & Method & $\tau=0$ & $4$ & $16$ & $32$ & $64$ \\
\midrule
GFC17 & Chronos-2 & $0.3495\pm0.0024$ & $0.3331\pm0.0055$ & $0.0388\pm0.0008$ & $0.0131\pm0.0006$ & $0.0029\pm0.0003$ \\
GFC17 & TimesFM-3 & $0.1756\pm0.0015$ & $0.0293\pm0.0003$ & $0.0097\pm0.0003$ & $0.0029\pm0.0002$ & $0.0050\pm0.0002$ \\
GFC17 & TiRex-2 & $0.0221\pm0.0005$ & $0.0090\pm0.0003$ & $0.0025\pm0.0002$ & $0.0089\pm0.0002$ & $-0.0024\pm0.0001$ \\
GFC17 & $t_0$ & $0.0204\pm0.0003$ & $-0.0006\pm0.0003$ & $-0.0013\pm0.0002$ & $-0.0021\pm0.0003$ & $0.0001\pm0.0002$ \\
GFC17 & CITRAS-FM & $0.0046\pm0.0003$ & $0.0082\pm0.0003$ & $0.0023\pm0.0002$ & $0.0041\pm0.0002$ & $-0.0005\pm0.0000$ \\
GFC17 & EventResponseFit & $0.4371\pm0.0045$ & $0.4697\pm0.0039$ & $0.4477\pm0.0026$ & $0.4375\pm0.0050$ & $0.3962\pm0.0029$ \\
\midrule
KDD & Chronos-2 & $0.2282\pm0.0035$ & $0.1793\pm0.0020$ & $0.0308\pm0.0005$ & $0.0052\pm0.0007$ & $0.0003\pm0.0006$ \\
KDD & TimesFM-3 & $0.1444\pm0.0029$ & $0.0152\pm0.0002$ & $0.0103\pm0.0006$ & $0.0033\pm0.0001$ & $0.0048\pm0.0002$ \\
KDD & TiRex-2 & $0.0380\pm0.0007$ & $0.0222\pm0.0005$ & $0.0060\pm0.0005$ & $0.0056\pm0.0004$ & $-0.0008\pm0.0001$ \\
KDD & $t_0$ & $0.0371\pm0.0008$ & $0.0065\pm0.0002$ & $0.0008\pm0.0002$ & $0.0008\pm0.0004$ & $0.0012\pm0.0002$ \\
KDD & CITRAS-FM & $0.0048\pm0.0001$ & $0.0071\pm0.0003$ & $0.0007\pm0.0002$ & $-0.0017\pm0.0001$ & $-0.0006\pm0.0002$ \\
KDD & EventResponseFit & $0.3599\pm0.0053$ & $0.3761\pm0.0054$ & $0.3439\pm0.0116$ & $0.3595\pm0.0057$ & $0.3309\pm0.0054$ \\
\bottomrule
\end{tabular}}
\end{table}

\clearpage
\section{Cross-Channel Dependency Structure}
\label{app:structural_experiments}
{
\subsection{Construction and Relative Changes}
\label{app:evidence_scaling}
We use $z_t=\rho z_{t-1}+\sqrt{1-\rho^2}u_t$,
$Y_t=z_t+\epsilon_t/\sqrt2$, and
$X_{j,t}=0.75z_t+\sqrt{1-0.75^2}\eta_{j,t}$.
The independent $\eta_j$ have the same stationary unit variance and
AR(1) persistence as $z$; all innovations are independent standard
Gaussians. We test $\rho\in\{0,.3,.6,.9,.95,.98\}$,
512-step contexts, horizon eight, and 256 paired windows in each of
five seeds after 512 burn-in steps.
Independent channels have distinct noise; duplicates repeat one
observation; distractors retain one informative channel and add
independent matched series. Counts are $K\in\{1,2,4,8,16\}$.
Conditions share targets and nested channel sets.

Let $R_{M,s}(1)$ denote seed $s$'s window-mean noisy-single-channel MAE.
Independent gains and duplicate/distractor penalties in the main figure are
\begin{equation}
100\frac{R_{M,s}(1)-R_{M,s}^{\mathrm{ind}}(K)}{R_{M,s}(1)},
\qquad
100\frac{R_{M,s}^{f}(K)-R_{M,s}(1)}{R_{M,s}(1)}.
\end{equation}
Direct-latent reductions instead use each method's target-only baseline.
Ratios are formed within seed before mean and five-seed SEM.
The analytic reference filters $(z,\eta_1,\ldots,\eta_K)$ using exactly
the supplied history and forecasts $\rho^h\mathbb E[z_T\mid\mathcal I_T]$.
With latent history this becomes $\rho^hz_T$; duplicates and independent
distractors leave the reference information unchanged.
}

\subsection{Structural Dependency Experiments: Numerical Results}
\label{app:structural_numerical_results}

\paragraph{Evaluation and reporting.}
These results use five seeds (0--4), with 256 paired target
windows per seed, context length 512, and forecast horizon 8.
All auxiliary observations are supplied as historical rows
through the same multivariate interface; no future latent values
are supplied.
We score the target's median forecast.
{For raw-unit comparisons,} we first average paired MAE differences
within each seed, then report their mean and sample standard
deviation divided by $\sqrt{5}$.
Oracle entries are empirical MAEs of the analytic conditional
forecasts on the same windows, rather than population expected MAEs.
Each oracle uses only the observations available in its condition
and accounts for the AR(1) idiosyncratic components.
The latent-state representation probes in
Appendix~\ref{app:distributed_diagnostics} are a separate
three-seed experiment and are not pooled with this sweep.

{
\paragraph{Direct latent access.}
Figure~\ref{fig:evidence_scaling}a reports relative MAE reductions
from latent history across persistence for all five TSFMs and
the same-information analytic reference.
Chronos-2 and TimesFM-3 obtain the largest relative gains in the
persistent settings; $t_0$ also benefits, while TiRex-2 and
CITRAS-FM obtain smaller gains.
The reference gain varies with persistence and is exactly zero
at $\rho=0$, where latent history does not predict the future state.

\paragraph{Independent observations.}
Figure~\ref{fig:evidence_scaling}b compares gains from independent
observations with each method's noisy single-channel reference.
At $\rho=0.95$, TimesFM-3 continues to benefit as observations
are added, whereas Chronos-2's mean gain declines beyond an
intermediate channel count.
$t_0$ and CITRAS-FM have smaller mean gains than TimesFM-3,
and SparseDirect has modest mean gains with substantial seed variation.
The fitted SSM attains gains close to the analytic reference.
The paired $K=4$ versus $K=16$ comparison is detailed in
Appendix~\ref{app:paired_channel_count}.

\paragraph{Duplicate and distractor controls.}
Figure~\ref{fig:evidence_scaling}c,d reports percentage MAE changes
from redundant or irrelevant channels.
At the largest channel count, Chronos-2 has a positive mean duplicate
penalty, while CITRAS-FM's mean error decreases as copies are added.
The mean duplicate changes of the other methods are smaller.
Distractors increase the mean errors of Chronos-2, TiRex-2, and $t_0$;
TimesFM-3 and CITRAS-FM change less.
SparseDirect's largest-count mean distractor penalty exceeds those
of all five TSFMs.
The SSM's exact duplicate invariance follows its explicit duplicate
removal; the analytic reference is invariant to both controls because
they do not change the predictive information set.

\paragraph{Absolute accuracy and measured scope.}
Table~\ref{tab:bc_structural_accuracy_summary} gives absolute MAEs
for the reference conditions and the largest channel count at
$\rho=0.95$, using the same 256 paired windows within each of five seeds.
SSM and SparseDirect are evaluated only at this persistence;
SparseDirect is not evaluated under direct latent access.
}

\begin{table}[tbp]
\centering
\caption{Absolute target MAE at $\rho=0.95$ (five-seed mean). Conditions share 256 paired windows per seed; --- denotes an unevaluated condition.}
\label{tab:bc_structural_accuracy_summary}
\footnotesize\setlength{\tabcolsep}{4pt}
\resizebox{\linewidth}{!}{%
\begin{tabular}{lrrrrrr}
\toprule
Method & Target only & Noisy $K=1$ & Latent & Ind $K=16$ & Dup $K=16$ & Dis $K=16$ \\
\midrule
Chronos-2 & 0.7957 & 0.7786 & 0.7600 & 0.7725 & 0.7865 & 0.7896 \\
TimesFM-3 & 0.7916 & 0.7791 & 0.7534 & 0.7587 & 0.7813 & 0.7792 \\
TiRex-2 & 0.7927 & 0.7845 & 0.7825 & 0.7819 & 0.7837 & 0.7923 \\
$t_0$ & 0.8022 & 0.7859 & 0.7753 & 0.7785 & 0.7860 & 0.7907 \\
CITRAS-FM & 0.7898 & 0.7871 & 0.7854 & 0.7827 & 0.7830 & 0.7881 \\
Oracle & 0.7752 & 0.7599 & 0.7318 & 0.7359 & 0.7599 & 0.7599 \\
SSM & 0.7811 & 0.7621 & 0.7358 & 0.7404 & 0.7621 & 0.7642 \\
SparseDirect & 0.8045 & 0.8009 & --- & 0.7921 & 0.8017 & 0.8217 \\
\bottomrule
\end{tabular}}
\end{table}

{
Removing exact source copies restores the single-observation input
by construction. Near-duplicate removal is not evaluated.
}

\subsection{A Fitted State-Space Baseline for Shared-Latent Dependencies}
\label{app:structural_fitted_baselines}

\paragraph{Input preprocessing.}
The state-space baseline receives the target history and the
covariate histories of the corresponding condition, without future
covariates.
Within each observed context, a covariate is removed as constant
when its population standard deviation is at most
$10^{-12}\max(1,\max_t|X_t|)$.
Remaining covariates are deduplicated only when their historical
columns are bitwise identical, and are ordered by a digest of
their own historical values.
The target is always retained.
The retained channels are centered and scaled using their context
means and population standard deviations; a constant target uses
a scaling denominator of one.
No generating parameter or condition label enters this preprocessing.

\paragraph{Fitted state-space model: assumptions and observations.}
The preprocessing above is applied to the full observed context,
without cross-validation.
Let $C$ be the number of retained channels, including the target.
In standardized units, define the state and observation vectors as
\begin{align}
    \mathbf s_t&=(m_t,\widetilde X_{1,t},\ldots,
        \widetilde X_{C-1,t})^\top,\\
    \mathbf v_t&=(\widetilde Y_t,\widetilde X_{1,t},\ldots,
        \widetilde X_{C-1,t})^\top.
\end{align}
The Gaussian state-space model is
\begin{align}
    \mathbf s_t&=r\mathbf s_{t-1}+\mathbf u_t,
        &\mathbf u_t&\sim\mathcal N(0,Q),\\
    \mathbf v_t&=\mathbf s_t+\mathbf e_t,
        &\mathbf e_t&\sim\mathcal N(0,R),\\
    R&=\operatorname{diag}(R_y,0,\ldots,0),
        &\mathbf s_1&\sim\mathcal N\!\left(0,\frac{Q}{1-r^2}\right).
\end{align}
The observation matrix is fixed to the identity and $Q$ is a
full innovation covariance.
Within the observed context, only the target signal $m_t$ is hidden;
the covariates are observed directly as components of the state.
Shared scalar persistence and white target measurement noise
are explicit model-class assumptions.
The true generating persistence $\rho$ is not supplied to the fit.

\paragraph{State-space fitting and start selection.}
Expectation--maximization (EM) estimates $r$, $Q$, and $R_y$,
including the stationary initial law in the complete-data likelihood.
{The M-step updates the innovation covariance and numerically
optimizes persistence over $r\in[-0.995,0.995]$.}
The target measurement variance is updated in closed form.
Both the eigenvalue floor for $Q$ and the floor for $R_y$ are
$10^{-8}$ in standardized units.
The zero covariate measurement variances are retained exactly;
covariance updates use the Joseph form and symmetrization.

Three deterministic starts initialize $R_y$ at $0.1$, $0.5$,
and $0.9$ of the standardized target variance.
Each permits at most 100 EM iterations, with relative tolerance
$10^{-5}$ for three consecutive iterations.
Likelihood-decreasing updates are damped by factors
$0.5$, $0.25$, and $0.125$, or rejected if they still decrease
the observed-history likelihood.
Among converged starts, the one with highest history likelihood
is selected. If no start converges, the best finite iterate across
all starts is retained and marked nonconverged.
No target values beyond the final forecast origin enter these decisions.
The standardized target forecast is
$r^h\mathbb E[m_T\mid\mathbf v_{1:T}]$, which is returned to raw
units using the target mean and scale from the same context.
This baseline is evaluated only at $\rho=0.95$, matching the
prespecified persistence level of the main structural figure.

\paragraph{Evaluation and convergence.}
The state-space baseline covers all 15 channel conditions at $\rho=0.95$.
It uses the original 512-step contexts, eight-step horizons,
256 windows per seed, and five seeds.
All 19,200 logical state-space forecast requests have a
converged selected solution and finite forecasts, although
not every initialization converges.
Results are reported over all finite forecasts without
excluding a difficult window based on convergence.

\paragraph{Accuracy and sensitivity to channel composition.}
The fitted state-space model has lower MAE than each TSFM
in every structural condition at $\rho=0.95$, with the same
direction in all five seeds.
At independent $K=16$, its MAE is $0.7404\pm0.0095$ and
its paired gap to the same-information oracle is
$0.0045\pm0.0018$.
Its improvement over its own noisy $K=1$ baseline is
$0.0217\pm0.0027$.
These results provide an achievable reference under the
specified fitted state-space class.

The state-space baseline's distractor-$K=16$ penalty relative
to its own noisy-$K=1$ condition is $0.0021\pm0.0019$.
Its duplicate penalties are exactly zero because of explicit
duplicate removal, not learned robustness to redundancy.

\subsection{Paired Changes in Channel Count}
\label{app:paired_channel_count}

We additionally analyze paired changes between channel counts
using the archived forecasts. The focal comparison uses
$\rho=0.95$, $\lambda=0.75$, independent observations, and
$K=4$ versus $K=16$.
For each method, we first compute each window's MAE difference
between the two channel counts, then average within each seed:
\begin{equation}
    d_s=\frac{1}{N}\sum_{w=1}^{N}
    \left[e_{s,w}(16)-e_{s,w}(4)\right].
\end{equation}
Positive values indicate worse performance with more channels.
All methods share the same 256 valid windows per seed.
We report the mean, SEM, and descriptive 95\% Student-$t$
interval over five seed-level differences.
The SEM is computed from these paired differences rather
than from two separately estimated gain SEMs.
This is an exploratory analysis of existing forecasts.

Increasing $K$ from 4 to 16 raises Chronos-2's MAE by
$0.00717\pm0.00200$ and lowers TimesFM-3's MAE by
$0.00943\pm0.00196$; both directions hold in all five seeds.
TiRex-2's change is $-0.00069\pm0.00043$, and its descriptive
95\% interval includes zero.
The oracle and the fitted state-space baseline improve in the same
comparison.
This comparison quantifies the loss of Chronos-2's gain beyond
an intermediate channel count; it does not imply deterioration
relative to $K=1$.

{
\subsection{Context-Selected Sparse Direct Reference}
\label{app:structural_sparse_direct}
SparseDirect retains raw constant and duplicate channels and fits a
separate Elastic Net regression for each forecast step, with an
unpenalized intercept. Candidates use target-only or all-channel
histories; lag lengths $\{1,2,4,8,16,32,64\}$; L1 ratios
$\gamma\in\{.1,.3,.5,.8,1\}$; and relative penalties
$a\in\{1,.3,.1,.03,.01,.003,.001\}$, giving 490 procedures.
For centered training design $D$ and response $y_h$, the squared-loss
normalization is $2n$ and penalty scale is
$\alpha_h=a\|D^\top y_h\|_\infty/(n\gamma)$.
Channel scalers use each training prefix; zero standard deviations
use scale one.

The two training cutoffs are $c=\lfloor0.625T\rfloor$ and
$\lfloor0.875T\rfloor$, with 25 validation origins $c-1,\ldots,c+23$
per fold. Training origins $p-1,\ldots,c-H-1$ keep their responses
inside the prefix. Coefficients remain fixed within each validation fold.
One input configuration and hyperparameter setting minimizes pooled
raw-unit MAE across folds and forecast steps.
Ties within $10^{-12}$ prefer target-only inputs, shorter lags,
stronger penalties, then larger L1 ratios.
Cyclic coordinate descent uses tolerance $10^{-5}$ and at most
10,000 iterations, retaining finite outputs at the iteration limit.
The selected procedure and scalers are refitted on the full history.
}

\clearpage
\section{Conditionally Irrelevant Future Covariates}
\label{app:known_future_experiments}
\label{app:known_future_numerical_results}
This section supplements the lagged-copy experiment in
Section~\ref{sec:irrelevant_future_covariates}.
Reliability mismatch is reported separately in
Appendix~\ref{app:reliability_experiments}, and lag-class representation
probes in Appendix~\ref{app:lag_class_diagnostics}.
{
\subsection{Construction and Measurements}
\label{app:known_future_conditioning}
We generate independent standard Gaussian $X_t,\epsilon_t$ and
$Y_t=\beta X_{t-L}+\sqrt{1-\beta^2}\epsilon_t$, with
$T=512$, $H=8$, $L\in\{8,16\}$, and
$\beta\in\{0.4,0.6,0.8,1\}$.
Each of five seeds supplies 256 matched windows.
Since $L\ge H$, every target-relevant source value is already observed;
future $X$ is conditionally irrelevant.
The oracle MAE is $\sqrt{2/\pi}\sqrt{1-\beta^2}$.
CI, MV, PastCov, and KC use the inputs defined in
Appendix~\ref{app:general_protocol}.
We report the paired benefit
$\Delta_{\mathrm{KC}}=\mathrm{MAE}_{\mathrm{PastCov}}-\mathrm{MAE}_{\mathrm{KC}}$.

For $E_J$, we perturb each predictive historical source value by
$\pm0.5$ on the first 64 windows per seed and use central differences.
The operator is averaged across windows and seeds before computing
$\|J-\beta I_H\|_F/\|\beta I_H\|_F$; it is undefined at $\beta=0$.
For $S_{\mathrm{future}}$, histories remain fixed while eight independent
Gaussian source continuations are supplied. We average the 28 pairwise
absolute differences between median forecasts over the horizon and
windows, then report five-seed mean $\pm$ SEM.
The oracle sensitivity is zero when $L\ge H$.

The no-dependence control uses $\beta=0,L=8$; relevant-future controls
use $\beta=0.8,L\in\{0,4\}$, with oracle sensitivity
$\beta(1-L/H)2/\sqrt\pi$.
Across the five models, the measured no-dependence sensitivities range
from 0.0073 to 0.0705, confirming that sensitivity alone does not
establish use of a predictive relationship.
}

{
\subsection{Accuracy and Future-Value Sensitivity}
\label{app:known_future_accuracy_details}
\label{app:known_future_sensitivity_details}

The exact- and noisy-copy comparisons in
Table~\ref{tab:irrelevant_future_main} distinguish the use of
historical information from sensitivity to future values.
\textsc{KC} improves TimesFM-3 and TiRex-2 relative to
\textsc{PastCov}; Chronos-2's benefit depends on the lag and
relation strength. At $\beta=1,L=8$, both MAE and $E_J$ increase
for Chronos-2. For $t_0$ and CITRAS-FM, mean MAE is higher under
\textsc{KC} than \textsc{PastCov} throughout the positive-$\beta$
grid, while the historical-source response errors remain close to one.
Supplementary weaker-relationship results are given in
Appendix~\ref{app:known_future_beta_sweep}.
Table~\ref{tab:ade_irrelevant_controls} reports the no-dependence
and relevant-future controls for all five models.
Nonzero sensitivity in the no-dependence control shows that future
sensitivity does not by itself imply use of a predictive relationship.

\begin{table}[p]
\centering\footnotesize
\setlength{\tabcolsep}{3pt}
\caption{No-dependence ($\beta=0,L=8$) and relevant-future ($\beta=0.8,L<8$) controls. Changes and sensitivity use five-seed SEM. The $\beta=0,L=16$ condition is not evaluated.}
\label{tab:ade_irrelevant_controls}
\resizebox{\linewidth}{!}{%
\begin{tabular}{lrrrrrrrr}
\toprule
Model & $\beta$ & $L$ & CI & MV & PastCov & KC & $\Delta_{\mathrm{KC}}$ & $S_{\mathrm{future}}$ \\
\midrule
Chronos-2 & 0.0 & 8 & 0.8055 & 0.8053 & 0.8053 & 0.8056 & $-0.0002\pm0.0003$ & $0.0152\pm0.0006$ \\
TimesFM-3 & 0.0 & 8 & 0.8049 & 0.8042 & 0.8042 & 0.8044 & $-0.0002\pm0.0004$ & $0.0218\pm0.0006$ \\
TiRex-2 & 0.0 & 8 & 0.8053 & 0.8052 & 0.8052 & 0.8052 & $-0.0000\pm0.0002$ & $0.0126\pm0.0006$ \\
$t_0$ & 0.0 & 8 & 0.8054 & 0.8057 & 0.8055 & 0.8079 & $-0.0024\pm0.0010$ & $0.0705\pm0.0006$ \\
CITRAS-FM & 0.0 & 8 & 0.8073 & 0.8067 & 0.8068 & 0.8072 & $-0.0004\pm0.0001$ & $0.0073\pm0.0002$ \\
Chronos-2 & 0.8 & 0 & 0.7925 & 0.7928 & 0.7928 & 0.4922 & $0.3006\pm0.0040$ & $0.9359\pm0.0020$ \\
TimesFM-3 & 0.8 & 0 & 0.7912 & 0.7911 & 0.7911 & 0.4856 & $0.3055\pm0.0053$ & $0.9130\pm0.0015$ \\
TiRex-2 & 0.8 & 0 & 0.7919 & 0.7919 & 0.7919 & 0.4941 & $0.2979\pm0.0046$ & $0.9124\pm0.0032$ \\
$t_0$ & 0.8 & 0 & 0.7922 & 0.7922 & 0.7924 & 0.5011 & $0.2913\pm0.0043$ & $0.8268\pm0.0008$ \\
CITRAS-FM & 0.8 & 0 & 0.7942 & 0.7939 & 0.7947 & 0.7842 & $0.0105\pm0.0004$ & $0.0420\pm0.0004$ \\
Chronos-2 & 0.8 & 4 & 0.8021 & 0.6840 & 0.6840 & 0.5181 & $0.1659\pm0.0047$ & $0.5429\pm0.0018$ \\
TimesFM-3 & 0.8 & 4 & 0.8018 & 0.7988 & 0.7988 & 0.7239 & $0.0749\pm0.0037$ & $0.2912\pm0.0019$ \\
TiRex-2 & 0.8 & 4 & 0.8026 & 0.8015 & 0.8015 & 0.6625 & $0.1390\pm0.0051$ & $0.3953\pm0.0006$ \\
$t_0$ & 0.8 & 4 & 0.8028 & 0.8029 & 0.8034 & 0.8021 & $0.0013\pm0.0011$ & $0.0781\pm0.0021$ \\
CITRAS-FM & 0.8 & 4 & 0.8044 & 0.8008 & 0.7993 & 0.7979 & $0.0015\pm0.0005$ & $0.0325\pm0.0006$ \\
\bottomrule
\end{tabular}%
}
\end{table}

}

{
\subsection{Weaker Relationships}
\label{app:known_future_beta_sweep}
Table~\ref{tab:slim_weak_beta} supplements the $\beta=0.8,1$
main comparison with $\beta=0.4,0.6$.
TimesFM-3 and TiRex-2 retain positive mean KC benefits, although
some weak-relation gains are small relative to seed variation.
For $t_0$ and CITRAS-FM the mean benefit is negative at both lags;
their historical-source response errors remain near one.
All five models remain sensitive to irrelevant future values.
\begin{table}[t]
\centering\footnotesize
\setlength{\tabcolsep}{3pt}
\caption{Weak-relation controls. $\Delta_{\mathrm{KC}}$ and $S_{\mathrm{future}}$ are mean $\pm$ SEM over five paired seeds; $E_J^{\mathrm{KC}}$ uses the pooled response operator. All lags satisfy $L\ge H=8$.}
\label{tab:slim_weak_beta}
\resizebox{\linewidth}{!}{%
\begin{tabular}{lrrrrrrr}
\toprule
& & \multicolumn{3}{c}{$\beta=0.4$} & \multicolumn{3}{c}{$\beta=0.6$} \\
\cmidrule(lr){3-5}\cmidrule(lr){6-8}
Model & $L$ & $\Delta_{\mathrm{KC}}$ & $E_J^{\mathrm{KC}}$ & $S_{\mathrm{future}}$ & $\Delta_{\mathrm{KC}}$ & $E_J^{\mathrm{KC}}$ & $S_{\mathrm{future}}$ \\
\midrule
Chronos-2 & 8 & $0.0074\pm0.0019$ & 0.700 & $0.0667\pm0.0014$ & $0.0254\pm0.0027$ & 0.193 & $0.1504\pm0.0006$ \\
Chronos-2 & 16 & $0.0009\pm0.0010$ & 0.803 & $0.0499\pm0.0013$ & $0.0226\pm0.0027$ & 0.446 & $0.1213\pm0.0013$ \\
TimesFM-3 & 8 & $0.0002\pm0.0002$ & 0.989 & $0.0277\pm0.0003$ & $0.0076\pm0.0016$ & 0.960 & $0.0524\pm0.0006$ \\
TimesFM-3 & 16 & $0.0003\pm0.0003$ & 0.994 & $0.0234\pm0.0005$ & $0.0026\pm0.0010$ & 0.983 & $0.0295\pm0.0007$ \\
TiRex-2 & 8 & $0.0074\pm0.0023$ & 0.892 & $0.0342\pm0.0011$ & $0.0479\pm0.0063$ & 0.733 & $0.0797\pm0.0015$ \\
TiRex-2 & 16 & $0.0008\pm0.0013$ & 0.978 & $0.0194\pm0.0003$ & $0.0154\pm0.0015$ & 0.908 & $0.0351\pm0.0007$ \\
$t_0$ & 8 & $-0.0022\pm0.0012$ & 0.999 & $0.0704\pm0.0006$ & $-0.0023\pm0.0008$ & 0.999 & $0.0719\pm0.0008$ \\
$t_0$ & 16 & $-0.0020\pm0.0008$ & 1.000 & $0.0709\pm0.0009$ & $-0.0019\pm0.0012$ & 1.000 & $0.0731\pm0.0008$ \\
CITRAS-FM & 8 & $-0.0009\pm0.0002$ & 0.998 & $0.0092\pm0.0003$ & $-0.0017\pm0.0002$ & 0.997 & $0.0132\pm0.0005$ \\
CITRAS-FM & 16 & $-0.0003\pm0.0002$ & 1.000 & $0.0069\pm0.0001$ & $-0.0010\pm0.0002$ & 1.000 & $0.0068\pm0.0002$ \\
\bottomrule
\end{tabular}}
\end{table}

}

{
\subsection{Context-Fitted Linear References}
\label{app:lagged_copy_fitted_baselines}
Lag-selected OLS fits $Y_t=a+bX_{t-\ell}$ for
$\ell\in\{0,\ldots,64\}$ and an intercept-only candidate.
All candidates use response times $65,\ldots,T$.
Blocked folds fit 65--256 and validate 257--384, then fit 65--384
and validate 385--512. Selection minimizes equally weighted fold MSE;
ties within relative tolerance $10^{-12}$ with a unit floor favor
intercept-only, then smaller lag. The selected model is refitted on
all available response times.
When its source value is unobserved, the history-only route uses an
AR(1) forecast fitted to source history (slope clipped to $[-0.995,0.995]$);
KC instead uses the supplied continuation.
At $\beta\in\{0.8,1\},L\in\{8,16\}$, OLS selects the true lag
in all 1,280 windows, attains near-oracle MAE, and is invariant to
future resampling. Its no-dependence sensitivity is
$0.0091\pm0.0005$, because finite-context lag selection can choose
spurious relationships.

SparseDirect uses the grid and rolling selection in
Appendix~\ref{app:structural_sparse_direct}.
Its all-channel KC candidate appends the full future source path to
historical features; target-only candidates do not use it.
Validation continuations remain inside history. History-only and KC
fits are selected separately, with the KC fit fixed across resampling.
Its exact-copy sensitivity is zero, but its noisy-copy sensitivity
is positive (Table~\ref{tab:irrelevant_future_main}).
In the no-dependence control ($\beta=0,L=8$), SparseDirect's
$S_{\mathrm{future}}$ is $0.0380\pm0.0011$, with history-only and KC
MAEs of $0.8150\pm0.0064$ and $0.8166\pm0.0064$, respectively.
Only $t_0$ has a higher mean sensitivity among the TSFMs.
With relevant future covariates ($\beta=0.8$), SparseDirect's KC MAEs
are $0.4901\pm0.0040$ at $L=0$ and $0.4926\pm0.0033$ at $L=4$.
These summaries use five-seed means and SEMs.
Separate SparseDirect CI, PastCov, and response-operator measurements
are not available.
}

{
\subsection{Effect Size Relative to Available Improvement}
\label{app:attainable_recovery}

The effect-size references differ across the lagged-copy, event-response,
and shared-latent tasks. The comparisons below retain these distinctions
rather than normalizing every diagnostic by an unavailable positive gain.

\paragraph{Lagged-copy recovery.}
For $\beta=0.8$ and $L\in\{8,16\}$, let $\overline R_M^{\mathrm{CI}}$
and $\overline R_M^{\mathrm{MV}}$ be each model's target-only and
history-informed MAEs, averaged over the five matched seed means.
Let $\overline R^\star$ be the empirical MAE of the generating-process
conditional mean $\beta X_{t-L}$ on those same target windows.
We define own-CI recovery as
\begin{equation}
 A_M^{\mathrm{own}}
 =100\frac{\overline R_M^{\mathrm{CI}}-\overline R_M^{\mathrm{MV}}}
 {\overline R_M^{\mathrm{CI}}-\overline R^\star}.
\end{equation}
These are ratios of seed-averaged risks, not averages of seed-specific
ratios. They compare information use within the same model and retain
negative values. Chronos-2 recovers substantial fractions at both lags,
while the other four TSFMs recover less than $2\%$.

SparseDirect has no evaluated target-only CI route, so its own-CI
recovery is undefined. To compare every method with one common
reference, let $\overline R_0$ be the empirical MAE of predicting zero
for the target. Zero is the target-only conditional median for this
independent, centered Gaussian process. We additionally report
\begin{equation}
 A_M^{0}
 =100\frac{\overline R_0-\overline R_M^{\mathrm{MV}}}
 {\overline R_0-\overline R^\star}.
\end{equation}
Here the available reduction is about $40\%$ of $\overline R_0$.
SparseDirect recovers about $97\%$ under this explicitly shared-zero
reference; this is not a SparseDirect own-CI measurement.
The oracle entries are finite-window empirical errors, not the population
value $\sqrt{2/\pi}\sqrt{1-\beta^2}$.
Table~\ref{tab:rebuttal_attainable_lagged} provides both denominators and
absolute errors so that the two summaries are not conflated.
\begin{table}[p]
\centering\footnotesize
\caption{Lagged-copy recovery at $\beta=0.8$, using actual five-seed mean risks on matched windows. Own-CI recovery and common-zero recovery use different denominators, defined in the text. The zero-forecast target-only risks are 0.79994 ($L=8$), 0.80126 ($L=16$). SparseDirect has no measured CI route, so its own-CI recovery is not reported.}
\label{tab:rebuttal_attainable_lagged}
\setlength{\tabcolsep}{4pt}
\resizebox{\linewidth}{!}{%
\begin{tabular}{llrrrrr}
\toprule
$L$ & Method & CI MAE & History MAE & Oracle MAE & Own-CI (\%) & Common-zero (\%) \\
\midrule
8 & Chronos-2 & 0.80136 & 0.53951 & 0.48184 & 81.95 & 81.87 \\
8 & TimesFM-3 & 0.80147 & 0.79646 & 0.48184 & 1.57 & 1.10 \\
8 & TiRex-2 & 0.80124 & 0.79920 & 0.48184 & 0.64 & 0.24 \\
8 & $t_0$ & 0.80264 & 0.80223 & 0.48184 & 0.13 & -0.72 \\
8 & CITRAS-FM & 0.80397 & 0.80051 & 0.48184 & 1.07 & -0.18 \\
8 & SparseDirect & --- & 0.49261 & 0.48184 & --- & 96.62 \\
\midrule
16 & Chronos-2 & 0.80270 & 0.61471 & 0.48184 & 58.59 & 58.40 \\
16 & TimesFM-3 & 0.80229 & 0.79901 & 0.48184 & 1.02 & 0.70 \\
16 & TiRex-2 & 0.80298 & 0.80215 & 0.48184 & 0.26 & -0.28 \\
16 & $t_0$ & 0.80368 & 0.80422 & 0.48184 & -0.17 & -0.93 \\
16 & CITRAS-FM & 0.80505 & 0.80280 & 0.48184 & 0.70 & -0.48 \\
16 & SparseDirect & --- & 0.49113 & 0.48184 & --- & 97.09 \\
\bottomrule
\end{tabular}}
\end{table}

\paragraph{Event-induced responses.}
Projection gain already uses the full injected response as its unit
reference: $G=1$ means unit signed projection along that template.
Table~\ref{tab:rebuttal_attainable_event} therefore reports $100G$,
with the context-fitted event-response regression alongside the TSFMs.
This is a response-functional normalization, not a fraction of MAE
improvement. Unit gain does not require an exact response profile,
and zero gain does not establish an absent response.
\begin{table}[p]
\centering\footnotesize
\caption{Event-response signed gain expressed as percent of the unit-gain injected-response reference. Entries are mean $\pm$ SEM over five seeds after equal weighting of the nine mappings. A zero projection does not imply a zero response profile.}
\label{tab:rebuttal_attainable_event}
\setlength{\tabcolsep}{4pt}
\resizebox{\linewidth}{!}{%
\begin{tabular}{lrrr}
\toprule
Method & $	\tau=0$ & $	\tau=16$ & $	\tau=64$ \\
\midrule
EventResponseFit & $68.61\pm0.33$ & $71.33\pm0.43$ & $67.00\pm0.42$ \\
Chronos-2 & $46.78\pm0.32$ & $34.18\pm0.20$ & $0.19\pm0.01$ \\
TimesFM-3 & $53.28\pm0.50$ & $3.51\pm0.08$ & $0.50\pm0.02$ \\
TiRex-2 & $1.80\pm0.10$ & $-0.46\pm0.06$ & $-0.11\pm0.01$ \\
$t_0$ & $4.74\pm0.08$ & $0.11\pm0.02$ & $0.03\pm0.01$ \\
CITRAS-FM & $0.20\pm0.00$ & $0.06\pm0.01$ & $-0.04\pm0.00$ \\
\bottomrule
\end{tabular}}
\end{table}

\paragraph{Independent observation channels.}
At $\rho=0.95$, define the paired raw-unit gain
$\overline\Delta_M=\overline R_M(1)-\overline R_M^{\mathrm{ind}}(16)$,
and let $\overline g_M$ be the mean of the five own-reference percentage
gains defined in Appendix~\ref{app:evidence_scaling}.
Table~\ref{tab:rebuttal_attainable_structural} gives the raw improvement,
$\overline g_M$, the fitted-reference comparison
$100\overline g_M/\overline g_{\mathrm{SSM}}$, and the analytic-oracle
raw-gain comparison $100\overline\Delta_M/\overline\Delta_\star$.
The SSM-relative column uses percentage gains with method-specific
baselines; it is not the fraction of an exact-oracle raw MAE gain.
The analytic reference and the fitted SSM are also distinct references.
TimesFM-3 reaches a gain close to the SSM's, whereas the other TSFMs
obtain smaller gains. SparseDirect's gain has substantial seed variation,
so these point ratios do not establish a TSFM shortfall relative to it.
Uncertainty is shown for the paired percentage gains; the ratio columns
are descriptive ratios of summary estimates.
\begin{table}[p]
\centering\footnotesize
\caption{Independent $K=16$ observations at $\rho=0.95$. Relative gains use each method\textquotesingle s own noisy-$K=1$ seed reference. The SSM column is the ratio of mean percentage gains; the oracle column uses raw MAE improvements instead. These distinct summaries are not interchangeable. Only the relative-gain column carries the paired five-seed SEM.}
\label{tab:rebuttal_attainable_structural}
\setlength{\tabcolsep}{4pt}
\resizebox{\linewidth}{!}{%
\begin{tabular}{lrrrr}
\toprule
Method & $\Delta$ MAE & Relative gain (\%) & SSM-relative (\%) & Oracle raw-gain (\%) \\
\midrule
Oracle & 0.02402 & $3.151\pm0.216$ & 111.0 & 100.0 \\
SSM & 0.02174 & $2.838\pm0.313$ & 100.0 & 90.5 \\
Chronos-2 & 0.00605 & $0.735\pm0.709$ & 25.9 & 25.2 \\
TimesFM-3 & 0.02043 & $2.598\pm0.461$ & 91.6 & 85.1 \\
TiRex-2 & 0.00255 & $0.317\pm0.269$ & 11.2 & 10.6 \\
$t_0$ & 0.00736 & $0.934\pm0.329$ & 32.9 & 30.6 \\
CITRAS-FM & 0.00439 & $0.557\pm0.065$ & 19.6 & 18.3 \\
SparseDirect & 0.00889 & $1.073\pm0.933$ & 37.8 & 37.0 \\
\bottomrule
\end{tabular}}
\end{table}

\paragraph{Zero-information controls.}
Duplicates, distractors, and conditionally irrelevant future values add
no information beyond the corresponding reference input. Their ideal
additional gain is zero, so an attainable-improvement recovery ratio
has a zero denominator and is not defined. We instead report signed
MAE changes or forecast sensitivity, with absolute accuracy and the
context-fitted reference, as in the main comparisons.
}

\clearpage
\section{Forecast-Time Covariate Reliability Shift}
\label{app:reliability_experiments}
{
\subsection{Predictive Process and Matched Degradation}
\label{app:reliability_shift}
The clean covariate provides information about the target's future
through a shared dynamic state:
\begin{align}
    z_t &= \rho z_{t-1}+\sqrt{1-\rho^2}\,u_t, \\
    X_t &= (z_t+\sigma_x\epsilon_t^x)/\sqrt{1+\sigma_x^2}, \\
    Y_t &= \kappa z_t+\sigma_y\epsilon_t^y.
\end{align}
The innovations are mutually independent standard Gaussian white
noise. We use $\rho=0.95$, $\kappa=1$, $\sigma_x=0.15$,
$\sigma_y=1/\sqrt{2}$, and 512 burn-in steps.
Each of five seeds ($0$--$4$) supplies 256 windows with a
512-step context and an eight-step forecast horizon.
Holding the target realization fixed, we degrade the covariate as
\begin{equation}
    X_\lambda=\lambda X+\sqrt{1-\lambda^2}\,\xi,
    \qquad \lambda\in\{1,0.75,0.5,0.25,0\},
    \label{eq:kc_reliability_degradation}
\end{equation}
where $\xi$ is independent and has the same marginal variance and
temporal dependence as $X$.
This preserves within-segment variance and autocorrelation;
one-sided degradation can also change cross-boundary dependence,
so the intervention represents a reliability mismatch rather than
a change in dependence strength alone.

The three known-covariate conditions are
\textsc{Stationary}: $(X_\lambda^-,X_\lambda^+)$,
\textsc{Future-Only}: $(X^-,X_\lambda^+)$, and
\textsc{History-Only}: $(X_\lambda^-,X^+)$, where $-$ and $+$ denote
historical and future segments.
Stationary and Future-Only share the same degraded future
realization, and all three inputs coincide at $\lambda=1$.
With the unchanged target history omitted, the shift penalty is
\begin{equation}
    \Delta_{\mathrm{shift}}(\lambda)
    =R_M(X^-,X_\lambda^+)-R_M(X_\lambda^-,X_\lambda^+),
    \label{eq:reliability_shift_penalty}
\end{equation}
where $R_M$ is model MAE.
A positive penalty means that the same degraded future is more
harmful after a clean history; this behavioral contrast does not
identify an internal reliability estimate.
We average paired window differences within each generation seed
and report their mean and SEM across five seeds.
Excess-over-VAR penalties subtract the VAR contrast within seed
before aggregation. Absolute nominal 80\% interval coverage and
width use the same five-seed mean and SEM; interval contrasts are
reported as mean Future-Only minus Stationary differences.
}

{
\subsection{Fitted Joint Conditional References}
\label{app:fitted_conditional_controls}
\paragraph{Joint state-space reference.}
The Gaussian SSM uses the history preprocessing and
stationary-initial-law EM procedure in
Appendix~\ref{app:structural_fitted_baselines}.
Its standardized joint state follows
$\mathbf s_t=r\mathbf s_{t-1}+\mathbf u_t$ with full innovation
covariance $Q$ and observation noise
$R=\operatorname{diag}(R_y,0)$.
The persistence, $Q$, and $R_y$ are fitted to each observed context;
forecast covariance includes filtered-state uncertainty, future
innovations, and target observation noise.

\paragraph{Lag-selected VAR.}
History-only validation selects among 144 combinations of lag
order $\{1,4,8,16,32,64\}$, penalty $\{10^{-4},10^{-2},1,100\}$,
L1 ratio $\{0,0.5,1\}$, and joint or target-independent structure.
The latter removes cross-channel lag coefficients in both
directions and sets innovation cross-covariance to zero.
Each training prefix uses channel means and population standard
deviations (scale one for constant channels), squared loss divided
by $2n$, an Elastic Net penalty, and an unpenalized intercept.
Synthetic-data prefixes end at $\lfloor0.625T\rfloor$ and
$\lfloor0.875T\rfloor$; each fold scores up to 25 chronological
$H=8$ validation origins in raw-unit target MAE.
Coefficients and scalers remain fixed within a fold while later
origins use newly observed history.
Ties within $10^{-12}\max(1,|\mathrm{best\ loss}|)$ favor shorter
lags, target-independent structure, larger penalty, and smaller
L1 ratio, in that order.
The selected model is refitted on the full context.
Coordinate descent uses tolerance $10^{-6}$ and at most 10,000
iterations, retaining finite outputs at the limit.
Innovation covariance uses 0.05 shrinkage toward its diagonal and
a $10^{-8}$ eigenvalue floor in standardized units;
fitted dynamics are not clipped for stability.

\paragraph{Conditional forecasts.}
Both methods form a joint future Gaussian for $(Y^+,X^+)$ from
history, then condition on the supplied trajectory $x^+$:
\begin{align*}
\mu_{Y\mid x}&=\mu_Y+\Sigma_{YX}\Sigma_{XX}^{-1}(x^+-\mu_X),\\
\Sigma_{Y\mid x}&=\Sigma_{YY}-\Sigma_{YX}\Sigma_{XX}^{-1}\Sigma_{XY}.
\end{align*}
The standardized conditioning block uses a $10^{-8}$ eigenvalue
floor and $10^{-10}$ roundoff tolerance; this numerical solve does
not add source observation noise.
Neither generating parameters nor degradation labels are supplied.
Fits and scalers use history only and stay fixed when future
values change; different histories are fitted separately, and the
clean endpoints share a fit. The CI reference fits target history
alone. Gaussian marginal quantiles give the synthetic intervals.
Real-data validation geometry is specified in
Appendix~\ref{app:forecast_reuse_experiments}.
}

{
\subsection{Accuracy and Interval Behavior}
\label{app:known_future_reliability_details}
Table~\ref{tab:ade_reliability_mae_tsfm} retains the reliability
ladder for all five TSFMs and both fitted conditional references.
The fitted SSM and VAR also incur a mismatch penalty;
CITRAS-FM is less affected and has negative mean penalties at some
intermediate levels. Excess over VAR is a paired comparison,
rather than a causal decomposition of error.

Table~\ref{tab:ade_reliability_intervals_tsfm} gives absolute
interval results at $\lambda=0$ and coverage/width contrasts at
all four degraded levels.
For Chronos-2, TimesFM-3, TiRex-2, and $t_0$, Future-Only intervals
are narrower and have lower coverage than Stationary intervals
at every degraded level. CITRAS-FM's width changes little, and its
coverage contrast varies with reliability.

\begin{table}[t]
\centering\footnotesize
\setlength{\tabcolsep}{4pt}
\caption{Reliability-shift penalties, $\Delta_{\mathrm{shift}}=\mathrm{MAE}_{\mathrm{Future\mbox{-}Only}}-\mathrm{MAE}_{\mathrm{Stationary}}$, across all degraded reliability levels. Values are mean $\pm$ SEM of five paired generation-seed means. The last column subtracts the VAR penalty within each seed at $\lambda=0$. At $\lambda=1$, all routes coincide and both contrasts are zero. The seven-method $\lambda=0$ route MAEs and CI/Clean references appear in Table~\ref{tab:kc_reliability_shift}.}
\label{tab:ade_reliability_mae_tsfm}
\label{tab:ade_reliability_mae_fitted}
\resizebox{\linewidth}{!}{%
\begin{tabular}{lrrrrr}
\toprule
Model & $\lambda=0.75$ & $\lambda=0.50$ & $\lambda=0.25$ & $\lambda=0$ & Excess over VAR ($\lambda=0$) \\
\midrule
Chronos-2 & $0.0890\pm0.0048$ & $0.2070\pm0.0088$ & $0.3320\pm0.0142$ & $0.4594\pm0.0200$ & $0.0827\pm0.0128$ \\
TimesFM-3 & $0.1072\pm0.0067$ & $0.2266\pm0.0089$ & $0.3491\pm0.0135$ & $0.4752\pm0.0180$ & $0.0985\pm0.0094$ \\
TiRex-2 & $0.0982\pm0.0052$ & $0.2058\pm0.0084$ & $0.3175\pm0.0130$ & $0.4351\pm0.0156$ & $0.0584\pm0.0094$ \\
$t_0$ & $0.0559\pm0.0045$ & $0.1410\pm0.0088$ & $0.2398\pm0.0150$ & $0.3481\pm0.0206$ & $-0.0285\pm0.0157$ \\
CITRAS-FM & $-0.0208\pm0.0023$ & $-0.0034\pm0.0048$ & $0.0291\pm0.0074$ & $0.0647\pm0.0092$ & $-0.3120\pm0.0187$ \\
SSM & $0.1000\pm0.0070$ & $0.2110\pm0.0087$ & $0.3247\pm0.0111$ & $0.4435\pm0.0153$ & $0.0669\pm0.0085$ \\
VAR & $0.0418\pm0.0073$ & $0.1462\pm0.0122$ & $0.2616\pm0.0176$ & $0.3766\pm0.0199$ & $0.0000\pm0.0000$ \\
\bottomrule
\end{tabular}%
}
\end{table}

\begin{table}[t]
\centering\footnotesize
\setlength{\tabcolsep}{4pt}
\caption{Nominal 80\% interval behavior under Stationary (S) and Future-Only (F) degradation. Top: absolute endpoint coverage (\%) and width, mean $\pm$ SEM across five generation-seed means. Bottom: mean F minus S contrasts at every degraded reliability level; coverage differences are percentage points and width differences are in target units. Negative values indicate lower coverage or narrower intervals after a clean history. At $\lambda=1$, the matched inputs coincide and all contrasts are zero.}
\label{tab:ade_reliability_intervals_tsfm}
\label{tab:ade_reliability_intervals_fitted}
\begin{tabular}{lrrrr}
\toprule
\multicolumn{5}{c}{Absolute intervals at $\lambda=0$} \\
Model & Coverage S & Coverage F & Width S & Width F \\
\midrule
Chronos-2 & $79.01\pm0.40$ & $45.21\pm0.97$ & $2.524\pm0.006$ & $1.879\pm0.006$ \\
TimesFM-3 & $79.70\pm0.27$ & $45.18\pm0.54$ & $2.541\pm0.004$ & $1.950\pm0.004$ \\
TiRex-2 & $78.62\pm0.45$ & $45.25\pm0.64$ & $2.484\pm0.004$ & $1.898\pm0.004$ \\
$t_0$ & $80.07\pm0.27$ & $54.30\pm0.74$ & $2.580\pm0.004$ & $2.139\pm0.006$ \\
CITRAS-FM & $79.48\pm0.34$ & $75.81\pm0.85$ & $2.538\pm0.005$ & $2.549\pm0.004$ \\
SSM & $78.76\pm0.51$ & $45.09\pm0.83$ & $2.485\pm0.001$ & $1.860\pm0.003$ \\
VAR & $75.40\pm0.61$ & $47.66\pm0.76$ & $2.414\pm0.008$ & $1.879\pm0.004$ \\
\bottomrule
\end{tabular}
\par\medskip
\begin{tabular}{lrrrrrrrr}
\toprule
& \multicolumn{2}{c}{$\lambda=0.75$} & \multicolumn{2}{c}{$\lambda=0.50$} & \multicolumn{2}{c}{$\lambda=0.25$} & \multicolumn{2}{c}{$\lambda=0$} \\
\cmidrule(lr){2-3}\cmidrule(lr){4-5}\cmidrule(lr){6-7}\cmidrule(lr){8-9}
Model & $\Delta$Cov. & $\Delta$Width & $\Delta$Cov. & $\Delta$Width & $\Delta$Cov. & $\Delta$Width & $\Delta$Cov. & $\Delta$Width \\
\midrule
Chronos-2 & $-16.44$ & $-0.471$ & $-24.25$ & $-0.607$ & $-29.91$ & $-0.642$ & $-33.80$ & $-0.645$ \\
TimesFM-3 & $-15.92$ & $-0.397$ & $-24.72$ & $-0.561$ & $-30.46$ & $-0.606$ & $-34.52$ & $-0.591$ \\
TiRex-2 & $-13.35$ & $-0.294$ & $-21.79$ & $-0.442$ & $-28.51$ & $-0.554$ & $-33.37$ & $-0.586$ \\
$t_0$ & $-9.49$ & $-0.259$ & $-16.04$ & $-0.386$ & $-21.90$ & $-0.440$ & $-25.77$ & $-0.441$ \\
CITRAS-FM & $+0.47$ & $-0.018$ & $-0.85$ & $-0.028$ & $-2.51$ & $-0.016$ & $-3.67$ & $+0.010$ \\
SSM & $-14.31$ & $-0.342$ & $-22.83$ & $-0.508$ & $-29.04$ & $-0.597$ & $-33.67$ & $-0.625$ \\
VAR & $-11.33$ & $-0.382$ & $-18.44$ & $-0.484$ & $-23.80$ & $-0.527$ & $-27.74$ & $-0.535$ \\
\bottomrule
\end{tabular}
\end{table}

}

\clearpage
\section{Conditioning on Predicted Covariates}
\label{app:forecast_reuse_experiments}
{
\subsection{Datasets and Evaluation Protocol}
\label{app:forecast_reuse_data}

We select three paired-channel tasks from fev-bench
\citep{shchur2026fevbenchrealisticbenchmarktime}: temperature$\to$load
(GFC17), wind speed$\to$active power (KDD), and customer count$\to$sales
(Rossmann). GFC17 represents weather-dependent demand forecasting
\citep{hong2019gefcom2017}, where prior work propagates temperature scenarios
to load forecasts \citep{dimoulkas2019neural}. KDD provides turbine-level
measurements from a challenge requiring forecasts without future meteorological
observations \citep{zhou2022sdwpf}. Rossmann extends the analysis to retail
\citep{rossmann2015}, where customer count is a historical covariate in the
benchmark task. We retain benchmark horizons and rolling origins, restrict
inputs to the selected source $X$ and target $Y$, and cap context at 512.
Primary forecasts withhold realized future sources, including the temperature
continuation exposed in the original GFC17 task; other covariates are excluded.

\paragraph{Cohorts and scoring.}
The primary five-model routes share the same series--origin pairs, restricted
to at least one finite target value and complete future source observations:
160 GFC17, 1,339 KDD, and 10,250 Rossmann pairs. Realized source values enter
only the named True-$X$ diagnostic. The supplementary native-quantile comparison
uses the target-eligible populations of 160, 1,340, and 10,610 pairs instead.
Rossmann excludes 540 scheduled pairs without finite target labels, then
360 more under source completeness; KDD loses one under source completeness.
Horizons are 168, 96, and 48, and seasonal periods are 24, 48, and seven,
respectively.

All primary target distributions use the common deciles
$\mathcal A=\{0.1,0.2,\ldots,0.9\}$ and the reconstruction below.
SQL averages twice the pinball loss over deciles and valid horizon positions,
divided by the mean absolute seasonal difference in the raw capped target
history. The denominator is identical across routes in a case; no epsilon
is added to an undefined scale. Coverage uses the nominal $80\%$ interval,
and width is its mean endpoint difference. We average valid horizon positions,
then origins within each series, then series equally within a dataset.
These diagnostic SQL scores differ from official benchmark scoring;
we do not attach synthetic-seed SEMs to dependent rolling-origin scores
or pool raw units across tasks.

\subsection{Point Accuracy and Predictive Intervals}
\label{app:forecast_reuse_results}

Conditioning on a predicted source median increases SQL and lowers coverage
relative to direct MV in every five-model, three-dataset comparison in
Table~\ref{tab:forecast_reuse_sql_coverage}. On KDD, it lowers point MAE
for all five models while narrowing intervals and lowering coverage.
Table~\ref{tab:f_point_intervals} retains the matched point and width evidence.
\begin{table}[t]
\centering\footnotesize
\setlength{\tabcolsep}{3pt}
\caption{KDD point accuracy and interval width under MV$\to$Median conditioning for all five models. Each entry gives direct MV $\to$ predicted-source median KC, on the 1,339 complete-source pairs with endpoint-clamped target reconstruction. MAE and nominal $80\%$ interval width are in target units. Coverage is reported in Table~\ref{tab:forecast_reuse_sql_coverage}.}
\label{tab:f_point_intervals}
\resizebox{\linewidth}{!}{%
\begin{tabular}{@{}lrr@{}}
\toprule
Model & MAE: MV$\to$Median & Width: MV$\to$Median \\
\midrule
Chronos-2 & $251.19\to 249.11$ & $788.76\to 261.65$ \\
TimesFM-3 & $243.83\to 239.40$ & $728.09\to 108.33$ \\
TiRex-2 & $211.68\to 205.29$ & $619.92\to 71.44$ \\
$t_0$ & $207.43\to 202.42$ & $758.10\to 107.01$ \\
CITRAS-FM & $227.83\to 222.99$ & $435.88\to 196.66$ \\
\bottomrule
\end{tabular}%
}
\end{table}

\subsection{Trajectory Propagation and Diagnostics}
\label{app:quantile_path_propagation}

\paragraph{Sampled source trajectories.}
The primary method draws 100 source paths from MV forecast marginals with a
Gaussian copula
\citep{baron2025efficientlygeneratingcorrelatedsample,choi2026interweavingmarginalsmultivariatesample}.
Source deciles are monotonically rearranged and interpolated linearly.
Below and above the decile grid, respectively, the quantile function is
\begin{align}
 Q_h(u)&=q_{1,h}-(q_{2,h}-q_{1,h})\log_2(0.1/u), &&u<0.1,\\
 Q_h(u)&=q_{9,h}+(q_{9,h}-q_{8,h})\log_2\{0.1/(1-u)\}, &&u>0.9,
\end{align}
with probabilities clipped to $[10^{-6},1-10^{-6}]$.
Finite source-history values receive average ranks and normal scores
$\Phi^{-1}\{\mathrm{rank}/(n+1)\}$. Adjacent valid normal-score pairs estimate
$\widehat\rho$, clipped to $[-0.99,0.99]$. Gaussian innovations generate
\begin{equation}
 Z_1=\epsilon_1,\qquad
 Z_h=\widehat\rho Z_{h-1}+\sqrt{1-\widehat\rho^2}\epsilon_h,
 \quad \epsilon_h\stackrel{\mathrm{iid}}{\sim}\mathcal N(0,1),
\end{equation}
and each source path is $X_h=Q_h\{\Phi(Z_h)\}$, with case-specific fixed seeds.
Each path is supplied separately to the same frozen KC interface with
unchanged observed histories. This estimates a source trajectory law;
it does not assume a common joint law for the model's source and conditional
target forecasts.

\paragraph{Target distributions and mixing.}
At every horizon step, we sort the nine conditional target deciles,
interpolate linearly between them, and clamp outside $[0.1,0.9]$.
Endpoint clamping assigns mass $0.1$ to each endpoint, with possible further
atoms from ties. For $B$ equally weighted source paths, we invert the
mixture CDF rather than average component quantiles:
\begin{equation}
 \widetilde F_{Y,h}(y\mid\mathcal I)
 =\frac1B\sum_{b=1}^{B}\widehat F^{(b)}_{Y,h}(y\mid\mathcal I),\qquad
 \widetilde Q_{Y,h}(p\mid\mathcal I)
 =\inf\{y:\widetilde F_{Y,h}(y\mid\mathcal I)\ge p\}.
 \label{eq:quantile_path_mixture}
\end{equation}
The mixture median supplies point MAE. Direct MV and Median use the same
target sorting and reconstruction as Trajectory.

\paragraph{Fixed-rank paths, tails, and realized sources.}
Rank9 concatenates each native source decile over the horizon to form nine
equally weighted paths of fixed rank, retaining possible native source
quantile crossings. Its prescribed temporal coupling differs from the
history-estimated copula. The target-tail diagnostic adds
$Q(0)=2q_1-q_2$ and $Q(1)=2q_9-q_8$ and interpolates linearly through the tails,
keeping the source paths and conditional predictions fixed.
True-$X$ instead supplies complete realized future sources to KC; it is
neither a deployable route nor a guaranteed upper bound.
Table~\ref{tab:quantile_path_true_source} retains these three diagnostics
on the primary complete-source cohorts. Trajectory improves SQL and coverage
relative to Median in every primary comparison, but coverage generally remains
below direct MV and the nominal level, with remaining model and task differences.
For coverage $C$, define recovery of the additional loss from Median as
$r_C=(C_{\rm Trajectory}-C_{\rm Median})/(C_{\rm MV}-C_{\rm Median})$.
Across the 15 primary model--dataset pairs, median recovery is $77.2\%$
(range $6.2$--$101.6\%$), with a median remaining MV-coverage gap of
$4.6$ percentage points. This measures recovery relative to direct MV,
rather than closeness to nominal $80\%$ coverage.
\begin{table}[t]
\centering\footnotesize
\setlength{\tabcolsep}{3pt}
\caption{Path diagnostics on the primary complete-source cohorts (160 GFC17, 1,339 KDD, 10,250 Rossmann pairs). Rank9 uses nine fixed native source ranks; Tail keeps the primary 100 temporal source paths and changes only target reconstruction to linear tail extension. True-$X$ supplies realized future sources. Rank9 and True-$X$ use endpoint clamping; the primary Trajectory comparison is in Table~\ref{tab:forecast_reuse_sql_coverage}. Coverage is in percent.}
\label{tab:quantile_path_true_source}
\resizebox{\linewidth}{!}{%
\begin{tabular}{@{}llrrrrrr@{}}
\toprule
Dataset & Model & \multicolumn{2}{c}{Rank9} & \multicolumn{2}{c}{Trajectory + Tail} & \multicolumn{2}{c}{True-$X$} \\
\cmidrule(lr){3-4}\cmidrule(lr){5-6}\cmidrule(lr){7-8}
 &  & SQL & Cov. & SQL & Cov. & SQL & Cov. \\
\midrule
GFC17 & Chronos-2 & 0.998 & 59.6 & 1.029 & 68.0 & 0.619 & 79.9 \\
GFC17 & TimesFM-3 & 0.954 & 64.2 & 0.950 & 69.0 & 0.648 & 80.3 \\
GFC17 & TiRex-2 & 0.933 & 62.9 & 0.927 & 68.5 & 0.753 & 75.0 \\
GFC17 & $t_0$ & 0.972 & 60.9 & 0.971 & 68.2 & 0.696 & 78.6 \\
GFC17 & CITRAS-FM & 1.011 & 40.7 & 0.985 & 45.4 & 0.815 & 52.0 \\
\addlinespace
KDD & Chronos-2 & 0.863 & 68.7 & 0.887 & 66.8 & 0.184 & 85.3 \\
KDD & TimesFM-3 & 0.813 & 65.7 & 0.799 & 73.0 & 0.150 & 81.7 \\
KDD & TiRex-2 & 0.737 & 72.1 & 0.728 & 80.3 & 0.142 & 90.4 \\
KDD & $t_0$ & 0.721 & 58.2 & 0.720 & 70.2 & 0.146 & 80.5 \\
KDD & CITRAS-FM & 0.815 & 42.6 & 0.795 & 45.9 & 0.372 & 60.1 \\
\addlinespace
Rossmann & Chronos-2 & 0.470 & 65.5 & 0.469 & 74.4 & 0.161 & 76.1 \\
Rossmann & TimesFM-3 & 0.497 & 72.6 & 0.497 & 78.6 & 0.149 & 78.3 \\
Rossmann & TiRex-2 & 0.523 & 77.4 & 0.524 & 78.4 & 0.194 & 87.7 \\
Rossmann & $t_0$ & 0.485 & 73.4 & 0.485 & 80.6 & 0.162 & 72.0 \\
Rossmann & CITRAS-FM & 0.534 & 58.5 & 0.532 & 60.0 & 0.485 & 66.9 \\
\bottomrule
\end{tabular}%
}
\end{table}

\subsection{Fitted Conditional Forecasters}
\label{app:f_conditional_controls}

SSM and VAR fitting follows Appendix~\ref{app:fitted_conditional_controls},
using the same supplied histories and complete-source cohorts.
Real-data VAR validation uses training prefixes $0.5T$ and $0.625T$,
the actual task horizon, and up to 25 complete origins per fold, all within
the observed history. The selected model is refitted on that history;
supplied-future routes do not refit it.
Their Joint100 route samples 100 entire source trajectories from the fitted
joint Gaussian, preserving temporal covariance, and mixes the conditional
target CDFs with the common decile reconstruction. This differs from the TSFM
marginal-copula construction. Under the fitted Gaussian law, conditioning on
the source mean leaves the target mean unchanged and reduces covariance;
propagation restores source-associated variation. The implementation checks
mean equality and total covariance.
Both controls show worse SQL and coverage under Median, whereas Joint100
brings SQL close to MV (Table~\ref{tab:f_conditional_controls}).
\begin{table}[t]
\centering\footnotesize
\setlength{\tabcolsep}{3pt}
\caption{Fitted conditional controls on the primary matched cohorts. SQL changes are absolute differences from MV; coverage changes are percentage points. Joint100 propagates full trajectories from the fitted joint Gaussian, using endpoint-clamped target reconstruction.}
\label{tab:f_conditional_controls}
\resizebox{\linewidth}{!}{%
\begin{tabular}{@{}llrrrrrr@{}}
\toprule
Dataset & Method & SQL$_{\rm MV}$ & $\Delta$SQL Median & $\Delta$SQL Joint100 & Cov.$_{\rm MV}$ & $\Delta$Cov. Median & $\Delta$Cov. Joint100 \\
\midrule
GFC17 & SSM & 1.859 & 0.0341 & 0.0061 & 77.8 & -10.7 & -3.9 \\
GFC17 & VAR & 1.273 & 0.0222 & 0.0044 & 57.9 & -6.8 & -1.5 \\
KDD & SSM & 0.819 & 0.0565 & 0.0035 & 81.5 & -40.3 & -1.2 \\
KDD & VAR & 0.832 & 0.0325 & 0.0018 & 76.7 & -19.3 & -0.6 \\
Rossmann & SSM & 0.982 & 0.1252 & 0.0065 & 74.9 & -46.8 & -0.5 \\
Rossmann & VAR & 0.574 & 0.0226 & 0.0020 & 84.7 & -28.4 & -0.8 \\
\bottomrule
\end{tabular}%
}
\end{table}

\subsection{Historical Source and Target Calibration}
\label{app:f_calibration}

\paragraph{Historical schedule and corrections.}
Calibration uses only each supplied $T\le512$ history.
Set $m=\max[0,\min\{8,\lfloor(T-128)/H\rfloor\}]$ and form complete
historical horizon blocks beginning at $T-mH,\ldots,T-H$, requiring $m\ge2$.
Each historical forecast uses only its preceding history. All but the latest
block fit the source map; the latest is reserved for target correction and
never refits that map. For GFC17, these are blocks after prefixes 176 and 344.
Each fit requires at least 32 finite labels.

Observed source labels yield mid-PIT values $\{F(x^-)+F(x)\}/2$,
retaining ties and atoms. Their empirical higher-quantile inverse map $G^{-1}$
is composed directly as
$Q_{\mathrm{cal}}(u)=Q_{\mathrm{raw}}\{G^{-1}(u)\}$,
without a second nine-knot approximation. Raw and calibrated paths share
copula probabilities. For each source state and target reconstruction,
the complete pipeline runs on the reserved block. From its intervals $[l,u]$
and labels $y$, compute $s=\max\{0,l-y,y-u\}$ and its empirical higher
$80$th percentile $c$. After final mixture inversion, replace only endpoints
by $q_{0.1}-c$ and $q_{0.9}+c$; interior quantiles and the median remain unchanged.
The endpoints are not remixed. The same target correction applies directly
to MV as a control. This historical correction provides no prospective coverage
guarantee under dependence or changing reliability.

\paragraph{Temporal-coupling comparisons.}
Tables~\ref{tab:f_calibration_temporal} and
\ref{tab:f_calibration_temporal_tail} report the four source/target calibration
states under the primary temporal coupling, for default endpoint clamping
and linear target tails. Target corrections are fitted separately for each
reconstruction; source maps and scenario probabilities are unchanged.
Within each row the four states share a calibration-eligible cohort:
160 GFC17 and 1,339 KDD pairs, and 10,070 Rossmann pairs after the reserved
block's label threshold. CITRAS-FM/KDD and $t_0$, CITRAS-FM/Rossmann are
unreported. Source-only Rossmann retains the 10,250-pair primary cohort
(Table~\ref{tab:f_source_only}); GFC17 and KDD coincide with the first two
states in Table~\ref{tab:f_calibration_temporal}. Source-only effects on target
SQL and coverage are mixed.

The following summary uses raw sources and endpoint clamping.
For Chronos-2, TimesFM-3, and TiRex-2 across the three tasks,
target correction reduces the equal-comparison mean absolute deviation
from nominal $80\%$ coverage from $8.4$ to $3.4$ percentage points,
with improvement in eight of nine comparisons. SQL increases range
from $0.08\%$ to $1.01\%$ in these comparisons.
The TiRex-2/KDD exception overshoots the nominal level; increased
coverage is not itself evidence of improved calibration.
Applying the correction directly to MV also increases coverage but raises SQL
in 11 of 12 reported comparisons (Table~\ref{tab:f_direct_ycal}).
\begin{table}[t]
\centering\footnotesize
\setlength{\tabcolsep}{3pt}
\caption{Temporal-coupling calibration with endpoint-clamped target reconstruction. Each cell is SQL / coverage (\%). R/C denotes raw/calibrated sources (first letter) and raw/corrected target intervals (second). Within each row all four states use the same $n$ calibration-eligible pairs. CITRAS-FM/KDD and $t_0$, CITRAS-FM/Rossmann are unreported and omitted.}
\label{tab:f_calibration_temporal}
\resizebox{\linewidth}{!}{%
\begin{tabular}{@{}llrcccc@{}}
\toprule
Dataset & Model & $n$ & R/R & C/R & R/C & C/C \\
\midrule
GFC17 & Chronos-2 & 160 & $1.032\,/\,65.7$ & $1.069\,/\,58.3$ & $1.035\,/\,71.9$ & $1.065\,/\,71.9$ \\
GFC17 & TimesFM-3 & 160 & $0.953\,/\,66.5$ & $1.010\,/\,60.6$ & $0.959\,/\,74.7$ & $1.008\,/\,73.4$ \\
GFC17 & TiRex-2 & 160 & $0.930\,/\,66.0$ & $0.967\,/\,61.0$ & $0.931\,/\,74.8$ & $0.962\,/\,73.7$ \\
GFC17 & $t_0$ & 160 & $0.974\,/\,65.1$ & $0.993\,/\,63.3$ & $0.979\,/\,73.7$ & $0.996\,/\,73.0$ \\
GFC17 & CITRAS-FM & 160 & $0.989\,/\,43.0$ & $1.000\,/\,46.2$ & $0.969\,/\,71.2$ & $0.978\,/\,73.3$ \\
\addlinespace
KDD & Chronos-2 & 1339 & $0.888\,/\,65.2$ & $0.886\,/\,65.1$ & $0.896\,/\,76.5$ & $0.898\,/\,78.6$ \\
KDD & TimesFM-3 & 1339 & $0.799\,/\,72.5$ & $0.777\,/\,76.4$ & $0.805\,/\,80.7$ & $0.788\,/\,82.4$ \\
KDD & TiRex-2 & 1339 & $0.728\,/\,79.8$ & $0.747\,/\,83.5$ & $0.735\,/\,85.6$ & $0.756\,/\,87.2$ \\
KDD & $t_0$ & 1339 & $0.720\,/\,68.5$ & $0.762\,/\,66.8$ & $0.726\,/\,81.9$ & $0.772\,/\,81.9$ \\
\addlinespace
Rossmann & Chronos-2 & 10070 & $0.476\,/\,73.6$ & $0.475\,/\,72.2$ & $0.478\,/\,80.1$ & $0.476\,/\,80.0$ \\
Rossmann & TimesFM-3 & 10070 & $0.505\,/\,77.8$ & $0.509\,/\,72.1$ & $0.506\,/\,81.4$ & $0.510\,/\,79.9$ \\
Rossmann & TiRex-2 & 10070 & $0.524\,/\,77.3$ & $0.532\,/\,75.0$ & $0.527\,/\,80.7$ & $0.535\,/\,80.4$ \\
\bottomrule
\end{tabular}%
}
\end{table}

\begin{table}[t]
\centering\footnotesize
\setlength{\tabcolsep}{3pt}
\caption{Temporal-coupling calibration with linear target-tail reconstruction; target corrections are fitted separately for this convention. Each cell is SQL / coverage (\%). R/C denotes raw/calibrated sources (first letter) and raw/corrected target intervals (second). Within each row all four states use the same $n$ calibration-eligible pairs. CITRAS-FM/KDD and $t_0$, CITRAS-FM/Rossmann are unreported and omitted.}
\label{tab:f_calibration_temporal_tail}
\resizebox{\linewidth}{!}{%
\begin{tabular}{@{}llrcccc@{}}
\toprule
Dataset & Model & $n$ & R/R & C/R & R/C & C/C \\
\midrule
GFC17 & Chronos-2 & 160 & $1.029\,/\,68.0$ & $1.064\,/\,61.0$ & $1.033\,/\,73.0$ & $1.063\,/\,72.5$ \\
GFC17 & TimesFM-3 & 160 & $0.950\,/\,69.0$ & $1.007\,/\,63.1$ & $0.957\,/\,75.6$ & $1.005\,/\,74.0$ \\
GFC17 & TiRex-2 & 160 & $0.927\,/\,68.5$ & $0.963\,/\,63.6$ & $0.929\,/\,76.1$ & $0.960\,/\,74.9$ \\
GFC17 & $t_0$ & 160 & $0.971\,/\,68.2$ & $0.989\,/\,66.1$ & $0.977\,/\,75.0$ & $0.994\,/\,74.4$ \\
GFC17 & CITRAS-FM & 160 & $0.985\,/\,45.4$ & $0.996\,/\,48.5$ & $0.968\,/\,71.2$ & $0.977\,/\,73.5$ \\
\addlinespace
KDD & Chronos-2 & 1339 & $0.887\,/\,66.8$ & $0.885\,/\,66.7$ & $0.895\,/\,77.0$ & $0.897\,/\,79.1$ \\
KDD & TimesFM-3 & 1339 & $0.799\,/\,73.0$ & $0.776\,/\,76.8$ & $0.804\,/\,80.9$ & $0.787\,/\,82.6$ \\
KDD & TiRex-2 & 1339 & $0.728\,/\,80.3$ & $0.747\,/\,83.9$ & $0.735\,/\,85.8$ & $0.756\,/\,87.4$ \\
KDD & $t_0$ & 1339 & $0.720\,/\,70.2$ & $0.761\,/\,68.3$ & $0.726\,/\,82.6$ & $0.771\,/\,82.4$ \\
\addlinespace
Rossmann & Chronos-2 & 10070 & $0.476\,/\,74.1$ & $0.475\,/\,72.9$ & $0.478\,/\,80.3$ & $0.476\,/\,80.2$ \\
Rossmann & TimesFM-3 & 10070 & $0.505\,/\,78.1$ & $0.509\,/\,72.7$ & $0.506\,/\,81.5$ & $0.509\,/\,80.0$ \\
Rossmann & TiRex-2 & 10070 & $0.524\,/\,77.9$ & $0.532\,/\,75.9$ & $0.527\,/\,81.0$ & $0.535\,/\,80.7$ \\
\bottomrule
\end{tabular}%
}
\end{table}

\begin{table}[t]
\centering\footnotesize
\setlength{\tabcolsep}{3pt}
\caption{Source-only recalibration on the 10,250-pair primary Rossmann cohort, without the reserved-target-block eligibility restriction. Cells are target SQL / coverage (\%); temporal coupling, endpoint clamping, and no target correction are used. GFC17 and KDD source-only comparisons coincide with R/R and C/R in Table~\ref{tab:f_calibration_temporal}. Unreported models follow that table.}
\label{tab:f_source_only}
\resizebox{\linewidth}{!}{%
\begin{tabular}{@{}lcc@{}}
\toprule
Model & Raw source & Calibrated source \\
\midrule
Chronos-2 & $0.469\,/\,73.9$ & $0.468\,/\,72.3$ \\
TimesFM-3 & $0.497\,/\,78.2$ & $0.501\,/\,72.4$ \\
TiRex-2 & $0.524\,/\,77.8$ & $0.532\,/\,75.1$ \\
\bottomrule
\end{tabular}%
}
\end{table}

\begin{table}[t]
\centering\footnotesize
\setlength{\tabcolsep}{3pt}
\caption{Direct MV $\to$ direct MV with the same historical target-interval correction, using the common calibration cohorts and endpoint-clamped reconstruction. Coverage is in percent; $n$ is the number of evaluated series--origin pairs. Unreported models follow Table~\ref{tab:f_calibration_temporal}.}
\label{tab:f_direct_ycal}
\resizebox{\linewidth}{!}{%
\begin{tabular}{@{}llrrr@{}}
\toprule
Dataset & Model & $n$ & SQL: raw$\to$corrected & Cov.: raw$\to$corrected \\
\midrule
GFC17 & Chronos-2 & 160 & $0.968\to 0.973$ & $69.4\to 74.4$ \\
GFC17 & TimesFM-3 & 160 & $0.954\to 0.960$ & $71.2\to 77.0$ \\
GFC17 & TiRex-2 & 160 & $0.930\to 0.935$ & $71.4\to 77.8$ \\
GFC17 & $t_0$ & 160 & $0.957\to 0.964$ & $70.3\to 76.4$ \\
GFC17 & CITRAS-FM & 160 & $0.990\to 0.972$ & $46.2\to 70.9$ \\
\addlinespace
KDD & Chronos-2 & 1339 & $0.828\to 0.832$ & $71.9\to 79.5$ \\
KDD & TimesFM-3 & 1339 & $0.807\to 0.812$ & $71.8\to 80.0$ \\
KDD & TiRex-2 & 1339 & $0.735\to 0.742$ & $84.1\to 85.1$ \\
KDD & $t_0$ & 1339 & $0.732\to 0.737$ & $79.9\to 85.3$ \\
\addlinespace
Rossmann & Chronos-2 & 10070 & $0.466\to 0.468$ & $76.2\to 81.3$ \\
Rossmann & TimesFM-3 & 10070 & $0.496\to 0.498$ & $82.6\to 84.0$ \\
Rossmann & TiRex-2 & 10070 & $0.520\to 0.522$ & $81.3\to 83.1$ \\
\bottomrule
\end{tabular}%
}
\end{table}

\subsection{Native-Quantile CI-Source and Source-Quality Comparisons}
\label{app:f_native_comparisons}

Table~\ref{tab:forecast_reuse_native_routes} preserves the native-quantile
CI-source comparison on the larger target-eligible populations for
Chronos-2, TimesFM-3, and TiRex-2. Predicting the source from its own history
and supplying its median to KC also worsens SQL and coverage relative to
native MV. These scores use native target quantiles rather than the primary
common reconstruction and therefore remain a separate comparison.
The same table reports native MV source SQL and coverage using available
source truth and its own seasonal scale. Imperfect source coverage means
that the target discrepancy cannot be assigned solely to source uncertainty
or solely to the KC interface.
\begin{table}[t]
\centering\footnotesize
\setlength{\tabcolsep}{3pt}
\caption{Supplementary native-quantile evidence for Chronos-2, TimesFM-3, and TiRex-2. Target comparisons use 160/1,340/10,610 target-eligible GFC17/KDD/Rossmann pairs, unlike the primary complete-source cohorts. CI$\to$KC predicts the source from its own history, then supplies its median to KC. Source SQL and coverage score native MV source forecasts using available source truth and the source seasonal scale; their source mask does not restrict the target cohort. Coverage is in percent.}
\label{tab:forecast_reuse_native_routes}
\resizebox{\linewidth}{!}{%
\begin{tabular}{@{}llrrrrrr@{}}
\toprule
Dataset & Model & \multicolumn{2}{c}{Target SQL} & \multicolumn{2}{c}{Target coverage} & \multicolumn{2}{c}{MV source quality} \\
\cmidrule(lr){3-4}\cmidrule(lr){5-6}\cmidrule(lr){7-8}
 &  & MV & CI$\to$KC & MV & CI$\to$KC & SQL & Cov. \\
\midrule
GFC17 & Chronos-2 & 0.968 & 1.027 & 69.4 & 48.6 & 1.079 & 68.9 \\
GFC17 & TimesFM-3 & 0.954 & 0.977 & 71.2 & 55.7 & 1.028 & 68.2 \\
GFC17 & TiRex-2 & 0.930 & 0.943 & 71.4 & 60.8 & 1.053 & 70.6 \\
KDD & Chronos-2 & 0.828 & 0.911 & 71.7 & 46.0 & 0.805 & 73.0 \\
KDD & TimesFM-3 & 0.806 & 0.863 & 71.7 & 27.8 & 0.756 & 72.9 \\
KDD & TiRex-2 & 0.735 & 0.834 & 84.0 & 30.3 & 0.692 & 77.7 \\
Rossmann & Chronos-2 & 0.492 & 0.556 & 76.1 & 36.4 & 0.542 & 78.7 \\
Rossmann & TimesFM-3 & 0.528 & 0.597 & 82.3 & 36.1 & 0.570 & 83.4 \\
Rossmann & TiRex-2 & 0.570 & 0.614 & 79.7 & 46.6 & 0.635 & 82.6 \\
\bottomrule
\end{tabular}%
}
\end{table}

}

\clearpage
\section{Representation Diagnostics}
\label{app:representation_experiments}
These experiments assess linearly accessible information in frozen target-side
representations. They are separate from the behavioral sweeps and do not by
themselves identify the mechanisms behind forecast changes.
\subsection{Representation Diagnostics for Distributed Evidence}
\label{app:distributed_diagnostics}
\paragraph{Data and comparison conditions.}
We use the shared-latent process described in Appendix~\ref{app:evidence_scaling}, with $\rho=0.95$ and $\lambda=0.75$. This separate experiment uses seeds $0,1,2$ and 256 windows per seed.
The main diagnostic fixes the total number of auxiliary channels at $D=16$ and varies
\begin{equation}
    K_{\mathrm{info}}\in\{1,2,4,8,16\}.
\end{equation}
The remaining $D-K_{\mathrm{info}}$ channels are matched distractors.
All gains in Table~\ref{tab:distributed_representation} compare $K_{\mathrm{info}}=16$ with $K_{\mathrm{info}}=1$ at this fixed dimensionality.
Unlike the channel-count sweep in Section~\ref{sec:distributed_underutilization}, the baseline therefore contains one informative channel and 15 distractors.

We additionally examine the original duplicate and distractor configurations as their total auxiliary-channel count increases from $K=1$ to $K=16$.

\paragraph{Target-side latent-state probes.}
For each layer $\ell$, we extract the hidden representation $h_Y^{(\ell)}$ associated with the scored target channel.
At the final layer, this is the representation consumed by the forecasting head.
We fit a linear ridge probe
\begin{equation}
    \widehat{z}_T
    =
    W h_Y^{(\ell)} + b
\end{equation}
to recover the latent state at the forecast origin.
The probe never receives auxiliary-channel representations directly.
For each model, we retain one target token without pooling:
Chronos-2 uses zero-based token 33 (the future patch), TiRex-2 token 63
(the token producing the first requested forecast steps), and TimesFM-3
token 15 (the last context patch). At the head input, the vectors have
dimensions 768, 512, and 1,280, respectively.
Block outputs and the input to the forecast head are extracted separately,
including any transformation between the final block and the head.

Evaluation uses leave-one-seed-out splits.
Feature normalization, regularization selection, and probe fitting use training seeds only.
Each outer split has 512 training windows and 256 held-out windows.
The ridge probe includes an intercept, uses feature means and standard deviations
estimated from the training windows, and centers $z_T$ by its training mean.
Inner leave-one-training-seed-out validation chooses the regularization coefficient
from $\alpha\in\{10^{j/2}:j=-8,-7,\ldots,16\}$ by mean validation MSE;
the probe is then refitted on both training seeds.

Probe gains are reported as mean $\pm$ SEM over three held-out-seed folds; forecast gains are reported as mean $\pm$ SEM over three seeds.

\paragraph{Oracle reference.}
We compute the exact Gaussian posterior mean
\begin{equation}
    \mu_K^\star
    =
    \mathbb{E}
    \left[
        z_T
        \mid
        Y_{1:T},
        X_{1,1:T},\ldots,X_{K,1:T}
    \right].
\end{equation}
For independent informative channels, the population oracle $R^2$ increases from $0.889$ at $K=1$ to $0.971$ at $K=16$.
Duplicate and distractor channels add no target-relevant information, so their oracle $R^2$ remains unchanged.

Oracle and probe $\Delta R^2$ values are not interpreted as directly comparable fractions of captured information.
The oracle is an exact conditional expectation, whereas the probe is an out-of-sample linear estimator fitted to a pretrained representation.
Likewise, probe $R^2$ and forecast MAE assess different quantities and are not compared on a common numerical scale.

\paragraph{Representation and forecast changes.}
Table~\ref{tab:distributed_representation} reports changes when
informative channels increase from 1 to 16 at a fixed total of
$D=16$ auxiliary channels.
Latent-state probe $R^2$ improves in all three models, with gains
ranging from $0.0806$ to $0.1042$.
Forecast MAE reductions range from $0.0086$ to $0.0149$, with all
point estimates below the corresponding oracle gain of $0.0219$.
This comparison combines increased predictive evidence with reduced
distraction, so it does not isolate the contribution of either change.
Improved linear accessibility of $z_T$ accompanies forecast
improvement.

\begin{table}[t]
    \centering
    \caption{
    Representation and forecast gains when informative channels
    increase from 1 to 16 at a fixed total of $D=16$ auxiliary channels.
    $\Delta R_z^2$ measures improved target-side latent-state recovery;
    $\Delta_M$ measures forecast MAE reduction.
    Oracle gains are $\Delta R_z^2=0.0825$ and
    $\Delta^\star=0.0219$.
    Values are mean $\pm$ SEM over three held-out-seed folds
    for probes and three seeds for forecasts.
    }
    \label{tab:distributed_representation}
    \small
    \setlength{\tabcolsep}{10pt}
    \renewcommand{\arraystretch}{1.1}
    \begin{tabular}{lcc}
        \toprule
        Model
        & Probe gain $\Delta R_z^2$
        & Forecast gain $\Delta_M$ \\
        \midrule
        Chronos-2
        & $0.1042 \pm 0.0015$
        & $0.0142 \pm 0.0065$ \\
        TimesFM-3
        & $0.1009 \pm 0.0087$
        & $0.0149 \pm 0.0038$ \\
        TiRex-2
        & $0.0806 \pm 0.0062$
        & $0.0086 \pm 0.0024$ \\
        \bottomrule
    \end{tabular}
\end{table}

\paragraph{Duplicate and distractor controls.}
For Chronos-2, increasing distractors from $K=1$ to $K=16$ reduces
head-input probe $R^2$ from $0.791$ to $0.768$ and increases MAE
by $0.0163$.
In the duplicate condition, latent-state decodability remains
nearly unchanged while MAE increases by $0.0082$.
\subsection{Target-Side Lag-Class Diagnostics}
\label{app:lag_class_diagnostics}

\paragraph{Data generation.}
We generate $X_t\overset{\mathrm{iid}}{\sim}\mathcal{N}(0,1)$ and, for each lag class $L$, construct
\begin{equation}
    Y_t=\beta X_{t-L}+\sqrt{1-\beta^2}\,\epsilon_t,
\end{equation}
where $\epsilon_t\overset{\mathrm{iid}}{\sim}\mathcal{N}(0,1)$ is independent of $X$.
For the no-dependence class, we instead use
\begin{equation}
    Y_t=\beta X'_t+\sqrt{1-\beta^2}\,\epsilon_t,
\end{equation}
with $X'$ an independent standard Gaussian series.
We set $\beta=0.8$, use 512 context steps, and request an eight-step forecast from each model.
Forecasts are used to check the extraction path, not to score predictive accuracy in this experiment.

For each base realization, all classes share the same $X$, $\epsilon$, and $X'$ arrays; lag classes differ only in the alignment of $X$ used to generate $Y$.
A fixed 64-step source prefix supports every lag without changing the supplied $X$ history across classes.
The predefined lag grids are
\begin{align}
    \mathcal{L}_{\mathrm{primary}}&=\{0,5,10,15,20\},\\
    \mathcal{L}_{\mathrm{fine}}&=\{0,1,2,4,8\},\\
    \mathcal{L}_{\mathrm{coarse}}&=\{0,8,16,32,64\}.
\end{align}
Each grid includes the no-dependence class, giving six balanced classes and chance accuracy $1/6$.
We use five seeds with 128 base realizations per seed, yielding 768 windows per seed and 3,840 windows per grid.
Results are evaluated separately for each grid; the grids are not treated as independent replications because they share base realizations and overlapping classes.

\paragraph{Input conditions and representation extraction.}
Under \textsc{MV}, the model receives $[X^-,Y^-]$; under \textsc{CI}, it receives the identical $Y^-$ history alone.
Neither condition receives observed future covariate values.
All pretrained checkpoints remain frozen.
At each block and at the forecasting-head input, we retain only the target row and one token, with no pooling or concatenation of covariate rows.
The selected zero-based token positions are 33 for Chronos-2 (the future-patch token), 63 for TiRex-2 (the token producing the first requested forecast steps), and 15 for TimesFM-3 (the last context patch).
The corresponding hidden dimensions are 768, 512, and 1,280.
At the head input, we use the representation after any intervening final transformation, rather than substituting the last block output.
The reported extraction check finds Spearman correlation one between the head median path at the selected token and the returned median forecast for every window in each model, mode, and grid.

\paragraph{Probe training and evaluation.}
We standardize features using training data and fit a multinomial logistic regression with an L2 penalty, using the \texttt{lbfgs} solver and a maximum of 2,000 iterations.
Outer evaluation holds out one generation seed at a time.
All class variants of a base realization remain in the same split.
Within the four training seeds, inner leave-one-seed-out validation selects
\begin{equation}
    C\in\{10^{-4},10^{-3},10^{-2},10^{-1},1,10,100\}
\end{equation}
by mean validation accuracy, breaking ties in favor of the smaller $C$.
The selected probe is refitted on all four training seeds and evaluated once on the held-out seed.
We report mean accuracy with SEM over the five outer folds.
Paired MV--CI differences are computed within each held-out seed before aggregation.

\paragraph{Controls.}
The target-only model representations provide a \textsc{CI} control.
Additional probes use raw $X^-$, raw $Y^-$, or a fixed random nonlinear map of the concatenated histories:
\begin{equation}
    \phi(X^-,Y^-)
    =\tanh\!\left([X^-;Y^-]W/\sqrt{1024}\right),
\end{equation}
where the entries of $W$ are independently standard Gaussian and its output width matches the respective model representation.
All controls use the same splitting, normalization, and regularization-selection procedure.
Low accuracy for these particular probes is not evidence that the raw joint histories lack lag information.

\paragraph{Results and interpretation.}
\label{app:known_future_probe_details}
Table~\ref{tab:lag_class_all_grids} summarizes head-input accuracy
and paired MV--CI differences across grids.
Raw and random-feature controls, coarse-grid class recall, and
confusion counts are retained in
Tables~\ref{tab:lag_class_controls},
\ref{tab:lag_class_coarse_recall}, and
\ref{tab:lag_confusion_coarse}.
For Chronos-2, coarse-grid recall decreases at longer lags,
with confusion between long-lag and no-dependence classes.
This concerns the fitted linear probe, not the absence of
information in the full representation.

\begin{figure}[t]
    \centering
    \includegraphics[width=\linewidth]{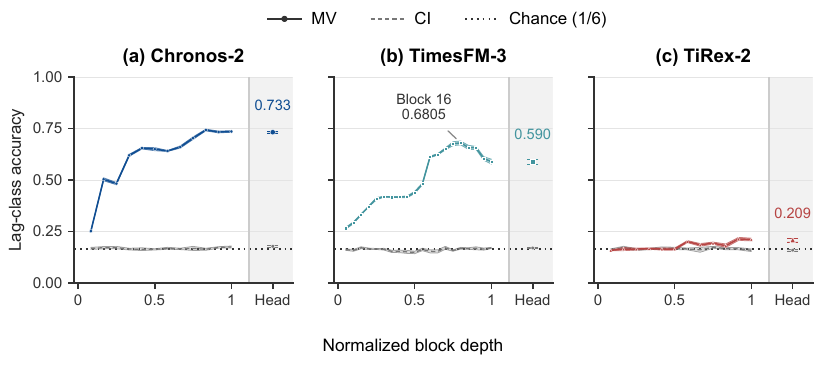}
    \caption{
    Layer-wise linear decodability of cross-channel lag relationships.
    Each layer is evaluated with a separate probe using only the target token
    at the forecast-readout position, under multivariate (MV) or target-only (CI) input.
    The coarse grid contains $L\in\{0,8,16,32,64\}$ and a no-dependence class;
    chance accuracy is $1/6$.
    Curves show mean accuracy with $\pm$ SEM over five held-out-seed folds.
    Block depth is normalized by the number of blocks (12, 20, and 12).
    The shaded \emph{Head} region reports the actual head input separately,
    with SEM error bars; it does not represent an additional block.
    }
    \label{fig:lag_layer_decodability}
\end{figure}

Across depth, TimesFM-3's coarse-grid accuracy peaks at block 16
before decreasing toward the head input, whereas Chronos-2
largely maintains its late-layer accuracy
(Figure~\ref{fig:lag_layer_decodability}).

These diagnostics assess linear accessibility of lag classes,
not recovery of predictive source values or their causal use
by the forecast head.
The probe experiment uses the same lagged Gaussian family as
Section~\ref{sec:irrelevant_future_covariates}, but separate
realizations.

\begin{table}[t]
    \centering
    \caption{Head-input lag-class accuracy across six-class grids (chance $1/6$). Values are mean $\pm$ SEM over five held-out-seed folds; MV--CI uses paired differences. Grids share base realizations and are not independent replications.}
    \label{tab:lag_class_all_grids}
    \label{tab:lag_class_paired}
    \footnotesize
    \setlength{\tabcolsep}{5pt}
    \renewcommand{\arraystretch}{1.05}
    \begin{tabular}{lccc}
        \toprule
        Model & MV & CI & MV--CI \\
        \midrule
        \multicolumn{4}{l}{\textbf{Primary}: $\{0,5,10,15,20\}$, plus no dependence} \\
        Chronos-2 & $0.9612 \pm 0.0042$ & $0.1693 \pm 0.0056$ & $0.7919 \pm 0.0091$ \\
        TimesFM-3 & $0.5398 \pm 0.0032$ & $0.1651 \pm 0.005$ & $0.3747 \pm 0.0046$ \\
        TiRex-2 & $0.1974 \pm 0.005$ & $0.1674 \pm 0.0048$ & $0.02995 \pm 0.0072$ \\
        \midrule
        \multicolumn{4}{l}{\textbf{Fine}: $\{0,1,2,4,8\}$, plus no dependence} \\
        Chronos-2 & $0.9359 \pm 0.0052$ & $0.1664 \pm 0.0082$ & $0.7695 \pm 0.0078$ \\
        TimesFM-3 & $0.5552 \pm 0.0075$ & $0.1581 \pm 0.0025$ & $0.3971 \pm 0.0087$ \\
        TiRex-2 & $0.1958 \pm 0.0069$ & $0.1714 \pm 0.0065$ & $0.02448 \pm 0.012$ \\
        \midrule
        \multicolumn{4}{l}{\textbf{Coarse}: $\{0,8,16,32,64\}$, plus no dependence} \\
        Chronos-2 & $0.7326 \pm 0.0032$ & $0.1781 \pm 0.0062$ & $0.5544 \pm 0.0045$ \\
        TimesFM-3 & $0.5896 \pm 0.012$ & $0.1701 \pm 0.0031$ & $0.4195 \pm 0.015$ \\
        TiRex-2 & $0.2091 \pm 0.0091$ & $0.1594 \pm 0.0032$ & $0.04974 \pm 0.0093$ \\
        \bottomrule
    \end{tabular}
\end{table}

\begin{table}[t]
    \centering
    \caption{Raw-history and fixed random-feature control accuracies (mean $\pm$ SEM over five folds). Random features use both histories, with output widths matching Chronos-2 (768), TimesFM-3 (1,280), and TiRex-2 (512). Chance accuracy is $1/6$.}
    \label{tab:lag_class_controls}
    \footnotesize
    \setlength{\tabcolsep}{4pt}
    \renewcommand{\arraystretch}{1.05}
    \begin{tabular}{lrccc}
        \toprule
        Probe input & Width & Primary & Fine & Coarse \\
        \midrule
        Raw $X^-$ & 512 & $0.1667 \pm 0$ & $0.1667 \pm 0$ & $0.1667 \pm 0$ \\
        Raw $Y^-$ & 512 & $0.1747 \pm 0.006$ & $0.1737 \pm 0.0043$ & $0.1669 \pm 0.0043$ \\
        \midrule
        Random features & 768 & $0.1622 \pm 0.0036$ & $0.1628 \pm 0.0034$ & $0.1672 \pm 0.0062$ \\
        Random features & 1,280 & $0.1651 \pm 0.0062$ & $0.168 \pm 0.007$ & $0.1609 \pm 0.0053$ \\
        Random features & 512 & $0.1674 \pm 0.0057$ & $0.1643 \pm 0.0064$ & $0.1682 \pm 0.0043$ \\
        \bottomrule
    \end{tabular}
\end{table}

\begin{table}[t]
    \centering
    \caption{Per-class recall at the head input for the coarse grid, aggregated over the held-out predictions. None denotes no cross-channel dependence. These are class-specific recalls, not six-class accuracies.}
    \label{tab:lag_class_coarse_recall}
    \footnotesize
    \setlength{\tabcolsep}{4pt}
    \renewcommand{\arraystretch}{1.05}
    \begin{tabular}{llrrrrrr}
        \toprule
        Model & Input & $L=0$ & $L=8$ & $L=16$ & $L=32$ & $L=64$ & None \\
        \midrule
        Chronos-2 & MV & 0.997 & 0.973 & 0.936 & 0.605 & 0.419 & 0.466 \\
         & CI & 0.217 & 0.211 & 0.128 & 0.113 & 0.183 & 0.217 \\
        \addlinespace[3pt]
        TimesFM-3 & MV & 0.989 & 0.498 & 0.469 & 0.534 & 0.48 & 0.567 \\
         & CI & 0.145 & 0.186 & 0.178 & 0.161 & 0.158 & 0.192 \\
        \addlinespace[3pt]
        TiRex-2 & MV & 0.223 & 0.216 & 0.206 & 0.178 & 0.227 & 0.205 \\
         & CI & 0.133 & 0.173 & 0.152 & 0.159 & 0.155 & 0.184 \\
        \bottomrule
    \end{tabular}
\end{table}

\begin{table}[t]
    \centering
    \caption{Coarse-grid head-input confusion matrices under MV input. Rows are true classes and columns are predicted classes. Counts are summed across five held-out-seed folds, giving 640 examples per true class. None denotes no cross-channel dependence. Bold entries are correct predictions.}
    \label{tab:lag_confusion_coarse}
    \footnotesize
    \setlength{\tabcolsep}{6pt}
    \begin{tabular}{lrrrrrr}
        \toprule
        True / predicted & $L=0$ & $L=8$ & $L=16$ & $L=32$ & $L=64$ & None \\
        \midrule
        \multicolumn{7}{l}{\textbf{Chronos-2}} \\
        $L=0$ & \textbf{638} & 0 & 0 & 0 & 0 & 2 \\
        $L=8$ & 1 & \textbf{623} & 9 & 4 & 1 & 2 \\
        $L=16$ & 0 & 2 & \textbf{599} & 36 & 2 & 1 \\
        $L=32$ & 0 & 1 & 15 & \textbf{387} & 131 & 106 \\
        $L=64$ & 1 & 0 & 3 & 116 & \textbf{268} & 252 \\
        None & 1 & 1 & 2 & 85 & 253 & \textbf{298} \\
        \midrule
        \multicolumn{7}{l}{\textbf{TimesFM-3}} \\
        $L=0$ & \textbf{633} & 2 & 1 & 0 & 2 & 2 \\
        $L=8$ & 2 & \textbf{319} & 126 & 46 & 64 & 83 \\
        $L=16$ & 1 & 133 & \textbf{300} & 102 & 62 & 42 \\
        $L=32$ & 0 & 66 & 85 & \textbf{342} & 98 & 49 \\
        $L=64$ & 5 & 55 & 41 & 74 & \textbf{307} & 158 \\
        None & 5 & 57 & 39 & 45 & 131 & \textbf{363} \\
        \midrule
        \multicolumn{7}{l}{\textbf{TiRex-2}} \\
        $L=0$ & \textbf{143} & 127 & 93 & 77 & 109 & 91 \\
        $L=8$ & 105 & \textbf{138} & 125 & 89 & 95 & 88 \\
        $L=16$ & 113 & 109 & \textbf{132} & 108 & 91 & 87 \\
        $L=32$ & 98 & 90 & 109 & \textbf{114} & 117 & 112 \\
        $L=64$ & 85 & 94 & 77 & 111 & \textbf{145} & 128 \\
        None & 98 & 98 & 83 & 103 & 127 & \textbf{131} \\
        \bottomrule
    \end{tabular}
\end{table}

{
\subsection{Event Identity and Predictive Source-Value Probes}
\label{app:rebuttal_probe_results}
The following probes distinguish recognizing an event from recovering the source
values needed to forecast a target. They use frozen Chronos-2, TimesFM-3, and
TiRex-2 checkpoints and fit diagnostic readouts on separate synthetic training
seeds. They complement the shared-latent and lag-class diagnostics above;
they do not identify a causal mechanism of forecast behavior.

\subsubsection{Historical Event Identity at the Response-Peak Representation}
\label{app:event_response_probes}

\paragraph{Data and representations.}
We use the spike-trigger, positive-response instance of the delayed-event
generator in Appendix~\ref{app:temporal_experiments}, with historical test-event
lags $\tau\in\{32,40,48,64\}$, a 768-step context, and a 96-step forecast.
This probe population covers this single trigger--response pair, rather than
the full set of trigger and response shapes in the behavioral sweep.
The response starts at forecast index 24 and peaks at index 29 (zero based).
The source and response amplitudes are three; the source background noise,
target noise, and target-level standard deviations are $0.3$, $0.10$, and
$0.15$, respectively. Each history contains nine teaching events and nine
historical distractors, with spacing 76 and jitter eight.
For each lag and each of five generation seeds, we generate 300 paired windows.
Within a window, event, no-event, and distractor arms share the background,
target history, and teaching examples; only the source pattern in the test
slot differs. The distractor is the downward counterpart of the upward trigger.
The native known-future interface supplies the source history and continuation;
the test event whose identity is probed is in the history.

For each arm, we retain the source-row token containing the test trigger and,
separately, the target-row token feeding the forecast head at the expected
response peak. We use one token for each readout, without pooling or
concatenating source and target representations. Token positions are derived
from each loaded model's patch and head configuration and checked against the
captured tensor shape. The reported representations are the actual head inputs,
including any intervening final normalization. A target-history-only reference
omits the covariate and assigns the three arm labels to the identical target
representation of each window.

\paragraph{Fitting and controls.}
We fit a training-set StandardScaler followed by multinomial logistic regression
with an L2 penalty, using \texttt{lbfgs}, a 2,000-iteration budget, and tolerance
$10^{-4}$. Outer evaluation holds out one generation seed. All three arms of a
window remain in the same split. Inner leave-one-training-seed-out validation
selects $C\in\{10^{-4},10^{-3},10^{-2},10^{-1},1,10,100\}$ by mean validation
accuracy, breaking ties toward smaller $C$; the selected readout is refitted on
the four training seeds. The event--distractor AUROC uses
$\log p(\mathrm{event})-\log p(\mathrm{distractor})$ from this same three-class
readout, restricted to the 600 event and distractor examples in the held-out
seed. The full test fold contains 900 examples.
The shuffled-label control repeats selection and fitting with labels permuted
within each training seed, with one permutation per outer fold, and evaluates
against the true held-out labels. All readouts use the same solver budget;
finite outputs are retained when this budget is reached.
Table~\ref{tab:probe_rebuttal_event_head} reports means and SEMs across the five
held-out seed folds. These summarize variation over generated datasets;
the training sets overlap between folds.

\paragraph{Results.}
Historical event identity remains linearly detectable in the target-side
head input at these lags. At lag 64, target event--distractor AUROC is
$0.928\pm0.001$ for Chronos-2 and $0.992\pm0.001$ for TimesFM-3, despite the
weaker delayed forecast responses in the behavioral experiment.
TiRex-2's target AUROC decreases with lag but remains above the shuffled
reference. The source-token scores remain high, whereas the target-history-only
reference has chance AUROC by construction.
This establishes linear accessibility of event identity at the selected target
token. It does not show that the native forecast head uses that information
with the required amplitude or timing.
}

\begin{table}[t]
\centering
\caption{Event--distractor AUROC from frozen representations in the delayed-event probe. Source and target denote the covariate token containing the historical test trigger and the target token feeding the response-peak forecast, respectively, both at the head input. The shuffled control permutes training labels for the target probe. Target-only attaches the three labels to identical target-history-only representations. Values are mean $\pm$ SEM over five held-out generation-seed folds.}
\label{tab:probe_rebuttal_event_head}
\small
\setlength{\tabcolsep}{4pt}
\begin{tabular}{llcccc}
\toprule
Model & Lag $\tau$ & Source & Target & Target shuffled & Target-only \\
\midrule
Chronos-2 & 32 & $1.0000 \pm 0.0000$ & $0.9955 \pm 0.0007$ & $0.4987 \pm 0.0102$ & $0.5000 \pm 0.0000$ \\
Chronos-2 & 40 & $1.0000 \pm 0.0000$ & $0.9889 \pm 0.0013$ & $0.4940 \pm 0.0071$ & $0.5000 \pm 0.0000$ \\
Chronos-2 & 48 & $1.0000 \pm 0.0000$ & $0.9650 \pm 0.0037$ & $0.5055 \pm 0.0044$ & $0.5000 \pm 0.0000$ \\
Chronos-2 & 64 & $1.0000 \pm 0.0000$ & $0.9285 \pm 0.0012$ & $0.5095 \pm 0.0049$ & $0.5000 \pm 0.0000$ \\
\midrule
TimesFM-3 & 32 & $1.0000 \pm 0.0000$ & $0.9994 \pm 0.0001$ & $0.4883 \pm 0.0107$ & $0.5000 \pm 0.0000$ \\
TimesFM-3 & 40 & $1.0000 \pm 0.0000$ & $0.9985 \pm 0.0002$ & $0.4910 \pm 0.0061$ & $0.5000 \pm 0.0000$ \\
TimesFM-3 & 48 & $1.0000 \pm 0.0000$ & $0.9941 \pm 0.0004$ & $0.5081 \pm 0.0066$ & $0.5000 \pm 0.0000$ \\
TimesFM-3 & 64 & $1.0000 \pm 0.0000$ & $0.9917 \pm 0.0013$ & $0.5042 \pm 0.0125$ & $0.5000 \pm 0.0000$ \\
\midrule
TiRex-2 & 32 & $1.0000 \pm 0.0000$ & $0.9454 \pm 0.0035$ & $0.5079 \pm 0.0038$ & $0.5000 \pm 0.0000$ \\
TiRex-2 & 40 & $1.0000 \pm 0.0000$ & $0.9255 \pm 0.0058$ & $0.4791 \pm 0.0121$ & $0.5000 \pm 0.0000$ \\
TiRex-2 & 48 & $1.0000 \pm 0.0000$ & $0.9132 \pm 0.0037$ & $0.5094 \pm 0.0070$ & $0.5000 \pm 0.0000$ \\
TiRex-2 & 64 & $1.0000 \pm 0.0000$ & $0.7399 \pm 0.0048$ & $0.4945 \pm 0.0057$ & $0.5000 \pm 0.0000$ \\
\bottomrule
\end{tabular}
\end{table}

{
\subsubsection{Decoding the Predictive Source Values}
\label{app:predictive_source_value_probes}

\paragraph{Data and interfaces.}
We generate independent standard Gaussian $X_t$ and $\epsilon_t$ and set
\begin{equation}
    Y_t=\beta X_{t-L}+\sqrt{1-\beta^2}\,\epsilon_t,
    \qquad \beta=0.8,\quad L\in\{8,16\}.
\end{equation}
We use 512 context steps and forecast eight steps, with 128 base histories per
seed and five seeds. A fixed 64-step source prefix permits matched source and
noise draws across lags. These are fresh realizations, separate from the
behavioral lagged-copy evaluation. Each model and lag is evaluated separately
under CI, MV, PastCov, and KC. CI receives only target history; MV supplies both
histories as target channels; PastCov supplies source history through the
past-covariate interface; KC additionally supplies the full source continuation.
Since $L\geq8$, the predictive source vector
\begin{equation}
    s=(X_{T+1-L},\ldots,X_{T+8-L})
\end{equation}
lies entirely in the observed history for both lags, and the conditional target
mean is $m^*=\beta s$.

\paragraph{Head-input extraction.}
We capture only the target row's single forecast-head input token that produces
steps one through eight. The target row is row one under MV and row zero under
the other interfaces (zero based); covariate rows are excluded from the probe.
The selected tokens are Chronos-2's first future-patch token, TiRex-2's token 63,
and TimesFM-3's context patch 15, with feature dimensions 768, 512, and 1,280,
respectively. We replay the loaded head and its native median readout, including
inverse scaling and any differencing, detrending, clipping, or quantile sorting
used by that model. Every window in all 24 model--lag--interface conditions
passes the extraction check
$|\widehat{Y}_{\mathrm{replay}}-\widehat{Y}_{\mathrm{native}}|
\leq10^{-4}+10^{-4}|\widehat{Y}_{\mathrm{native}}|$
at every requested step. The primary probe receives the hidden vector alone,
without source rows, raw source values, or stored normalization statistics.

\paragraph{Readouts and evaluation.}
A multi-output ridge readout maps this target-side vector to $s$.
Features are standardized on each training split and the intercept is
unpenalized. Outer evaluation again holds out one generation seed; the
remaining four seeds provide 512 training histories.
Inner leave-one-training-seed-out validation selects
$\alpha\in\{10^{-4},10^{-3},10^{-2},10^{-1},1,10,100,1000,10000\}$ by
horizon-averaged validation MSE, breaking ties toward larger $\alpha$.
The selected readout is refitted on all four training seeds and tested on
128 histories from the held-out seed. We compute $R^2$ separately at each of
the eight horizons within a test fold, average across horizons, then report
mean and SEM over the five seed folds. Negative $R^2$ values are retained.
A shuffled-source control permutes the complete source vectors within each
training seed and repeats the full selection procedure. A raw-target-history readout uses the same fitting procedure and splits;
a true-source reference returns $s$ directly without fitting.

For each held-out history, we also compare the native median forecast and the
diagnostic forecast $\beta\widehat{s}$ against the same realized target $Y$.
The latter tests whether a supervised linear readout of the frozen
representation can yield useful source estimates; it is not a zero-shot
forecasting procedure. Table~\ref{tab:probe_rebuttal_source_head} reports source
recovery and these paired-population target errors.

\paragraph{Results and scope.}
Under the history-only MV and PastCov interfaces, Chronos-2's target-side
representation supports substantially more accurate source-value recovery
than those of TimesFM-3 and TiRex-2. The corresponding diagnostic forecasts
also improve on Chronos-2's native forecast. CI, raw target history, and
shuffled-source controls do not recover the independent predictive source
values. KC changes the pattern: source-value recovery improves, including
for TiRex-2, so the history-only comparison should not be extended to every
interface.

These readouts distinguish event classification from decoding the values
needed for a forecast. Accessible information need not be used by the native
head, and poor linear decoding does not establish that information is absent
from a nonlinear representation. Neither experiment locates the first stage
at which forecast behavior departs from the constructed relation or establishes
that an internal feature causes the forecast response.
}

\begin{table}[t]
\centering
\caption{Predictive source-value recovery and target forecast error on the same held-out windows. The hidden probe decodes the eight historical source values from one target-side forecast-head input; its diagnostic forecast is $\beta\widehat{s}$. Shuffled uses training-source vectors permuted within each seed. The target MAEs compare the native median and the probe-decoded diagnostic forecast with $Y$. Each cell is mean $\pm$ SEM over five held-out generation-seed folds, with 128 histories per fold. KC supplies the full future covariate path; MV and PastCov supply historical covariates only.}
\label{tab:probe_rebuttal_source_head}
\small
\setlength{\tabcolsep}{3pt}
\begin{tabular}{lllcccc}
\toprule
 & & & \multicolumn{2}{c}{Source $R^2$} & \multicolumn{2}{c}{Target MAE} \\
\cmidrule(lr){4-5}\cmidrule(lr){6-7}
Model & $L$ & Interface & Hidden & Shuffled & Native & $\beta\widehat{s}$ \\
\midrule
Chronos-2 & 8 & CI & $-0.024 \pm 0.005$ & $-0.026 \pm 0.003$ & $0.795 \pm 0.009$ & $0.797 \pm 0.009$ \\
Chronos-2 & 8 & MV & $0.903 \pm 0.006$ & $-0.015 \pm 0.011$ & $0.524 \pm 0.008$ & $0.507 \pm 0.007$ \\
Chronos-2 & 8 & PastCov & $0.903 \pm 0.006$ & $-0.015 \pm 0.011$ & $0.524 \pm 0.008$ & $0.507 \pm 0.007$ \\
Chronos-2 & 8 & KC & $0.938 \pm 0.002$ & $-0.018 \pm 0.014$ & $0.520 \pm 0.005$ & $0.500 \pm 0.005$ \\
\addlinespace
Chronos-2 & 16 & CI & $-0.022 \pm 0.002$ & $-0.023 \pm 0.003$ & $0.811 \pm 0.007$ & $0.813 \pm 0.008$ \\
Chronos-2 & 16 & MV & $0.666 \pm 0.005$ & $-0.020 \pm 0.006$ & $0.602 \pm 0.007$ & $0.593 \pm 0.004$ \\
Chronos-2 & 16 & PastCov & $0.666 \pm 0.005$ & $-0.020 \pm 0.006$ & $0.602 \pm 0.007$ & $0.593 \pm 0.004$ \\
Chronos-2 & 16 & KC & $0.775 \pm 0.004$ & $-0.017 \pm 0.006$ & $0.566 \pm 0.009$ & $0.555 \pm 0.007$ \\
\midrule
TimesFM-3 & 8 & CI & $-0.026 \pm 0.004$ & $-0.031 \pm 0.007$ & $0.795 \pm 0.009$ & $0.797 \pm 0.010$ \\
TimesFM-3 & 8 & MV & $0.075 \pm 0.005$ & $-0.032 \pm 0.008$ & $0.791 \pm 0.009$ & $0.774 \pm 0.008$ \\
TimesFM-3 & 8 & PastCov & $0.075 \pm 0.005$ & $-0.032 \pm 0.008$ & $0.791 \pm 0.009$ & $0.774 \pm 0.008$ \\
TimesFM-3 & 8 & KC & $0.169 \pm 0.017$ & $-0.039 \pm 0.009$ & $0.760 \pm 0.006$ & $0.743 \pm 0.007$ \\
\addlinespace
TimesFM-3 & 16 & CI & $-0.020 \pm 0.004$ & $-0.033 \pm 0.004$ & $0.811 \pm 0.007$ & $0.813 \pm 0.008$ \\
TimesFM-3 & 16 & MV & $0.020 \pm 0.005$ & $-0.033 \pm 0.004$ & $0.808 \pm 0.007$ & $0.802 \pm 0.009$ \\
TimesFM-3 & 16 & PastCov & $0.020 \pm 0.005$ & $-0.033 \pm 0.004$ & $0.808 \pm 0.007$ & $0.802 \pm 0.009$ \\
TimesFM-3 & 16 & KC & $0.076 \pm 0.006$ & $-0.032 \pm 0.005$ & $0.790 \pm 0.008$ & $0.787 \pm 0.011$ \\
\midrule
TiRex-2 & 8 & CI & $-0.024 \pm 0.004$ & $-0.023 \pm 0.004$ & $0.796 \pm 0.009$ & $0.797 \pm 0.009$ \\
TiRex-2 & 8 & MV & $0.002 \pm 0.004$ & $-0.022 \pm 0.004$ & $0.794 \pm 0.009$ & $0.790 \pm 0.009$ \\
TiRex-2 & 8 & PastCov & $0.001 \pm 0.004$ & $-0.022 \pm 0.004$ & $0.794 \pm 0.009$ & $0.791 \pm 0.009$ \\
TiRex-2 & 8 & KC & $0.505 \pm 0.012$ & $-0.026 \pm 0.009$ & $0.663 \pm 0.003$ & $0.635 \pm 0.003$ \\
\addlinespace
TiRex-2 & 16 & CI & $-0.022 \pm 0.004$ & $-0.019 \pm 0.003$ & $0.812 \pm 0.007$ & $0.814 \pm 0.008$ \\
TiRex-2 & 16 & MV & $-0.017 \pm 0.004$ & $-0.019 \pm 0.003$ & $0.811 \pm 0.007$ & $0.813 \pm 0.008$ \\
TiRex-2 & 16 & PastCov & $-0.017 \pm 0.005$ & $-0.019 \pm 0.003$ & $0.811 \pm 0.007$ & $0.813 \pm 0.008$ \\
TiRex-2 & 16 & KC & $0.242 \pm 0.008$ & $-0.018 \pm 0.002$ & $0.739 \pm 0.007$ & $0.740 \pm 0.008$ \\
\bottomrule
\end{tabular}
\end{table}

\clearpage
{\section{Fine-Tuning on Delayed Event--Response Data}
\label{app:temporal_finetuning}

\subsection{Training and Evaluation Protocol}

We evaluate task-specific adaptation of Chronos-2, $t_0$, and
CITRAS-FM. The adaptation task uses a spike trigger and a positive
target response, with a 768-step context, a 96-step horizon, and
nine historical teaching pairs. In variable-lag training
(FT-Variable), each example has a lag in $0$--$48$, and its
historical teaching pairs use the same lag. Trigger positions
are drawn independently of the assigned lag from a common
support. The training construction has no historical distractors.

The primary recipe uses learning rate $10^{-5}$ and 1,000 training
steps, with training seeds 42, 43, and 44. Each training run has
16,000 window exposures; these are repeated exposures, not
16,000 distinct windows. Additional controls use learning rate
$10^{-6}$ and lag-zero training (FT-Local). Frozen and fine-tuned
checkpoints use matched inference and normalization rules within
each evaluation panel and backbone. Native quantile grids and
numerical conventions differ across backbones; a common learning
rate and step count do not imply identical optimization strength.

\paragraph{Evaluation panels and metrics.}
The moving-trigger evaluation covers all 49 integer lags from
$0$ to $48$, with trigger onsets in $0$--$36$. The completed
evaluation uses five generation draws and 64 paired windows per
lag and draw. For extrapolation, we evaluate the same in-range
lags and the unseen lags 56, 64, and 80 with trigger onsets
restricted to $0$--$4$ in both groups. This keeps the complete
response inside the forecast horizon and matches trigger
positions across lag ranges.

We report signed response gain $G$ and normalized full-profile
error $E$ from the paired event and no-event forecasts. The
reference values are $G=1$ and $E=0$; a zero response has error
about one. Gain alone does not establish that the whole response
curve is correct. Moving-trigger metrics are computed per window
before aggregation, with equal weighting over evaluation lags
within each generation draw. They are not profile errors computed
after averaging all windows into one response curve.

For fine-tuned checkpoints, we average the five generation draws
within each training seed and report the mean and sample
standard deviation across the three training seeds. Frozen
uncertainty is the sample standard deviation across the five
generation draws. Neither is a SEM, and the $3\times5$
training--generation combinations are not 15 independent
training replicates. These panels differ from Figure~\ref{fig:temporal_response_gain}
in lag weighting, trigger position, and evaluation population,
so their frozen values need not coincide with that figure.

\subsection{Response Recovery within the Training Range}

Table~\ref{tab:temporal_finetuning_inrange} reports the primary
FT-Variable recipe with normal teaching history. Chronos-2
produces a substantially larger response component along the
injected response, while its profile error remains nonzero.
The gains of $t_0$ and CITRAS-FM increase but remain small under
this budget.

\begin{table}[t]
\centering
\caption{Moving-trigger evaluation over integer lags $0$--$48$,
with trigger onsets $0$--$36$. FT denotes FT-Variable at learning
rate $10^{-5}$ and step 1,000. Entries are mean $\pm$ sample SD:
five generation draws for Frozen, three training seeds for FT.}
\label{tab:temporal_finetuning_inrange}
\small
\setlength{\tabcolsep}{4pt}
\resizebox{\linewidth}{!}{%
\begin{tabular}{lrrrr}
\toprule
& \multicolumn{2}{c}{Gain $G$} & \multicolumn{2}{c}{Profile error $E$ $\downarrow$} \\
\cmidrule(lr){2-3}\cmidrule(lr){4-5}
Model & Frozen & FT & Frozen & FT \\
\midrule
Chronos-2 & $0.2141\pm0.0049$ & $1.0169\pm0.0106$ & $0.8210\pm0.0044$ & $0.3330\pm0.0051$ \\
$t_0$ & $0.0083\pm0.0007$ & $0.0864\pm0.0113$ & $0.9997\pm0.0006$ & $0.9540\pm0.0016$ \\
CITRAS-FM & $-0.0033\pm0.0002$ & $0.0991\pm0.0092$ & $1.0045\pm0.0002$ & $0.9758\pm0.0038$ \\
\bottomrule
\end{tabular}%
}
\end{table}

\paragraph{Training and history controls.}
For Chronos-2, FT-Local at the same learning rate and step has
in-range gain $0.3346\pm0.0198$, while FT-Variable at learning
rate $10^{-6}$ has gain $0.6532\pm0.0226$. In the lag-$33$--$48$
stratum, frozen, FT-Local, and primary FT-Variable gains are
0.0057, 0.0081, and 1.0098, respectively.
For the primary Chronos-2 checkpoint, mean in-range gain changes
from 1.0169 with normal teaching history to 0.0337 after removing
teaching pairs and $-0.0007$ with mismatched history. Each history
variant also changes the context-based normalization, so this
comparison is not an isolated intervention on teaching
information. The additional models retain some response changes
under control histories; the same degree of history specificity
is not established for them.

\subsection{Transfer beyond the Training Range}

Table~\ref{tab:temporal_finetuning_extrapolation} uses the
position-matched panel for Chronos-2. Its in-range reference
therefore differs from Table~\ref{tab:temporal_finetuning_inrange}.
Gain decreases and profile error increases beyond the training
range. The unchanged architecture can express stronger in-range
responses after adaptation, but these results do not uniquely
identify the cause of the pretrained deficit or establish
general recovery across unseen lags.

\begin{table}[t]
\centering
\caption{Chronos-2 extrapolation with trigger onsets matched
to $0$--$4$ for both in-range and unseen lags. Recipe and SD
definitions follow Table~\ref{tab:temporal_finetuning_inrange}.}
\label{tab:temporal_finetuning_extrapolation}
\small
\setlength{\tabcolsep}{4pt}
\resizebox{\linewidth}{!}{%
\begin{tabular}{lrrrr}
\toprule
& \multicolumn{2}{c}{Gain $G$} & \multicolumn{2}{c}{Profile error $E$ $\downarrow$} \\
\cmidrule(lr){2-3}\cmidrule(lr){4-5}
Lag & Frozen & FT & Frozen & FT \\
\midrule
$0$--$48$ & $0.3011\pm0.0041$ & $1.0161\pm0.0247$ & $0.7456\pm0.0029$ & $0.3421\pm0.0079$ \\
56 & $0.0068\pm0.0023$ & $0.3808\pm0.0313$ & $0.9979\pm0.0023$ & $0.7101\pm0.0221$ \\
64 & $0.0026\pm0.0041$ & $0.0977\pm0.0136$ & $1.0089\pm0.0047$ & $0.9463\pm0.0103$ \\
80 & $0.0155\pm0.0081$ & $-0.0135\pm0.0095$ & $1.0253\pm0.0032$ & $1.1902\pm0.0260$ \\
\bottomrule
\end{tabular}%
}
\end{table}

\paragraph{Transfer to the original generator format.}
A separate evaluation uses the original spike-to-positive
generator format, a fixed response anchor, 300 windows per
generation draw, and 13 lags spanning known-future and historical
trigger positions. Table~\ref{tab:temporal_finetuning_transfer}
summarizes the historical-trigger lags $32,40,48,64,96,128$.
Here profile error is computed from each generation draw's mean
response profile before aggregation, unlike the moving-trigger
panels above. This transfer evaluation uses new generation draws
and the adaptation inference conventions; for Chronos-2 these
include BF16, batch size 32, and target-history normalization,
rather than the FP32/batch-100 protocol of the original sweep.
It is therefore a separate transfer comparison, not a replacement
of the frozen results in Figure~\ref{fig:temporal_response_gain}.

\begin{table}[t]
\centering
\caption{Historical-trigger transfer under the primary
FT-Variable recipe. Entries are means over the six listed lags;
the population and profile-error aggregation differ from the
moving-trigger panels.}
\label{tab:temporal_finetuning_transfer}
\small
\setlength{\tabcolsep}{5pt}
\begin{tabular}{lrrrr}
\toprule
& \multicolumn{2}{c}{Gain $G$} & \multicolumn{2}{c}{Profile error $E$ $\downarrow$} \\
\cmidrule(lr){2-3}\cmidrule(lr){4-5}
Model & Frozen & FT & Frozen & FT \\
\midrule
Chronos-2 & 0.0213 & 0.2026 & 0.9795 & 0.8085 \\
$t_0$ & $-0.0006$ & 0.0065 & 1.0007 & 0.9941 \\
CITRAS-FM & 0.0049 & 0.0391 & 0.9954 & 0.9728 \\
\bottomrule
\end{tabular}
\end{table}

\subsection{General-Forecasting Evaluation and Scope}

We also evaluate frozen and fine-tuned checkpoints on the same
26-task, five-family multivariate fev-bench subset
\citep{shchur2026fevbenchrealisticbenchmarktime}, using the
latest two rolling windows and all eligible items. We measure
scaled quantile loss (SQL). For each
training seed, we form the geometric mean of task-level
fine-tuned/frozen SQL ratios, then geometrically aggregate the
three seed-level ratios. Table~\ref{tab:temporal_finetuning_retention}
reports the relative changes for the primary recipe. Positive
values indicate higher SQL. These are aggregate changes on this
subset, not statements that every task deteriorates.

\begin{table}[t]
\centering
\caption{General-forecasting SQL change after primary
FT-Variable training; positive values indicate higher loss.
The same task subset is used within each frozen/FT comparison.}
\label{tab:temporal_finetuning_retention}
\small
\begin{tabular}{lr}
\toprule
Model & Geometric SQL change (\%) \\
\midrule
Chronos-2 & $+10.7$ \\
$t_0$ & $+6.8$ \\
CITRAS-FM & $+16.5$ \\
\bottomrule
\end{tabular}
\end{table}
}

\clearpage
{
\section{Interventions, Mean-Response Structure, and Patch Alignment}
\label{app:response_mechanisms}

\subsection{Scope and Measurements}
\label{app:mechanism_setup}

These exploratory interventions examine Chronos-2 and TimesFM-3,
whose short-lag responses are large enough to compare with their
attenuated responses. They use the same nine trigger--response mappings,
768-step context, 96-step horizon, response anchor 24, and generation
seeds $0$--$4$ as Appendix~\ref{app:temporal_setup}.
For each mapping and seed, the screen retains the fixed first 32
windows of the 300-window draw and evaluates lags $0,4,16$.
The corresponding trigger anchors are 24, 20, and 8: all test triggers
are in the known-future segment. This screen therefore differs from
both the full-window temporal sweep and the historical-trigger probes.
It is not a completed intervention study across all temporal lags.

Within each condition, the event and no-event runs share their
observed target history, backgrounds, historical demonstrations, and
model batching, differing only in the test trigger. For every
intervention we first average paired forecast-difference profiles
across the 32 windows, then compute signed projection gain against
the condition's own response template. The nine mapping quantities
receive equal weight within each seed; reported uncertainties are
sample SEMs over the resulting five seed quantities.
Blocks, mappings, and windows are not additional independent seeds.
Table~\ref{tab:mechanism_profiles} reports this screen's native values;
its 32-window estimates need not equal the primary 300-window results.

\subsection{Activation Patching and Attention Blocking}
\label{app:activation_interventions}

\paragraph{Activation interventions.}
At the output of one Transformer block, we replace the complete
source-channel or target-channel activation sequence of a no-event
recipient with that of the matched event donor. Injection gain is
computed from patched no-event minus native no-event forecasts.
The replaced sequence includes historical and future token positions;
it is not a patch of only the trigger token or a single neuron.
The complementary removal intervention places no-event activations
into an event recipient and measures native event minus patched event.
Thus a removal quantity is removed response, not the response left
after the intervention. Distractor-donor replacements provide
additional controls using the same intervention locations.
Donor and recipient always have the same lag. No lag-zero donor was
used to rescue a lag-four or lag-sixteen recipient.

All block numbers here are one-based. Chronos-2 has 12 blocks and
TimesFM-3 has 20; the same block number does not identify equivalent
relative depth or computation across the two architectures.
Table~\ref{tab:mechanism_focal_injections} gives focal injection
comparisons.

In TimesFM-3, source-injection gain falls from block 11 to 12 while
target-injection gain rises sharply at that transition for all three
lags. In Chronos-2, target-injection effects increase over several
late blocks: block 11 reproduces more than 95\% of the native macro
gain at each tested lag, but the increase is distributed across depths.
The largest successive target-injection increase occurs at block 11
for lags zero and four and block 10 for lag sixteen.
This identifies locations where the native response is accessible
to the target readout, without establishing a single layer where the
delayed relation first fails.

\paragraph{Attention blocking.}
We remove the direct attention edge from target queries to source
keys and values before softmax, within one cross-channel module,
a depth interval, or all modules. The reported blocked gain uses
blocked event minus blocked no-event forecasts and is therefore the
response remaining under the mask. With $G_s$ and $G_s^{\mathrm{mask}}$
denoting the nine-mapping macro gains of seed $s$, suppression is
\begin{equation}
 B_s=100\frac{G_s-G_s^{\mathrm{mask}}}{G_s}.
\end{equation}
We average these five ratios before computing their SEM.
Suppression is relative to the model's own native response, not to
unit ground-truth gain, and is not a ratio of separately pooled means.

Chronos-2 depends on both early and late cross-channel computations.
At block 2 the absolute macro-gain reduction is larger at lag sixteen
than at zero, approximately 0.249 versus 0.137. Blocking block 8 has
the largest single-block macro effect at each tested lag.
For TimesFM-3, blocking block 12 has the largest single-block effect
in 45, 43, and 33 of the 45 mapping--seed cells at lags zero, four,
and sixteen, respectively; it is not the uniquely dominant block in
every cell. Table~\ref{tab:mechanism_attention} reports the focal
suppression percentages with seed-level SEMs.
Blocking every direct source-to-target edge makes the paired response
exactly zero in both models. Effects of individual masks are conditional
on the paths left intact and cannot be summed as independent contributions.

\paragraph{Interpretive limits.}
The interventions show which computations support the native response.
They do not distinguish erased information from information that is
available but not converted to the appropriate response, nor identify
architecture or pretraining as the original cause. Target replacement
at the final block reproduces the native gain and source replacement
there has zero effect, as expected from the target readout; this endpoint
is not evidence that source information disappears at the final layer.
Distractor-donor effects are not uniformly zero, and some TimesFM-3
lag-sixteen mapping--seed native gains are nonpositive.
We consequently base relative suppression on the positive seed-level
macro gain rather than unstable cell-wise divisions by a near-zero gain.
}
\begin{table}[p]
\centering\footnotesize
\caption{Focal activation injections. The donor event activations replace the source or target channel at one block output of the no-event run at the same lag. Entries are projection gains of patched-minus-native no-event forecasts, mean $\pm$ five-seed SEM. Blocks are numbered from one.}
\label{tab:mechanism_focal_injections}
\setlength{\tabcolsep}{3pt}
\resizebox{\linewidth}{!}{%
\begin{tabular}{llrrr}
\toprule
Model & Block & $\tau$ & Source injection & Target injection \\
\midrule
Chronos-2 & 7 & 0 & $0.3135\pm0.0052$ & $0.0138\pm0.0005$ \\
Chronos-2 & 7 & 4 & $0.3210\pm0.0055$ & $0.0183\pm0.0007$ \\
Chronos-2 & 7 & 16 & $0.2693\pm0.0091$ & $0.0161\pm0.0006$ \\
Chronos-2 & 8 & 0 & $0.1634\pm0.0040$ & $0.0759\pm0.0027$ \\
Chronos-2 & 8 & 4 & $0.1592\pm0.0031$ & $0.0899\pm0.0034$ \\
Chronos-2 & 8 & 16 & $0.0978\pm0.0057$ & $0.1200\pm0.0038$ \\
Chronos-2 & 10 & 0 & $0.0179\pm0.0009$ & $0.2620\pm0.0049$ \\
Chronos-2 & 10 & 4 & $0.0200\pm0.0010$ & $0.2855\pm0.0070$ \\
Chronos-2 & 10 & 16 & $0.0149\pm0.0013$ & $0.2784\pm0.0080$ \\
Chronos-2 & 11 & 0 & $0.0003\pm0.0001$ & $0.4473\pm0.0067$ \\
Chronos-2 & 11 & 4 & $-0.0001\pm0.0001$ & $0.4491\pm0.0051$ \\
Chronos-2 & 11 & 16 & $0.0001\pm0.0000$ & $0.3593\pm0.0097$ \\
TimesFM-3 & 11 & 0 & $0.4230\pm0.0069$ & $0.0093\pm0.0010$ \\
TimesFM-3 & 11 & 4 & $0.0485\pm0.0019$ & $0.0063\pm0.0006$ \\
TimesFM-3 & 11 & 16 & $0.0228\pm0.0017$ & $0.0091\pm0.0006$ \\
TimesFM-3 & 12 & 0 & $0.0927\pm0.0022$ & $0.3275\pm0.0063$ \\
TimesFM-3 & 12 & 4 & $0.0046\pm0.0004$ & $0.0574\pm0.0031$ \\
TimesFM-3 & 12 & 16 & $0.0040\pm0.0005$ & $0.0296\pm0.0015$ \\
\bottomrule
\end{tabular}}
\end{table}

\begin{table}[p]
\centering\footnotesize
\caption{Native gain suppressed by attention blocking, in percent. Within each seed the nine-mapping native gain minus blocked gain is divided by that seed\textquotesingle s native gain. Entries are mean $\pm$ SEM of these five ratios, not fractions of ground-truth gain. The direct edge is target query to source keys/values. All-depth blocks every such module.}
\label{tab:mechanism_attention}
\setlength{\tabcolsep}{3pt}
\resizebox{\linewidth}{!}{%
\begin{tabular}{llrrr}
\toprule
Model & Blocked blocks & $\tau=0$ & $\tau=4$ & $\tau=16$ \\
\midrule
Chronos-2 & 2 & $29.20\pm1.23$ & $32.45\pm1.90$ & $68.90\pm1.67$ \\
Chronos-2 & 8 & $50.93\pm0.59$ & $55.03\pm0.99$ & $71.88\pm0.65$ \\
Chronos-2 & All-depth & $100.00\pm0.00$ & $100.00\pm0.00$ & $100.00\pm0.00$ \\
TimesFM-3 & 11 & $49.65\pm0.67$ & $72.92\pm1.01$ & $51.73\pm1.17$ \\
TimesFM-3 & 12 & $72.90\pm0.40$ & $88.23\pm1.14$ & $68.36\pm0.99$ \\
TimesFM-3 & All-depth & $100.00\pm0.00$ & $100.00\pm0.00$ & $100.00\pm0.00$ \\
\bottomrule
\end{tabular}}
\end{table}

{
\subsection{Size, Alignment, and Translation of Mean Responses}
\label{app:mean_response_structure}
\label{app:response_size_alignment}

For a window-averaged response $\overline D_{c,s}$ and its true template
$Q_c$, define size and alignment as
\begin{align}
 A_{c,s}&=\frac{\|\overline D_{c,s}\|_2}{\|Q_c\|_2},\\
 C_{c,s}&=\frac{\langle\overline D_{c,s},Q_c\rangle}
 {\|\overline D_{c,s}\|_2\|Q_c\|_2},\\
 G_{c,s}&=A_{c,s}C_{c,s}.
\end{align}
Alignment is undefined for a zero response. The native mean profiles
in this screen have nonzero norms, so all these quantities are defined.
Table~\ref{tab:mechanism_profiles} averages each measure over mappings
within seed and then over seeds. In particular, the identity applies
before aggregation: the mean gain is the mean of the products,
not the product of the reported mean size and mean alignment.

Chronos-2's mean profile remains closely aligned at the three tested
lags while its size decreases at lag sixteen. TimesFM-3 loses both
size and alignment at nonzero lags. These are descriptions of
window-averaged profiles. A smaller mean can result from weaker
individual responses, cancellation across windows, or both;
these measurements do not establish preserved individual-window
alignment, or Chronos-2 alignment at longer historical lags.

As a separate diagnostic, we translate the response template across
the horizon and find the largest absolute projection, retaining the
signed projection for each candidate location before taking its
absolute value. This descriptive search does not replace the primary
fixed-anchor gain. Even after allowing the best translation,
TimesFM-3's mean absolute projection remains about 0.093 at lag four
and 0.050 at lag sixteen, versus 0.566 at zero lag.
Its attenuation is therefore not explained solely by a translated
mean response of otherwise preserved size.
}
\begin{table}[p]
\centering\footnotesize
\caption{Native responses of the exploratory 32-window screen. Values are mean $\pm$ five-seed SEM after equal weighting of nine mappings within seed. Size and alignment are computed for each seed--mapping mean response before aggregation. Best-shift $|G|$ is the largest absolute projection over translations, not a corrected primary gain.}
\label{tab:mechanism_profiles}
\setlength{\tabcolsep}{3pt}
\resizebox{\linewidth}{!}{%
\begin{tabular}{llrrrrr}
\toprule
Model & $\tau$ & Gain $G$ & Size $A$ & Alignment $C$ & Error $E$ & Best-shift $|G|$ \\
\midrule
Chronos-2 & 0 & $0.4687\pm0.0072$ & $0.4897\pm0.0070$ & $0.9492\pm0.0013$ & $0.5606\pm0.0064$ & $0.4687\pm0.0072$ \\
Chronos-2 & 4 & $0.4599\pm0.0055$ & $0.4749\pm0.0054$ & $0.9552\pm0.0029$ & $0.5599\pm0.0058$ & $0.4599\pm0.0055$ \\
Chronos-2 & 16 & $0.3617\pm0.0099$ & $0.3768\pm0.0107$ & $0.9571\pm0.0015$ & $0.6499\pm0.0088$ & $0.3617\pm0.0099$ \\
TimesFM-3 & 0 & $0.5416\pm0.0073$ & $0.6244\pm0.0079$ & $0.8621\pm0.0019$ & $0.5714\pm0.0058$ & $0.5656\pm0.0084$ \\
TimesFM-3 & 4 & $0.0764\pm0.0037$ & $0.1235\pm0.0048$ & $0.5577\pm0.0239$ & $0.9298\pm0.0035$ & $0.0933\pm0.0040$ \\
TimesFM-3 & 16 & $0.0352\pm0.0020$ & $0.0762\pm0.0015$ & $0.3962\pm0.0152$ & $0.9678\pm0.0019$ & $0.0505\pm0.0022$ \\
\bottomrule
\end{tabular}}
\end{table}

{
\subsection{Joint Trigger--Response Offset Sweep}
\label{app:patch_phase_sweep}
\label{app:joint_alignment}

The separate TimesFM-3 alignment experiment evaluates all response
anchors $16,17,\ldots,47$ at lags $0,4,8,16$.
It uses nine mappings, five seeds, and all 300 windows per mapping
and seed, retaining the same history and paired backgrounds within
each lag while jointly shifting the test trigger and target response.
For each fixed lag, these 32 offsets visit one complete input-patch
phase cycle. Across lags, historical teaching-response alignment
also follows the lag-specific generation rules.
For each anchor and lag, mapping-specific window-mean projection
gains receive equal weight within seed before averaging over the
five generation seeds.

Across anchors, lag-zero macro gains range from 0.499 to 0.698,
whereas lag-four gains range from 0.049 to 0.148.
Lag-eight and lag-sixteen gains range from 0.023 to 0.077 and
from 0.012 to 0.064, respectively.
At every matched anchor the lag-four mean is lower, retaining only
7.9--23.1\% of the corresponding lag-zero mean.
The largest lag-four mean is below the smallest lag-zero mean.
The same direction holds in all 160 anchor--seed macro comparisons;
these overlapping comparisons are not 160 independent generation seeds.
Thus the short-lag deficit is not confined to one unfavorable
joint placement. This 32-offset experiment is distinct from the
four-anchor, single-mapping control in
Appendix~\ref{app:temporal_position_controls}.

TimesFM-3 uses 32-step input patches. In the default spike-to-positive
case at response anchor 24 and lag four, the nominal trigger occupies
future positions 20--23 and its nonzero support occupies 21--22;
both lie within one input patch. The nominal target response occupies
24--35 and intersects the output stitching boundary at 32.
The sweep changes input phase, response position, and output stitching
together. It therefore constrains a boundary-placement explanation
without independently identifying an effect of input patching or
compression; subpatch temporal resolution remains a candidate,
not an established unique cause.
}

\clearpage
{
\section{Context-Length Controls}
\label{app:context_controls}
\label{app:bc_temporal_context}
\label{app:bc_structural_context}
\label{app:kc_context_sensitivity}
\label{app:ade_reliability_context}
\label{app:f_context_sensitivity}

\paragraph{Paired synthetic populations.}
The synthetic summary reports $T\in\{512,768,1024\}$, using exact suffixes of
one longest-context history per draw, with identical future targets
and counterfactuals across lengths. These paired populations are
separate from the primary fixed-context experiments, including
where the nominal context length agrees.
The temporal experiment uses nine mappings, thirteen lags, five
seeds, 300 windows per mapping and seed, and a 96-step horizon.
Its teaching schedule fills the longer interval at the same spacing,
so longer histories also contain more demonstrations.
The structural, lagged-copy, and reliability experiments use five
seeds, 256 paired windows per seed, and an eight-step horizon.
Their source and noise parameters follow the corresponding primary
experiments; the structural conditional reference is recomputed
from exactly the supplied context. Fitted SSM, VAR, SparseDirect,
and event-response results from the primary experiments are not
relabeled as refits on these populations.

\paragraph{Selected synthetic comparisons.}
Table~\ref{tab:context_synthetic_summary} retains zero-lag and
historical-lag temporal gains, direct latent access, and independent,
duplicate, and distractor channel comparisons at strong persistence.
It also retains irrelevant-future sensitivity and a forecast-time
reliability penalty. The table selects representative conditions;
it does not reproduce every lag, persistence, channel count, relation
strength, or reliability level in the context sweep. The 2,048-step
synthetic results are omitted from this summary.
The structural percentages use each method's own seed-level
reference. Means receive equal seed weights, with equal mapping
weights for the temporal experiment.
Context length changes the size of the effects but does not
uniformly remove them. CITRAS-FM's 2,048-step synthetic requests
exceed its recorded 1,032-step context limit and are unavailable,
without substituting shorter requests.

\paragraph{Real-data histories.}
Tables~\ref{tab:context_real_sql} and~\ref{tab:context_real_coverage}
retain within-context SQL comparisons and interval coverage.
The fixed target-eligible cohorts contain 160 GFC17, 1,340 KDD,
and 10,610 Rossmann series--origin pairs. At each cap, official
origins, targets, and masks remain fixed, while source forecasts
are recomputed from the available history.
This control uses nine fixed-rank paths and fits no calibration;
it is distinct from the primary 100-sample Trajectory construction.
The seasonal SQL denominator is recomputed from the supplied history.
Within-context routes share that denominator, but cross-context SQL
changes do not isolate changes in forecast error.
CITRAS-FM cannot accept the requested 2,048-step GFC17 and KDD
histories. Rossmann histories saturate below the larger caps,
so the 1,024- and 2,048-cap scores coincide and remain supported
for CITRAS-FM. Coverage is shown at the shortest and longest
requested caps to retain interval evidence without the full grid.
}
\begin{table}[t]
\centering\footnotesize
\caption{Focal synthetic context comparisons (five-seed means). $G_0,G_{32}$ are temporal gains. The four $g$ columns are percentage MAE changes: direct latent access ($g_{\rm latent}$) uses the target-only reference, while independent, duplicate, and distractor channels use its own noisy-$K=1$ reference. Structural conditions fix $\rho=0.95$ and $K=16$ for the channel-count comparisons. $S_{\rm future}$ uses $\beta=0.8,L=8$; $\Delta_{\rm shift}$ uses $\lambda=0$. The analytic row concerns the structural process only. Dashes denote unsupported or unevaluated combinations; this table selects conditions rather than reproducing the full context grids.}
\label{tab:context_synthetic_summary}
\setlength{\tabcolsep}{3pt}
\resizebox{\linewidth}{!}{%
\begin{tabular}{rlrrrrrrrr}
\toprule
$T$ & Model & $G_0$ & $G_{32}$ & $g_{\rm latent}$ & $g_{\rm ind}$ & $g_{\rm dup}$ & $g_{\rm dis}$ & $S_{\rm future}$ & $\Delta_{\rm shift}$ \\
\midrule
512 & Chronos-2 & $0.297$ & $0.012$ & $4.88$ & $2.41$ & $0.37$ & $0.53$ & $0.162$ & $0.473$ \\
512 & TimesFM-3 & $0.487$ & $0.007$ & $4.95$ & $2.95$ & $-0.21$ & $-0.05$ & $0.114$ & $0.498$ \\
512 & TiRex-2 & $0.014$ & $0.002$ & $1.02$ & $1.17$ & $-0.09$ & $-0.02$ & $0.129$ & $0.457$ \\
512 & $t_0$ & $0.032$ & $-0.001$ & $3.75$ & $0.69$ & $0.29$ & $0.80$ & $0.075$ & $0.371$ \\
512 & CITRAS-FM & $0.003$ & $0.002$ & $0.66$ & $0.81$ & $-0.37$ & $-0.07$ & $0.021$ & $0.076$ \\
512 & Analytic & --- & --- & $5.08$ & $3.12$ & $0.00$ & $0.00$ & --- & --- \\
\midrule
768 & Chronos-2 & $0.390$ & $0.023$ & $4.77$ & $1.93$ & $0.33$ & $1.51$ & $0.135$ & $0.472$ \\
768 & TimesFM-3 & $0.467$ & $0.007$ & $4.84$ & $3.09$ & $-0.14$ & $-0.10$ & $0.121$ & $0.497$ \\
768 & TiRex-2 & $0.009$ & $0.001$ & $1.34$ & $1.26$ & $-0.02$ & $0.16$ & $0.135$ & $0.458$ \\
768 & $t_0$ & $0.042$ & $-0.001$ & $3.38$ & $0.91$ & $0.34$ & $0.69$ & $0.065$ & $0.395$ \\
768 & CITRAS-FM & $0.004$ & $0.003$ & $0.57$ & $0.69$ & $-0.31$ & $0.10$ & $0.019$ & $0.074$ \\
768 & Analytic & --- & --- & $5.08$ & $3.12$ & $0.00$ & $0.00$ & --- & --- \\
\midrule
1024 & Chronos-2 & $0.457$ & $0.027$ & $4.56$ & $1.73$ & $0.19$ & $1.15$ & $0.120$ & $0.472$ \\
1024 & TimesFM-3 & $0.481$ & $0.005$ & $4.58$ & $3.15$ & $0.12$ & $0.00$ & $0.124$ & $0.501$ \\
1024 & TiRex-2 & $0.009$ & $0.001$ & $1.47$ & $1.38$ & $0.04$ & $0.23$ & $0.136$ & $0.455$ \\
1024 & $t_0$ & $0.049$ & $-0.001$ & $2.94$ & $0.87$ & $0.43$ & $0.70$ & $0.059$ & $0.405$ \\
1024 & CITRAS-FM & $0.004$ & $0.003$ & $0.55$ & $0.69$ & $-0.32$ & $0.11$ & $0.017$ & $0.085$ \\
1024 & Analytic & --- & --- & $5.08$ & $3.12$ & $0.00$ & $0.00$ & --- & --- \\
\midrule
2048 & Chronos-2 & $0.496$ & $0.035$ & $4.48$ & $1.75$ & $0.19$ & $0.98$ & $0.095$ & $0.481$ \\
2048 & TimesFM-3 & $0.519$ & $0.007$ & $4.51$ & $2.88$ & $0.18$ & $0.01$ & $0.134$ & $0.497$ \\
2048 & TiRex-2 & $0.007$ & $0.000$ & $2.18$ & $1.22$ & $0.14$ & $0.87$ & $0.139$ & $0.454$ \\
2048 & $t_0$ & $0.066$ & $-0.002$ & $2.17$ & $1.04$ & $0.32$ & $0.43$ & $0.048$ & $0.425$ \\
2048 & CITRAS-FM & --- & --- & --- & --- & --- & --- & --- & --- \\
2048 & Analytic & --- & --- & $5.08$ & $3.12$ & $0.00$ & $0.00$ & --- & --- \\
\bottomrule
\end{tabular}}
\end{table}

\begin{table}[t]
\centering\footnotesize
\caption{Real-data context comparisons. Each cell is $(\Delta_M,\Delta_P)$ in percent, where $\Delta_M=100(\mathrm{SQL}_{\rm Median}/\mathrm{SQL}_{\rm MV}-1)$ and $\Delta_P=100(\mathrm{SQL}_{\rm Rank9}/\mathrm{SQL}_{\rm Median}-1)$. Negative values denote improvement. Rank9 uses nine fixed-rank paths, separately from the primary 100-path Trajectory method. Dashes indicate unsupported requests.}
\label{tab:context_real_sql}
\setlength{\tabcolsep}{3pt}
\resizebox{\linewidth}{!}{%
\begin{tabular}{llrrrr}
\toprule
Dataset & Model & $T=512$ & $T=768$ & $T=1024$ & $T=2048$ \\
\midrule
GFC17 & Chronos-2 & $(9.8,-6.1)$ & $(12.8,-7.2)$ & $(10.7,-6.8)$ & $(8.0,-6.0)$ \\
 & TimesFM-3 & $(4.4,-4.2)$ & $(4.6,-3.9)$ & $(3.3,-4.3)$ & $(-2.7,-2.7)$ \\
 & TiRex-2 & $(1.3,-1.0)$ & $(1.2,-0.8)$ & $(0.6,-1.0)$ & $(-0.6,-0.5)$ \\
 & $t_0$ & $(2.7,-1.1)$ & $(2.0,-1.1)$ & $(2.6,-1.9)$ & $(-0.7,-1.9)$ \\
 & CITRAS-FM & $(3.1,-0.9)$ & $(1.0,-0.5)$ & $(0.6,-0.5)$ & --- \\
\midrule
KDD & Chronos-2 & $(11.8,-6.7)$ & $(3.1,-7.1)$ & $(6.9,-9.4)$ & $(10.3,-11.6)$ \\
 & TimesFM-3 & $(15.5,-12.8)$ & $(16.7,-14.4)$ & $(19.4,-15.1)$ & $(14.9,-14.4)$ \\
 & TiRex-2 & $(14.2,-12.1)$ & $(16.4,-14.4)$ & $(16.0,-15.0)$ & $(15.6,-15.4)$ \\
 & $t_0$ & $(10.5,-10.7)$ & $(17.8,-13.5)$ & $(15.3,-12.9)$ & $(13.6,-14.7)$ \\
 & CITRAS-FM & $(5.9,-4.1)$ & $(6.6,-5.1)$ & $(4.3,-4.8)$ & --- \\
\midrule
Rossmann & Chronos-2 & $(12.8,-9.3)$ & $(11.1,-8.6)$ & $(10.9,-8.6)$ & $(10.9,-8.6)$ \\
 & TimesFM-3 & $(12.9,-9.9)$ & $(12.0,-9.1)$ & $(12.2,-9.1)$ & $(12.2,-9.1)$ \\
 & TiRex-2 & $(7.6,-6.3)$ & $(7.6,-6.5)$ & $(7.9,-6.6)$ & $(7.9,-6.6)$ \\
 & $t_0$ & $(12.3,-9.1)$ & $(11.7,-8.8)$ & $(11.7,-8.6)$ & $(11.7,-8.6)$ \\
 & CITRAS-FM & $(1.9,-0.5)$ & $(1.9,-0.5)$ & $(2.0,-0.6)$ & $(2.0,-0.6)$ \\
\bottomrule
\end{tabular}}
\end{table}

\begin{table}[t]
\centering\footnotesize
\caption{Real-data 80\% interval coverage at the shortest and longest requested context caps (percent). Each triple gives MV, Median, and Rank9 coverage. These selected endpoints retain interval evidence without the complete route-by-context grid. The nominal reference is 80\%; dashes indicate unsupported requests.}
\label{tab:context_real_coverage}
\setlength{\tabcolsep}{3pt}
\resizebox{\linewidth}{!}{%
\begin{tabular}{llrr}
\toprule
Dataset & Model & $T=512$: MV / Median / Rank9 & $T=2048$: MV / Median / Rank9 \\
\midrule
GFC17 & Chronos-2 & $69.4 / 46.2 / 59.6$ & $72.8 / 45.4 / 62.2$ \\
 & TimesFM-3 & $71.2 / 53.2 / 64.2$ & $80.8 / 66.3 / 77.0$ \\
 & TiRex-2 & $71.4 / 60.9 / 62.9$ & $77.7 / 67.8 / 68.6$ \\
 & $t_0$ & $70.3 / 59.3 / 60.9$ & $78.8 / 66.3 / 71.6$ \\
 & CITRAS-FM & $46.2 / 39.0 / 40.7$ & --- \\
\midrule
KDD & Chronos-2 & $71.7 / 46.1 / 68.7$ & $78.6 / 44.5 / 72.2$ \\
 & TimesFM-3 & $71.7 / 25.5 / 65.7$ & $84.7 / 23.1 / 80.7$ \\
 & TiRex-2 & $84.0 / 30.4 / 72.1$ & $90.3 / 23.0 / 83.9$ \\
 & $t_0$ & $79.8 / 30.1 / 58.2$ & $86.5 / 35.4 / 77.0$ \\
 & CITRAS-FM & $53.0 / 34.3 / 42.6$ & --- \\
\midrule
Rossmann & Chronos-2 & $76.1 / 36.6 / 65.3$ & $75.2 / 37.0 / 65.1$ \\
 & TimesFM-3 & $82.3 / 36.1 / 72.0$ & $81.5 / 35.8 / 70.5$ \\
 & TiRex-2 & $79.7 / 46.8 / 75.3$ & $78.8 / 45.5 / 74.0$ \\
 & $t_0$ & $80.6 / 36.0 / 71.7$ & $79.8 / 37.4 / 70.8$ \\
 & CITRAS-FM & $64.8 / 55.3 / 57.1$ & $64.7 / 55.3 / 57.4$ \\
\bottomrule
\end{tabular}}
\end{table}

\end{document}